\documentclass[10pt,twocolumn,letterpaper]{article}

\usepackage{cvpr}
\usepackage{times}
\usepackage{epsfig}
\usepackage{graphicx}
\usepackage{amsmath}
\usepackage{amssymb}
\usepackage{multirow}
\usepackage{url}
\usepackage{subfigure}
\usepackage{booktabs}
\usepackage[misc]{ifsym}

\DeclareMathOperator*{\argmin}{arg\,min}

\newcommand{\rulev}{\unskip\ \vrule\ }
\newcommand{\ruleh}{\unskip\ \hrule\ }

\usepackage[breaklinks=true,bookmarks=false]{hyperref}

\cvprfinalcopy 

\def\cvprPaperID{****} 

\ifcvprfinal\fi

\begin{document}

\title{Separating Style and Content for Generalized Style Transfer}

\author{Yexun Zhang\\
Shanghai Jiao Tong University\\
{\tt\small zhyxun@sjtu.edu.cn}
\and
Ya Zhang\textsuperscript{\Letter}\\
Shanghai Jiao Tong University\\
{\tt\small ya$\_$zhang@sjtu.edu.cn}
\and
Wenbin Cai\\
Microsoft \\
{\tt\small wenbca@microsoft.com}
\and
Jie Chang\\
Shanghai Jiao Tong University\\
{\tt\small j$\_$chang@sjtu.edu.cn}
}


\maketitle
\thispagestyle{empty}

\begin{abstract}
Neural style transfer has drawn broad attention in recent years. However, most existing methods aim to explicitly model the transformation between different styles, and the learned model is thus not generalizable to new styles. We here attempt to separate the representations for styles and contents, and propose a generalized style transfer network consisting of style encoder, content encoder, mixer and decoder. The style encoder and content encoder are used to extract the style and content factors from the style reference images and content reference images, respectively. The mixer employs a bilinear model to integrate the above two factors and finally feeds it into a decoder to generate images with target style and content. To separate the style features and content features, we leverage the conditional dependence of styles and contents given an image. During training, the encoder network learns to extract styles and contents from two sets of reference images in limited size, one with shared style and the other with shared content. This learning framework allows simultaneous style transfer among multiple styles and can be deemed as a special `multi-task' learning scenario. The encoders are expected to capture the underlying features for different styles and contents which is generalizable to new styles and contents. For validation, we applied the proposed algorithm to the Chinese Typeface transfer problem. Extensive experiment results on character generation have demonstrated the effectiveness and robustness of our method.

\end{abstract}

\section{Introduction}

In recent years, style transfer, as an interesting application of deep neural networks (DNNs), has increasingly attracted attention among the research community. Existing studies either apply an iterative optimization mechanism~\cite{Gatys} or directly learn a feed-forward generator network to force the output image to be with target style and target contents~\cite{johnson,ulyanov}. A set of losses are accordingly proposed for the transfer network, such as the pix-wise loss~\cite{isola}, the perceptual loss~\cite{johnson,zhang2017multi}, and the histogram loss~\cite{wilmot}. Recently, several variations of generative adversarial networks (GANs)~\cite{liu2016,Zhu2017} are introduced by adding a discriminator to the style transfer network which incorporates adversarial loss with transfer loss to generate better images. However, these studies aim to explicitly learn the transformation from a certain source style to a given target style, and the learned model is thus not generalizable to new styles, i.e. retraining is needed for transformations of new styles which is time-consuming.

\begin{figure}[!tbp]
\setlength{\abovecaptionskip}{-2pt}
\centering
\hspace{-5pt}
\includegraphics[height=1.0in,width=3.3in]{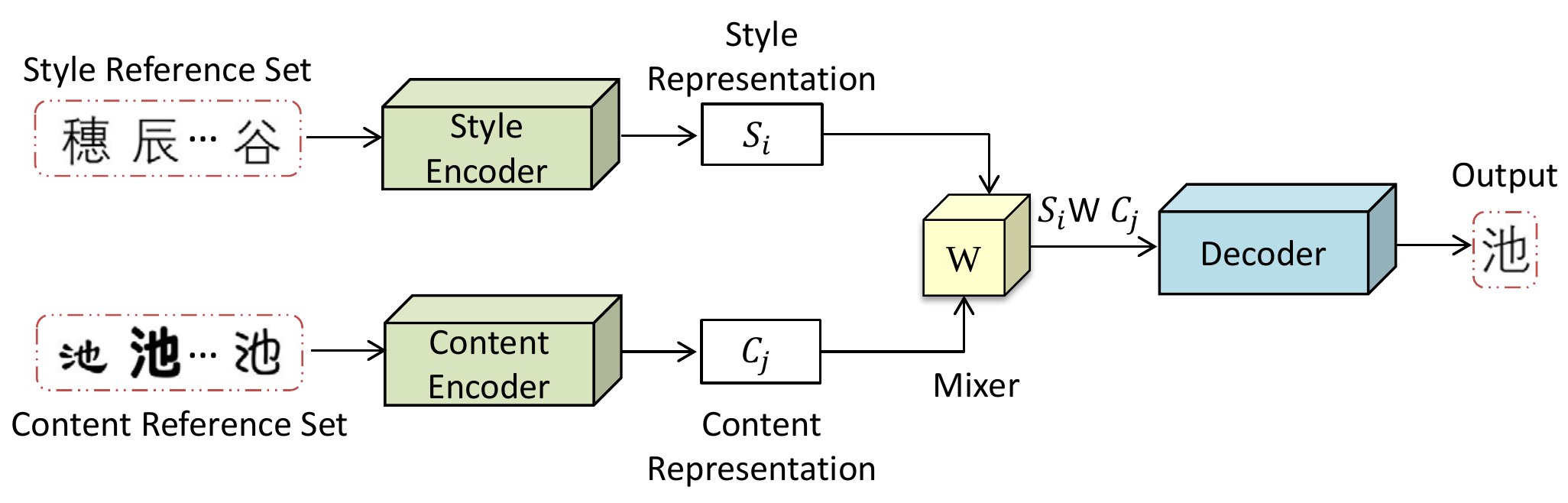}
\caption{The framework of the proposed {\em EMD} model.}
\label{fig:framework1}
\vspace{-15pt}
\end{figure}

\begin{table*}[!t]\footnotesize{}
    \centering
    \caption{Comparison of {\em EMD} with existing methods.}
    \begin{tabular}{c|c|c|c|c}
    \hline
        Methods &  Data format & Generalizable to new styles? & Requirements for new style transfer & What the model learned? \\
    \hline Pix2pix~\cite{isola} & paired & \multirow{8}{3.2cm}{The learned model can only transfer images to styles which appeared in the training set. For new styles, the model has to be retrained.}
    & \multirow{6}{4cm}{Retrain on a lot of training images for a source style and a target style.}  & \multirow{6}{4.5cm}{The translation from a certain source style to a specific target style.}\\
    \cline{1-2} CoGAN~\cite{liu2016} &  unpaired &  & & \\
    \cline{1-2} CycleGAN~\cite{Zhu2017} &  unpaired  & &  \\
    \cline{1-2} Rewrite~\cite{rewrite} &  paired  &  & &\\
    \cline{1-2} Zi-to-zi~\cite{zitozi} & paired  &  & &\\
    \cline{1-2} AEGN~\cite{lyu} & paired  &   & & \\
    \cline{1-2} \cline{4-5} Perceptual~\cite{johnson} & unpaired  &  &  \multirow{2}{4cm}{Retrain on many input content images and one style image.} &
    \multirow{2}{4.5cm}{Transformation among specific styles.}\\
    \cline{1-2} StyleBank~\cite{chen2017} & unpaired  &  & & \\
    \hline  Patch-based~\cite{chen2016fast} & unpaired  & \multirow{3}{3.2cm}{The learned model can be generalized to new styles.}  & \multirow{3}{4cm}{One or a small set of style/content reference images.} & The swap of style/content feature maps. \\
    \cline{1-2} \cline{5-5} AdaIn~\cite{Huang_2017_ICCV} & unpaired & & & The transferring of feature statistics. \\
    \cline{1-2} \cline{5-5} EMD & triplet &  &  & The feature representation of style/content. \\

    \hline
    \end{tabular}%
    \vspace{-10pt}
  \label{tab:comp_methods}%
\end{table*}%

In this paper, we propose a novel generalized style transfer network which can extend well to new styles or contents. Different from existing supervised style transfer methods, where an individual transfer network is built for each pair of style transfer, the proposed network represents each style or content with a small set of reference images and attempts to learn separate representations for styles and contents. Then, to generate an image of a given style-content combination is simply to mix the corresponding two representations. This learning framework allows simultaneous style transfer among multiple styles and can be deemed as a special `multi-task' learning scenario. Through separated style and content representations, the network is able to generate images of all style-content combination given the corresponding reference sets, and is therefore expected to generalize well to new styles and contents.
To our best knowledge, the study most resembles to ours is the bilinear model proposed by Tenenbaum and Freeman~\cite{Tenenbaum}, which obtained independent style and content representations through matrix decomposition. However, it usually requires an exhaustive enumeration of examples for accurate decomposition of new styles and contents, which may not be readily available for some styles/contents.

As shown in Figure~\ref{fig:framework1}, the proposed style transfer network, denoted as {\em EMD} thereafter, consists of a style encoder, a content encoder, a mixer, and a decoder. Given a set of reference images, the style/content encoder leverages the conditional dependence of styles and contents to learn style/content representations. The mixer then combines the corresponding style and content representations using a bilinear model. The decoder finally generates the target images based on the combined representations. Each training example for the proposed network is provided as a triplet $<$$\mathcal{R}_{S_i}$, $\mathcal{R}_{C_j}$, $I_{ij}$$>$, where $I_{ij}$ is the target image of style $S_i$ and content $C_j$. $\mathcal{R}_{S_i}$ and $\mathcal{R}_{C_j}$ are respectively the style and content reference sets, each consisting of $r$ random images of the corresponding style $S_i$ and content $C_j$. The entire network is trained end-to-end with a weighted $L1$ loss measuring the difference between the generated images and the target images.
As it is difficult to validate the decomposition of style and content for images, we here use the character typeface transfer as a special case of style transfer to validate the proposed method. Extensive experiment results have demonstrated the effectiveness and robustness of our method for style transfer.
The main contributions of our study are summarized as follows.
\begin{itemize}
\item We propose a generalized style transfer network which is able to generate images of any unseen style/content given a small set of reference images.
\item The network decomposes an image into separate style and content representations, taking advantages of the conditional dependence of contents and styles.
\item This learning framework allows simultaneous style transfer among multiple styles and can be deemed as a special `multi-task' learning scenario.
\end{itemize}

\begin{figure*}[!tbp]
\setlength{\abovecaptionskip}{-2pt}
\centering
\includegraphics[height=2.5in,width=7in]{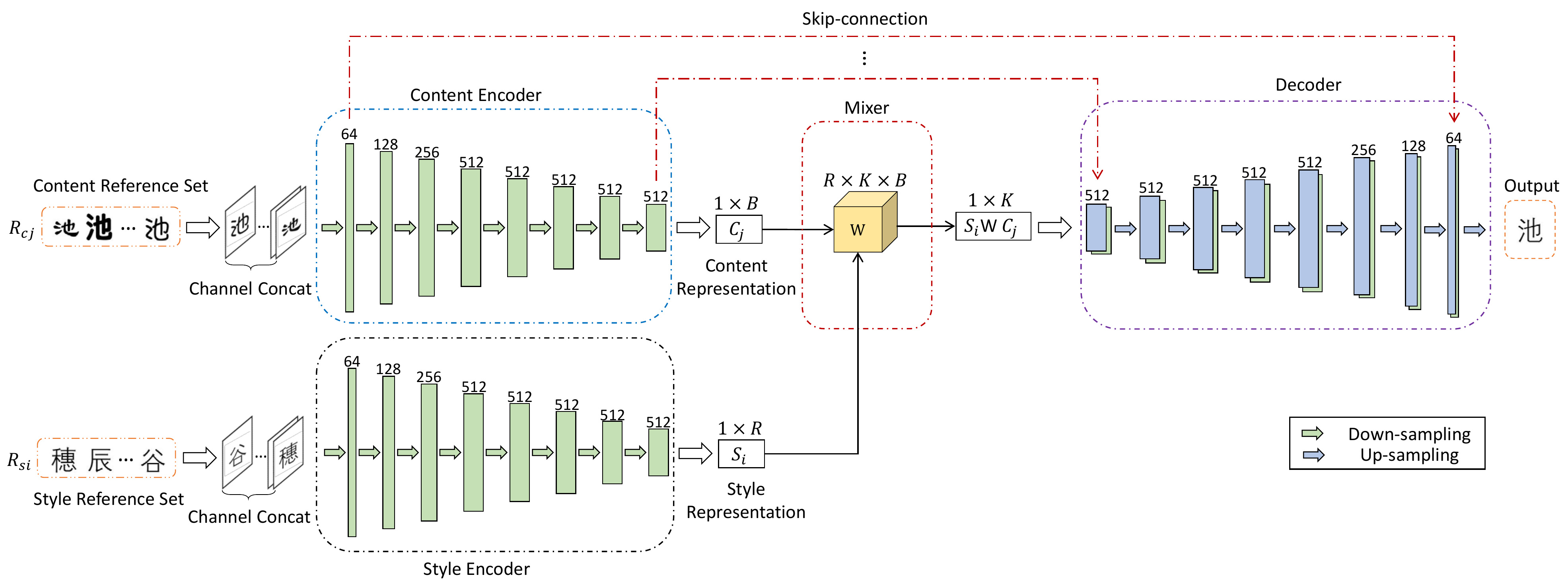}
\caption{The detailed architecture of the proposed generalized {\em EMD} model for style transfer.}
\label{fig:framework2}
\vspace{-5pt}
\end{figure*}

\section{Related Work}

\textbf{Neural Style Transfer.} DeepDream~\cite{mordvintsev} may be the first attempt to generate artistic work using Convolution Neural Networks (CNNs). Then Gatys et. al successfully applied CNNs to neural style transfer~\cite{Gatys}. They generate the target image by optimizing a noise image iteratively using a pretrained network, which is time-consuming. Therefore, many studies have been done for finding a way to directly learn a feed-forward generator network. Johnson et. al proposed a perceptual loss function to help neural style transfer~\cite{johnson}. Ulyanov et. al proposed a texture network for both texture synthesis and style transfer~\cite{ulyanov}. Further, Chen et. al proposed the stylebank to represent each style by a convolution filter, which can simultaneously learn numerous styles~\cite{chen2017}. For arbitrary neural style transfer, \cite{chen2016fast} proposed a patch-based method to replace each content feature patch with the nearest style feature. Further, \cite{Huang_2017_ICCV} proposed a faster method based on adaptive instance normalization which performed style transfer in the feature space by transferring feature statistics.

\textbf{Image-to-Image Translation.} Image-to-image translation is to learn the mapping from the input image to output image, such as from edges to real objects. Pix2pix~\cite{isola} used a conditional GAN based network which needs paired data for training. However, paired data are hard to collect in many applications. Therefore, some methods with no need for paired data are proposed. Liu and Tuzel proposed the coupled GAN (CoGAN)~\cite{liu2016} for learning a joint distribution of two domains by a weight sharing way. Later, Liu~\cite{liu2017} extended the CoGAN to unsupervised image-to-image translation problem. Some other studies~\cite{Bousmalis,Shrivastava,Taigmand} encourage the input and output to share certain content even though they may differ in style by enforcing the output to be close to the input in a predefined metric space, such as class label space and so on. Recently, Zhu et. al proposed the cycle-consistent adversarial network (CycleGAN)~\cite{Zhu2017} which performs well for many vision and graphics tasks.

\textbf{Character Style Transfer.} Most existing studies take character style transfer as an image translation task. The ``Rewrite" project uses a simple traditional flavour top-down CNNs structure and can transfer a typographic font to another stylized typographic font~\cite{rewrite}. As the improvement version, the ``zi-to-zi" project can transfer multiple styles by assigning each style an one-hot category label and training the network by a supervised way~\cite{zitozi}. The recent work ``From A to Z" also adopts a supervised method and assigns each character an one-hot label~\cite{upchurch}. Lyu et. al proposed an auto-encoder guided GAN network (AEGN) which can synthesize calligraphy images with specified style from standard Chinese font images~\cite{lyu}.

However, most of the methods reviewed above can only transfer styles in the training set and the network must be retrained for new styles. In contrast, the proposed {\em EMD} network can generate images with novel styles/contents given only a small set of reference images. We present a comparison of the methods in Table~\ref{tab:comp_methods}.

\section{Generalized Style Transfer Model}

In this section, we present the details of the proposed generalized style transfer model {\em EMD}. The whole model is an encoder-decoder network which consists of four subnets: $\textsl{Style Encoder}$, $\textsl{Content Encoder}$, $\textsl{Mixer}$ and $\textsl{Decoder}$, as shown in Figure~\ref{fig:framework2}. First, the $\textsl{Style/Content Encoder}$  extracts style/content representations given style/content reference images. Next, the $\textsl{Mixer}$ integrates the style feature and content feature and the combined feature is then fed into the $\textsl{Decoder}$. Finally, the $\textsl{Decoder}$ generates the image with the target style and content.

\subsection{Encoder Network}
To achieve the generation of images with arbitrary style and content, it is crucial to separate the style and content explicitly. The $\textsl{Style Encoder}$ and $\textsl{Content Encoder}$ are designed for this purpose. They both have the same architecture, consisting of a series of Convolution-BatchNorm-LeakyReLU down-sampling blocks which yield 1$\times$1 latent feature representations of the input style/content reference images. The first convolution layer is with $5\times5$ kernel and stride 1 and the rest are with $3\times3$ kernel and stride 2. All ReLUs are leaky, with slope 0.2.

The input to the $\textsl{Style Encoder}$ and $\textsl{Content Encoder}$ are style reference set $\mathcal{R}_{S_i}$ and content reference set $\mathcal{R}_{C_j}$, respectively. $\mathcal{R}_{S_i}$ consists of $r$ reference images with the same style $S_i$ but different contents ${C_{j_1},C_{j_2},\ldots,C_{j_r}}$
\begin{equation}
\mathcal{R}_{S_i} = \{I_{ij_1}, I_{ij_2}, \ldots, I_{ij_r} \},
\end{equation}
where $I_{ij}$ represents the image with style $S_i$ and content $C_j$.

Similarly, $\mathcal{R}_{C_j}$ is for content $C_j$ $(j=1,2,\ldots,m)$ and consists of $r$ reference images with the same content $C_j$ but different styles ${S_{i_1},S_{i_2},\ldots,S_{i_r}}$
\begin{equation}
\mathcal{R}_{C_j} = \{I_{i_1j}, I_{i_2j}, \ldots, I_{i_rj} \}.
\end{equation}
The $r$ reference images are concatenated in the channel dimension to feed in to the encoders. This allows the encoders to capture the common characteristics among images of the same style/content.

\subsection{Mixer Network}
With the style representations and content representations obtained by the $\textsl{Style Encoder}$ and $\textsl{Content Encoder}$, we combine the two factors by the $\textsl{Mixer}$ which is a bilinear model. The bilinear models are two-factor models with the mathematical property of separability: their outputs are linear in either factor when the other is held constant, which has been demonstrated that the influences of two factors can be efficiently separated and combined in a flexible representation that can be naturally generalized to unfamiliar factor classes~\cite{Tenenbaum}, such as new styles. Furthermore, the bilinear model has also been successfully used in zero-shot learning as a compatibility function to associate visual representation and auxiliary class text description~\cite{changpinyo,frome,xian}. The learned compatibility function can be seen as the shared knowledge and transferred to new classes. Here, we take the bilinear model to integrate styles and contents together and the combination function can be formulated as
\begin{equation}
F_{ij} = S_i\textbf{W}C_j,
\end{equation}
where $\textbf{W}$ is a tensor with size $R\times K \times B$, $S_i$ is the $R$-dimensional style feature and $C_j$ is the $B$-dimensional content feature. $F_{ij}$ can be seen as the $K$-dimensional feature vector of image $I_{ij}$ which will be taken as the input of the $\textsl{Decoder}$ to generate the image with style $S_i$ and content $C_j$.

\subsection{Decoder Network}

The image generator is a typical decoder network which is symmetrical to the encoder and maps the combined feature representation to the output image with target style and content. The $\textsl{Decoder}$ roughly follows the architectural guidelines set forth by Radford et. al~\cite{Radford} and consists of a series of Deconvolution-BatchNorm-ReLU up-sampling blocks except the last layer which only contains the deconvolution layer. Other than the last layer which uses 5$\times$5 kernels and stride 1, all deconvolution layers use 3$\times$3 kernels and stride 2. The outputs are transformed into $0\sim 1$ by the sigmoid function.

In addition, since the stride convolution in $\textsl{Style Encoder}$ and $\textsl{Content Encoder}$ is detrimental to spatial information extraction, we adopt the skip-connection which has been commonly used in semantic segmentation tasks~\cite{jegou,long,ronneberger} to refine the segmentation using spatial information from different resolutions. Here, based on the fact that though the content inputs and outputs differ in appearances, they share the same structure, we concatenate the input feature map of each up-sampling block with the corresponding output of the symmetrical down-sampling block in $\textsl{Content Encoder}$ to allow the $\textsl{Decoder}$ to learn back the relevant structure information lost during the down-sampling process.

\subsection{Loss Function}

Given a set of training examples $\mathcal{D}_t$, the training objective is defined as
\begin{alignat}{3}
\theta = \argmin_{\theta} \sum_{I_{ij} \in \mathcal{D}_t} L(\hat{I}_{ij},I_{ij}|\mathcal{R}_{S_i},\mathcal{R}_{C_j}),
\label{eq:model}
\end{alignat}
where $\theta$ represents model parameters, $\hat{I}_{ij}$ is the generated image and $L(\hat{I}_{ij},I_{ij}|\mathcal{R}_{S_i},\mathcal{R}_{C_j})$ is the generation loss which can be written as
\begin{equation}
L(\hat{I}_{ij},I_{ij}|\mathcal{R}_{S_i},\mathcal{R}_{C_j}) = W^{ij}_{st}\times W^{ij}_b \times ||\hat{I}_{ij} - I_{ij}||.
\end{equation}
We use pixel-wise L1 loss as our generation loss for character typeface transfer problem rather than L2 loss since L1 loss tends to yield sharper and cleaner images~\cite{isola,lyu}.

$W^{ij}_{st}$ and $W^{ij}_b$ are two weights for target image $I_{ij}$ which are added to alleviate the imbalance in the target set induced by the random sampling. In each learning iteration, the size and thickness of target images in the target set may vary greatly and the model will be optimized mainly for target images containing characters which have more pixels and cause more losses, such as those big and thick characters. Moreover, models trained using L1 loss may pay more attention to blacker characters and perform poorly on images with lighter characters. To alleviate these imbalance, we add these two weights on the generation loss: $W^{ij}_{st}$ about the size and thickness of characters, and $W^{ij}_b$ about the darkness of characters.

As for $W^{ij}_{st}$, we first calculate the number of black pixels, i.e. pixels covered by the characters. Then $W^{ij}_{st}$ is defined as the reciprocal of the number of black pixels in each target image
\begin{equation}
W^{ij}_{st} = 1/N_b^{ij},
\end{equation}
where $N_b^{ij}$ is the number of black pixels of target image $I_{ij}$.

As for $W^{ij}_b$, we calculate the mean value of black pixels for each target image and set a softmax weight
\begin{equation}
W^{ij}_b = \frac{exp(\textrm{mean}_{ij})}{\sum_{I_{ij}\in \mathcal{D}_t}exp(\textrm{mean}_{ij})},
\end{equation}
where $\textrm{mean}_{ij}$ is the mean value of the black pixels of the target image $I_{ij}$.

\section{Experiments}
In this section, we evaluate the proposed network for Chinese Typeface transfer problem. We first introduce the data set we used followed by the implementation details. Finally, we present our experimental results.

\begin{figure}[!t]
\setlength{\abovecaptionskip}{-2pt}
\centering
\hspace{-5pt}
\includegraphics[height=1.5in,width=3.3in]{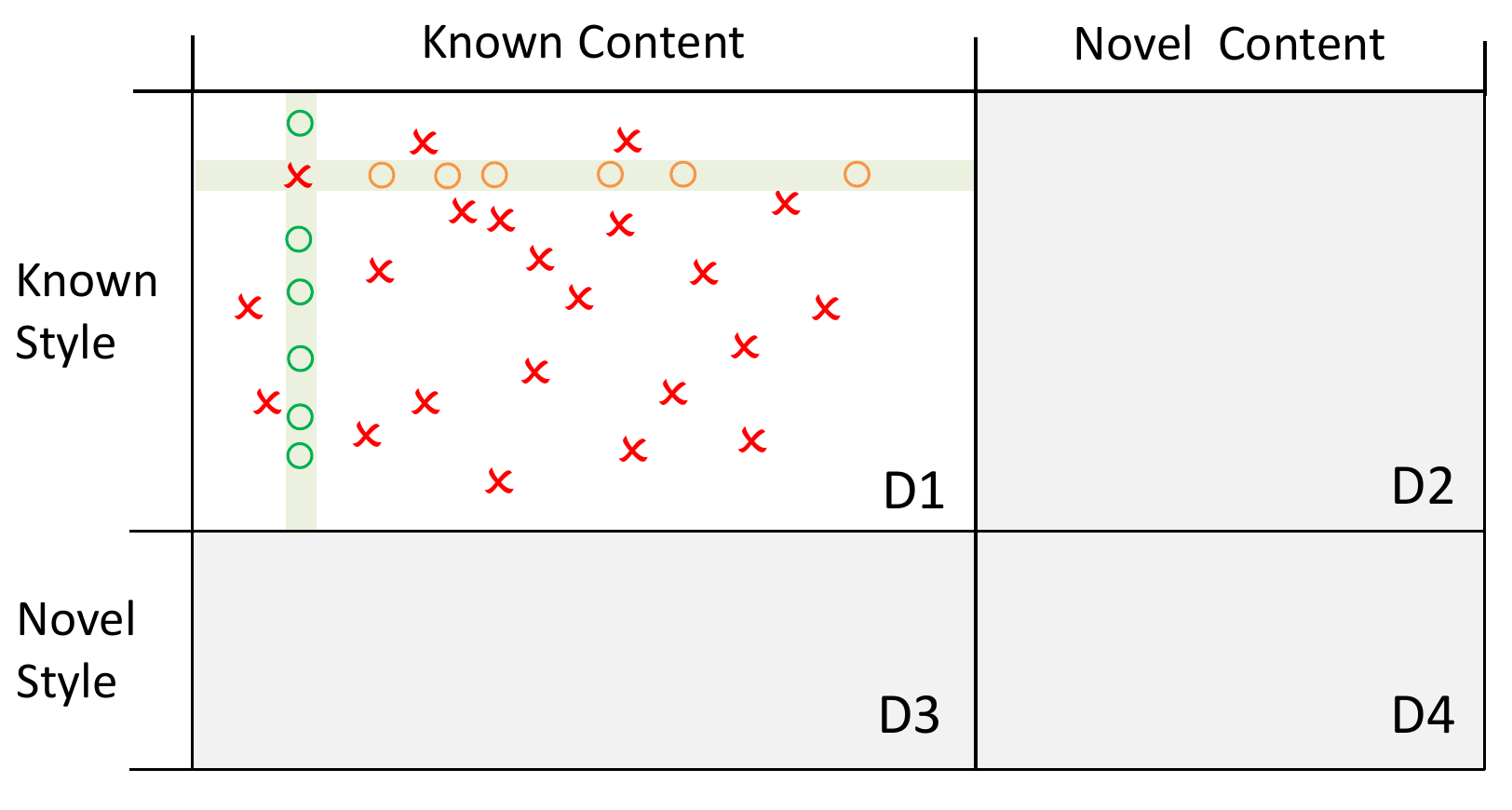}
\caption{The illustration of data set partition, target images selection and reference set construction (best viewed in color).}
\label{fig:train}
\vspace{-10pt}
\end{figure}

\begin{figure}[!t]
\centering
\setlength{\abovecaptionskip}{5pt}
\hspace{-6pt}
\subfigure{
\begin{minipage}{0.24\textwidth}{\vspace{-5pt}
\begin{minipage}{0.12\textwidth}TG:\end{minipage}
\begin{minipage}{0.15\textwidth}
\includegraphics[width=1.5in,height=0.21in]{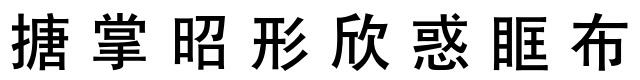}
\end{minipage}\\
\begin{minipage}{0.12\textwidth}O1:\end{minipage}
\begin{minipage}{0.15\textwidth}
\includegraphics[width=1.5in,height=0.21in]{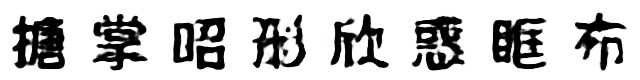}
\end{minipage}\\
\begin{minipage}{0.12\textwidth}O2:\end{minipage}
\begin{minipage}{0.15\textwidth}
\includegraphics[width=1.5in,height=0.21in]{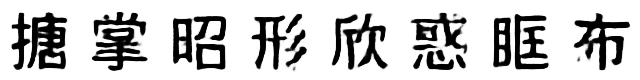}
\end{minipage}\\
\begin{minipage}{0.12\textwidth}O3:\end{minipage}
\begin{minipage}{0.15\textwidth}
\includegraphics[width=1.5in,height=0.21in]{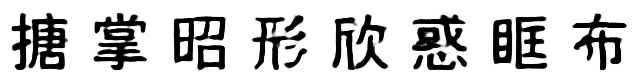}
\end{minipage}\\
\begin{minipage}{0.12\textwidth}O4:\end{minipage}
\begin{minipage}{0.15\textwidth}
\includegraphics[width=1.5in,height=0.21in]{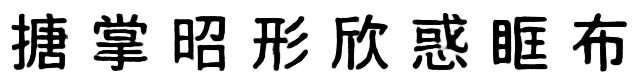}
\end{minipage}\\
\begin{minipage}{0.12\textwidth}O5:\end{minipage}
\begin{minipage}{0.15\textwidth}
\includegraphics[width=1.5in,height=0.21in]{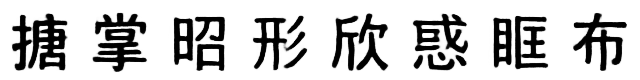}
\end{minipage}
\ruleh
\\
\begin{minipage}{0.12\textwidth}TG:\end{minipage}
\begin{minipage}{0.15\textwidth}
\includegraphics[width=1.5in,height=0.21in]{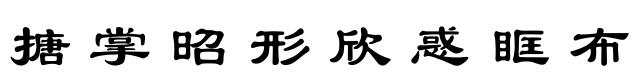}
\end{minipage}\\
\begin{minipage}{0.12\textwidth}O1:\end{minipage}
\begin{minipage}{0.15\textwidth}
\includegraphics[width=1.5in,height=0.21in]{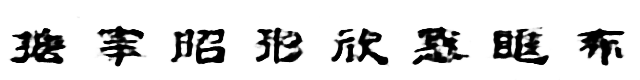}
\end{minipage}\\
\begin{minipage}{0.12\textwidth}O2:\end{minipage}
\begin{minipage}{0.15\textwidth}
\includegraphics[width=1.5in,height=0.21in]{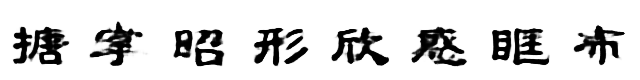}
\end{minipage}\\
\begin{minipage}{0.12\textwidth}O3:\end{minipage}
\begin{minipage}{0.15\textwidth}
\includegraphics[width=1.5in,height=0.21in]{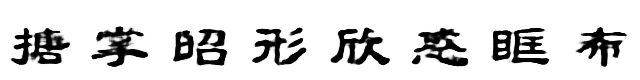}
\end{minipage}\\
\begin{minipage}{0.12\textwidth}O4:\end{minipage}
\begin{minipage}{0.15\textwidth}
\includegraphics[width=1.5in,height=0.21in]{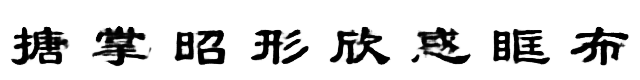}
\end{minipage}\\
\begin{minipage}{0.12\textwidth}O5:\end{minipage}
\begin{minipage}{0.15\textwidth}
\includegraphics[width=1.5in,height=0.21in]{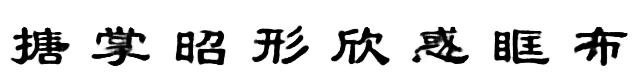}
\end{minipage}
}
\end{minipage}
}
\vspace{-5pt}
\rulev
\hspace{-3pt}
\subfigure{
\begin{minipage}{0.21\textwidth}{\vspace{-5pt}
\includegraphics[width=1.5in,height=0.21in]{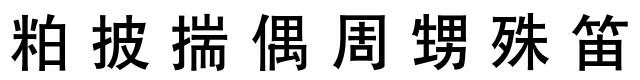}\\
\includegraphics[width=1.5in,height=0.21in]{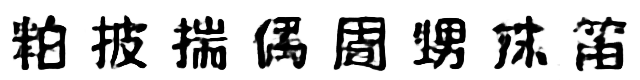}\\
\includegraphics[width=1.5in,height=0.21in]{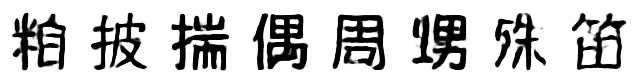}\\
\includegraphics[width=1.5in,height=0.21in]{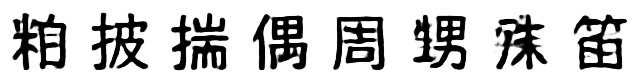}\\
\includegraphics[width=1.5in,height=0.21in]{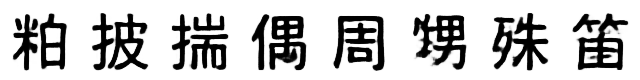}\\
\includegraphics[width=1.5in,height=0.21in]{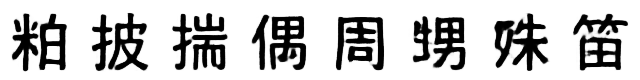}
\ruleh
\\
\includegraphics[width=1.5in,height=0.21in]{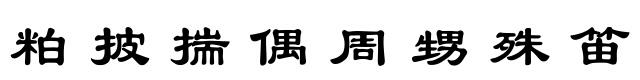}\\
\includegraphics[width=1.5in,height=0.21in]{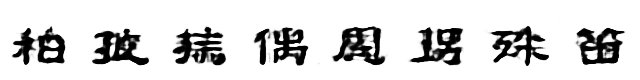}\\
\includegraphics[width=1.5in,height=0.21in]{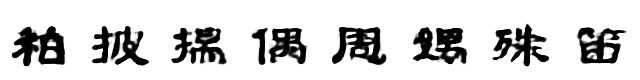}\\
\includegraphics[width=1.5in,height=0.21in]{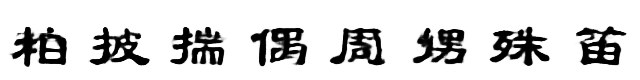}\\
\includegraphics[width=1.5in,height=0.21in]{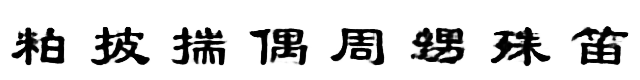}\\
\includegraphics[width=1.5in,height=0.21in]{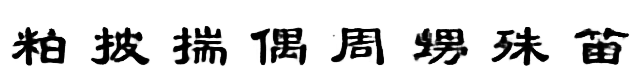}

}
\end{minipage}
}
\caption{Generation results for $D_1$, $D_2$, $D_3$, $D_4$ (from upper left to lower right) with different training set size. TG: Target image, O1: Output for $N_{t}$=20k, O2: Output for $N_{t}$=50k, O3: Output for $N_{t}$=100k, O4: Output for $N_{t}$=300k, O5: Output for $N_{t}$=500k. In all cases, $r$=10.}
\label{fig:train-size}
\vspace{-15pt}
\end{figure}

\subsection{Data Set}
To evaluate the proposed {\em EMD} model with Chinese Typeface transfer tasks, we construct a data set which contains 832 fonts (styles) and each font has 1732 commonly used Chinese characters (contents). All images are $80\times 80$ pixels. We randomly select 75\% of the styles and contents as known styles and contents (i.e. 624 train styles and 1299 train contents) and leave the rest 25\% as novel styles and contents (i.e. 208 novel styles and 433 novel contents). The entire data set is therefor partitioned into four subsets as shown in Figure~\ref{fig:train}: $D_1$, images with known styles and contents namely train styles and contents, $D_2$, images with known styles but novel contents, $D_3$, images with known contents but novel styles, and $D_4$, images with both novel styles and novel contents. The four data sets represent different levels of style transfer challenges.

\subsection{Implementation Details}
In our experiment, the output channels of convolution layers in the $\textsl{Style Encoder}$ and $\textsl{Content Encoder}$ are 1, 2, 4, 8, 8, 8, 8, 8 times of $C$ respectively, where $C$=64. And for the $\textsl{Mixer}$, we set $R$=$B$=$K$ in our implementation. The output channels of the first seven deconvolution layers in $\textsl{Decoder}$ are 8, 8, 8, 8, 4, 2, 1 times of $C$ respectively. We set the initial learning rate as 0.0002 and train the model end-to-end with the Adam optimization method until the output is stable.

In each experiment, we first randomly sample $N_{t}$ target images with known content and known styles as training examples. We then construct the two reference sets for each target image by randomly sampling $r$ images of the corresponding style/content. Figure~\ref{fig:train} provides an illustration of target images selection and reference set construction. Each row represents one style and each column represents a content. The target images are represented by randomly scattered red ``x" marks. The reference images for the target image are selected from corresponding style/content, shown as the orange circles for the style reference images and green circles for content reference images. When testing, taking D4 as an example, each target image in D4 can be generated with $r$ style/content reference images. The style reference images can be randomly sampled from images with target style in D3 and the content reference images are randomly sampled from images with target content in D2.

\begin{figure}[!t]
\centering
\setlength{\abovecaptionskip}{-15pt}
\hspace{-7pt}
\subfigure{
\begin{minipage}{0.24\textwidth}{\vspace{-5pt}
\begin{minipage}{0.12\textwidth}TG:\end{minipage}
\begin{minipage}{0.15\textwidth}
\includegraphics[width=1.5in,height=0.21in]{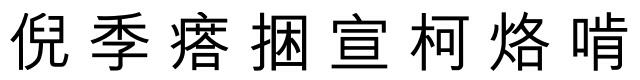}
\end{minipage}\\
\begin{minipage}{0.12\textwidth}O1:\end{minipage}
\begin{minipage}{0.15\textwidth}
\includegraphics[width=1.5in,height=0.21in]{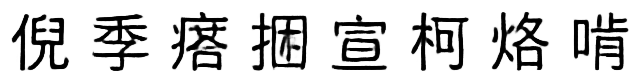}
\end{minipage}\\
\begin{minipage}{0.12\textwidth}O2:\end{minipage}
\begin{minipage}{0.15\textwidth}
\includegraphics[width=1.5in,height=0.21in]{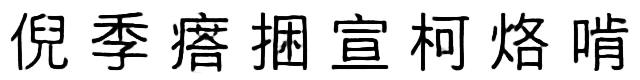}
\end{minipage}\\
\begin{minipage}{0.12\textwidth}O3:\end{minipage}
\begin{minipage}{0.15\textwidth}
\includegraphics[width=1.5in,height=0.21in]{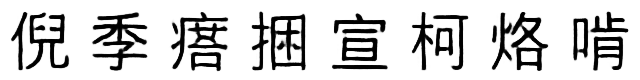}
\end{minipage}
\vspace{3pt}\\
\begin{minipage}{0.12\textwidth}TG:\end{minipage}
\begin{minipage}{0.15\textwidth}
\includegraphics[width=1.5in,height=0.21in]{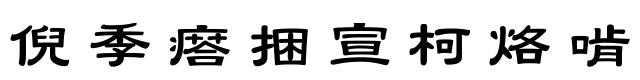}
\end{minipage}\\
\begin{minipage}{0.12\textwidth}O1:\end{minipage}
\begin{minipage}{0.15\textwidth}
\includegraphics[width=1.5in,height=0.21in]{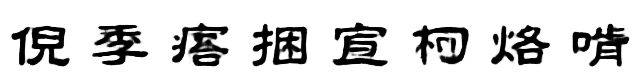}
\end{minipage}\\
\begin{minipage}{0.12\textwidth}O2:\end{minipage}
\begin{minipage}{0.15\textwidth}
\includegraphics[width=1.5in,height=0.21in]{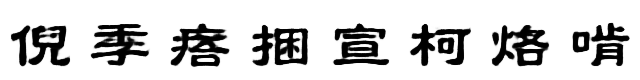}
\end{minipage}\\
\begin{minipage}{0.12\textwidth}O3:\end{minipage}
\begin{minipage}{0.15\textwidth}
\includegraphics[width=1.5in,height=0.21in]{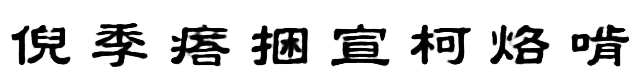}
\end{minipage}
\ruleh
\\
\begin{minipage}{0.12\textwidth}TG:\end{minipage}
\begin{minipage}{0.15\textwidth}
\includegraphics[width=1.5in,height=0.21in]{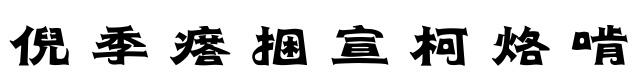}
\end{minipage}\\
\begin{minipage}{0.12\textwidth}O1:\end{minipage}
\begin{minipage}{0.15\textwidth}
\includegraphics[width=1.5in,height=0.21in]{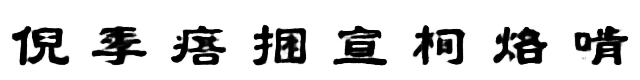}
\end{minipage}\\
\begin{minipage}{0.12\textwidth}O2:\end{minipage}
\begin{minipage}{0.15\textwidth}
\includegraphics[width=1.5in,height=0.21in]{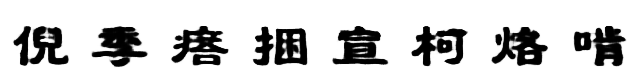}
\end{minipage}\\
\begin{minipage}{0.12\textwidth}O3:\end{minipage}
\begin{minipage}{0.15\textwidth}
\includegraphics[width=1.5in,height=0.21in]{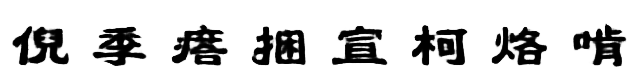}
\end{minipage}
\vspace{3pt}\\
\begin{minipage}{0.12\textwidth}TG:\end{minipage}
\begin{minipage}{0.15\textwidth}
\includegraphics[width=1.5in,height=0.21in]{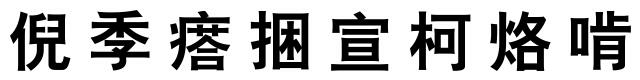}
\end{minipage}\\
\begin{minipage}{0.12\textwidth}O1:\end{minipage}
\begin{minipage}{0.15\textwidth}
\includegraphics[width=1.5in,height=0.21in]{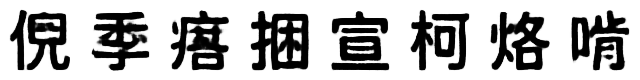}
\end{minipage}\\
\begin{minipage}{0.12\textwidth}O2:\end{minipage}
\begin{minipage}{0.15\textwidth}
\includegraphics[width=1.5in,height=0.21in]{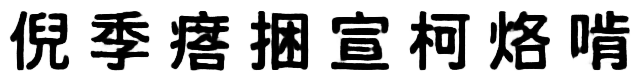}
\end{minipage}\\
\begin{minipage}{0.12\textwidth}O3:\end{minipage}
\begin{minipage}{0.15\textwidth}
\includegraphics[width=1.5in,height=0.21in]{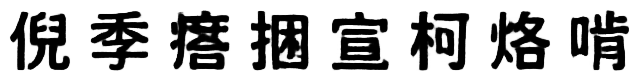}
\end{minipage}
}
\end{minipage}
}
\hspace{1pt}
\rulev
\hspace{-3pt}
\subfigure{
\begin{minipage}{0.21\textwidth}{\vspace{-5pt}
\includegraphics[width=1.5in,height=0.21in]{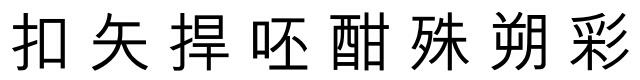}\\
\includegraphics[width=1.5in,height=0.21in]{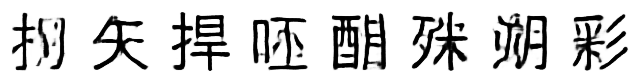}\\
\includegraphics[width=1.5in,height=0.21in]{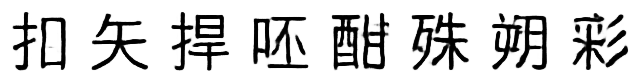}\\
\includegraphics[width=1.5in,height=0.21in]{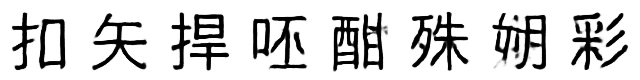}\vspace{3pt}\\
\includegraphics[width=1.5in,height=0.21in]{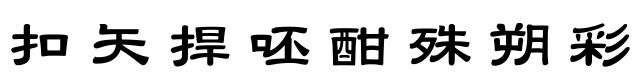}\\
\includegraphics[width=1.5in,height=0.21in]{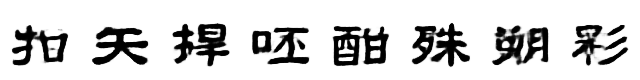}\\
\includegraphics[width=1.5in,height=0.21in]{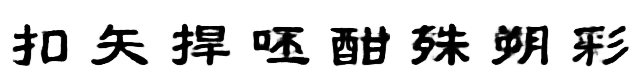}\\
\includegraphics[width=1.5in,height=0.21in]{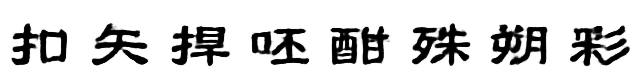}
\ruleh
\\
\includegraphics[width=1.5in,height=0.21in]{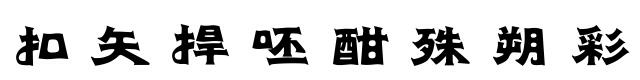}\\
\includegraphics[width=1.5in,height=0.21in]{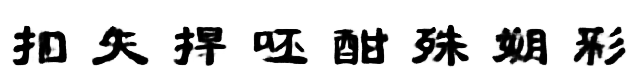}\\
\includegraphics[width=1.5in,height=0.21in]{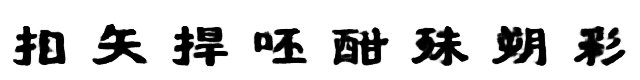}\\
\includegraphics[width=1.5in,height=0.21in]{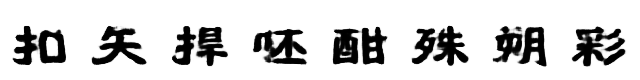}\vspace{3pt}\\
\includegraphics[width=1.5in,height=0.21in]{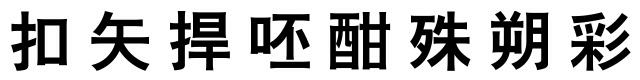}\\
\includegraphics[width=1.5in,height=0.21in]{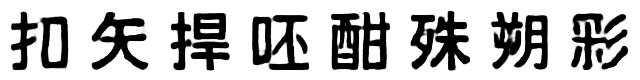}\\
\includegraphics[width=1.5in,height=0.21in]{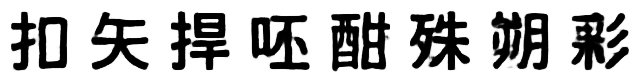}\\
\includegraphics[width=1.5in,height=0.21in]{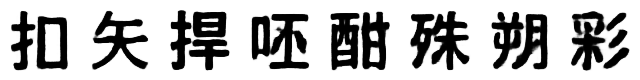}

}
\end{minipage}
}
\caption{The impact of the number of reference images on the generation of images in $D_1$, $D_2$, $D_3$, $D_4$, respectively (from upper left to lower right). TG: Target image, O1: Output for $r$=5, O2: Output for $r$=10, O3: Output for $r$=15. In all cases, $N_{t}$=300k.}
\vspace{-15pt}
\label{fig:ref-num}
\end{figure}

\subsection{Experimental Results}

In this section, we present the experimental results. First, we analyze the influence of some factors influencing the model performance. Then, we validate the separation of style and content. Finally, we compare the proposed method with some baseline networks to prove the effectiveness of our method.

\subsubsection{Influence of the Training Set Size }
To evaluate the influence of the training set size on style transfer, we conduct experiments for $N_{t}$=20k, 50k, 100k, 300k and 500k. The generation results for $D_{1}$, $D_{2}$, $D_{3}$ and $D_{4}$ are shown in Figure~\ref{fig:train-size}. As we can see, the larger the training set, the better the performance, which is consistent with our intuition. The generated images with $N_{t}$=300k and 500k are clearly better than images generated with $N_{t}$=20k, 50k and 100k. Besides, the performance of $N_{t}$=300k and $N_{t}$=500k is close which implies that with more training images, the network performance tends to be saturated and $N_{t}$=300k is enough for good results. Therefore, we take $N_{t}$=300k for the following experiments.

\begin{figure}[!tpb]
\centering
\setlength{\abovecaptionskip}{-10pt}
\hspace{-7pt}
\subfigure{
\begin{minipage}{0.24\textwidth}{\vspace{-5pt}
\begin{minipage}{0.12\textwidth}TG:\end{minipage}
\begin{minipage}{0.15\textwidth}
\includegraphics[width=1.5in,height=0.21in]{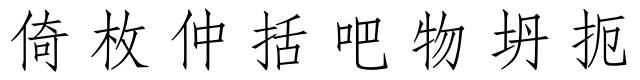}
\end{minipage}\\
\begin{minipage}{0.12\textwidth}O1:\end{minipage}
\begin{minipage}{0.15\textwidth}
\includegraphics[width=1.5in,height=0.21in]{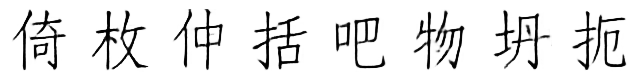}
\end{minipage}\\
\begin{minipage}{0.12\textwidth}O2:\end{minipage}
\begin{minipage}{0.15\textwidth}
\includegraphics[width=1.5in,height=0.21in]{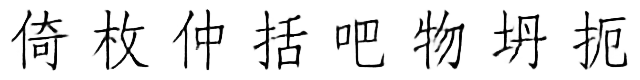}
\end{minipage}
\vspace{3pt}\\
\begin{minipage}{0.12\textwidth}TG:\end{minipage}
\begin{minipage}{0.15\textwidth}
\includegraphics[width=1.5in,height=0.21in]{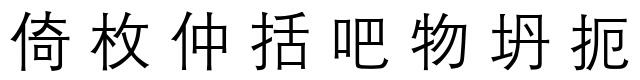}
\end{minipage}\\
\begin{minipage}{0.12\textwidth}O1:\end{minipage}
\begin{minipage}{0.15\textwidth}
\includegraphics[width=1.5in,height=0.21in]{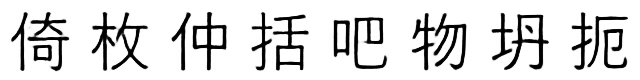}
\end{minipage}\\
\begin{minipage}{0.12\textwidth}O2:\end{minipage}
\begin{minipage}{0.15\textwidth}
\includegraphics[width=1.5in,height=0.21in]{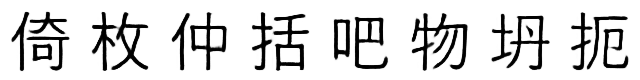}
\end{minipage}
\ruleh
\\
\begin{minipage}{0.12\textwidth}TG:\end{minipage}
\begin{minipage}{0.15\textwidth}
\includegraphics[width=1.5in,height=0.21in]{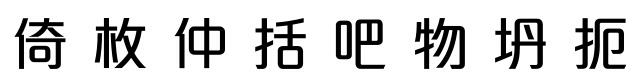}
\end{minipage}\\
\begin{minipage}{0.12\textwidth}O1:\end{minipage}
\begin{minipage}{0.15\textwidth}
\includegraphics[width=1.5in,height=0.21in]{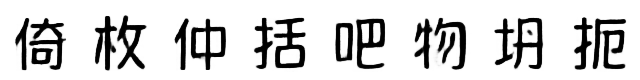}
\end{minipage}\\
\begin{minipage}{0.12\textwidth}O2:\end{minipage}
\begin{minipage}{0.15\textwidth}
\includegraphics[width=1.5in,height=0.21in]{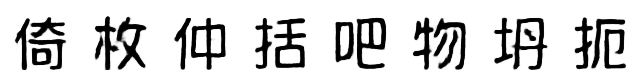}
\end{minipage}
\vspace{3pt}\\
\begin{minipage}{0.12\textwidth}TG:\end{minipage}
\begin{minipage}{0.15\textwidth}
\includegraphics[width=1.5in,height=0.21in]{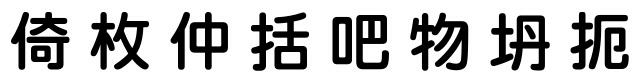}
\end{minipage}\\
\begin{minipage}{0.12\textwidth}O1:\end{minipage}
\begin{minipage}{0.15\textwidth}
\includegraphics[width=1.5in,height=0.21in]{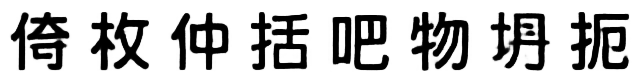}
\end{minipage}\\
\begin{minipage}{0.12\textwidth}O2:\end{minipage}
\begin{minipage}{0.15\textwidth}
\includegraphics[width=1.5in,height=0.21in]{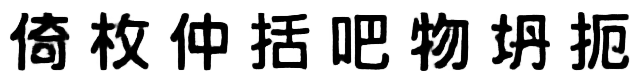}
\end{minipage}
}
\end{minipage}
}
\hspace{1pt}
\rulev
\hspace{-3pt}
\subfigure{
\begin{minipage}{0.21\textwidth}{\vspace{-5pt}
\includegraphics[width=1.5in,height=0.21in]{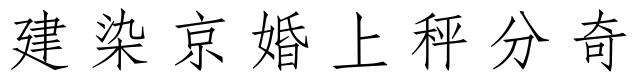}\\
\includegraphics[width=1.5in,height=0.21in]{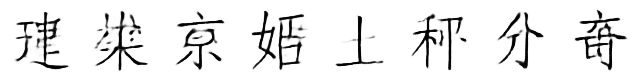}\\
\includegraphics[width=1.5in,height=0.21in]{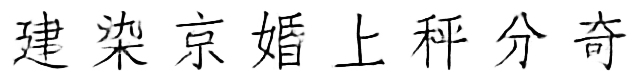}\vspace{3pt}\\
\includegraphics[width=1.5in,height=0.21in]{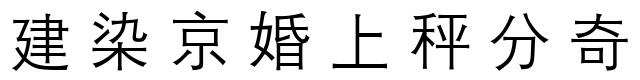}\\
\includegraphics[width=1.5in,height=0.21in]{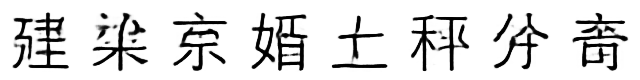}\\
\includegraphics[width=1.5in,height=0.21in]{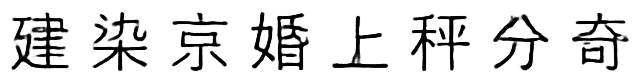}
\ruleh
\vspace{2pt}
\includegraphics[width=1.5in,height=0.21in]{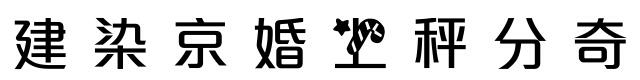}\\
\includegraphics[width=1.5in,height=0.21in]{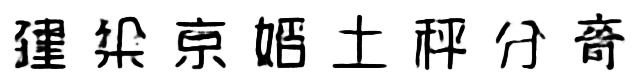}\\
\includegraphics[width=1.5in,height=0.21in]{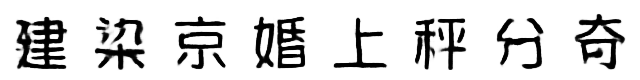}\vspace{3pt}\\
\includegraphics[width=1.5in,height=0.21in]{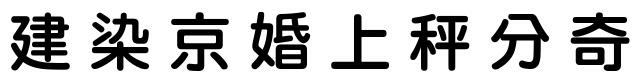}\\
\includegraphics[width=1.5in,height=0.21in]{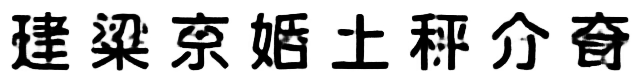}\\
\includegraphics[width=1.5in,height=0.21in]{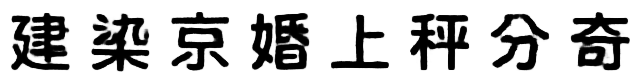}

}
\end{minipage}
}
\caption{The impact of the skip-connection on generation of images in $D_1$, $D_2$, $D_3$, $D_4$, respectively (from upper left to lower right). TG is the target image, O1 and O2 are outputs of models without and with skip-connection. In all cases $N_{t}$=300k, $r$=10.}
\vspace{-18pt}
\label{fig:connection}
\end{figure}

\subsubsection{Influence of the Reference Set Size}
In addition, we conduct experiments with different number of reference images. Figure~\ref{fig:ref-num} displays the image generation results of $N_{t}$=300k with $r$=5, $r$=10 and $r$=15 respectively. From the figure, we can observe that with more reference images, characters are generated better in details. Besides, characters generated with $r$=5 are overall okay, meaning that our model can generalize to novel styles using only a few reference images. The generation results of $r$=10 and $r$=15 are close, therefore we take $r$=10 in our other experiments. Intuitively, more reference images will support more information about strokes and styles of characters and the common points in the reference sets will be more obvious. Therefore, given $r>1$, our model can achieve co-learning of images with the same style/content. Moreover, with $r>1$ we can learn more images at once which will improve the efficiency but if we split the $<$r, r, 1$>$ triplets to be $r^2$ $<$1, 1, 1$>$ triplets, the time will increase nearly $r^2$ times under the same condition.

\subsubsection{Effect of the Skip-connection}

To evaluate the effectiveness of the skip-connection during image generation, we compare the results with and without skip-connection in Figure~\ref{fig:connection}. As shown in the figure, images in $D_1$ are generated best, next is $D_3$ and last is $D_2$ and $D_4$, which conforms to the difficulty level and indicates that novel contents are more challenging to extract than novel styles. For known contents, models with and without skip-connection perform closely but for novel contents, images generated with skip-connection are much better in details. Besides, the model without skip-connection may generate images of novel characters to be similar characters which it has seen before. This is because the structure of novel characters is more challenging to extract and the structure information losing during down-sampling will lead the model to generate blurry even wrong characters. However, with content skip-connection, the location and structure information lost will be recaptured by the $\textsl{Decoder}$ network.

\begin{figure}[!t]
\centering
\setlength{\abovecaptionskip}{-10pt}
\includegraphics[height=1.7in,width=3.3in]{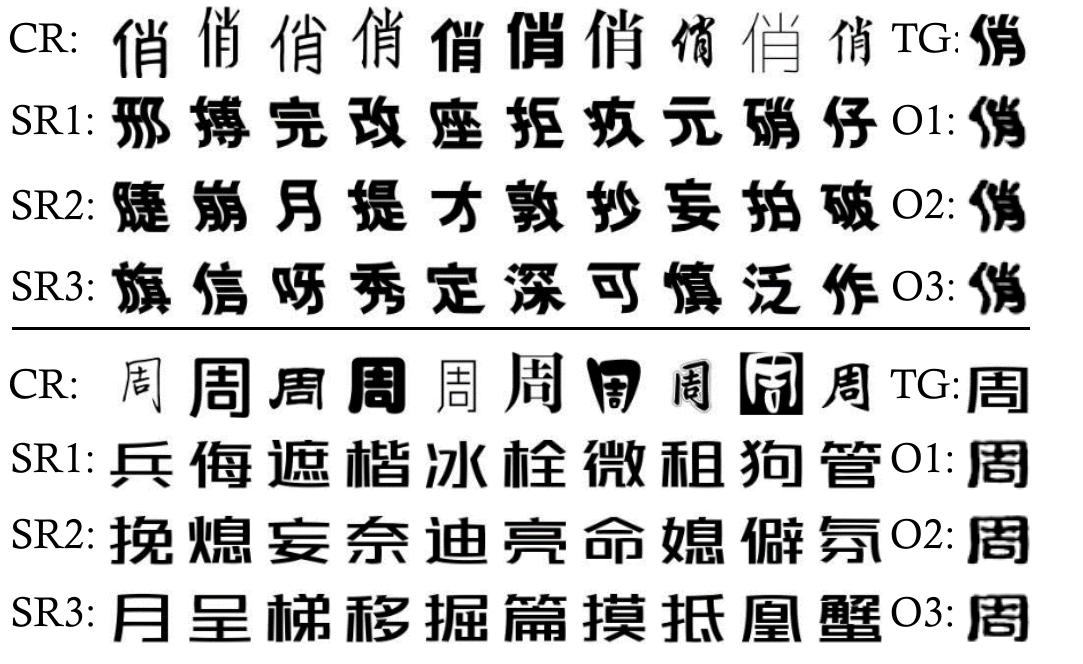}
\caption{Validation of pure style extraction. CR: the content reference set, TG: the target image, O1, O2 and O3 are generated by CR and three different style reference sets SR1, SR2 and SR3.}
\vspace{-10pt}
\label{fig:vali-style}
\end{figure}


\begin{figure}[!t]
\centering
\setlength{\abovecaptionskip}{-2pt}
\includegraphics[height=1.7in,width=3.3in]{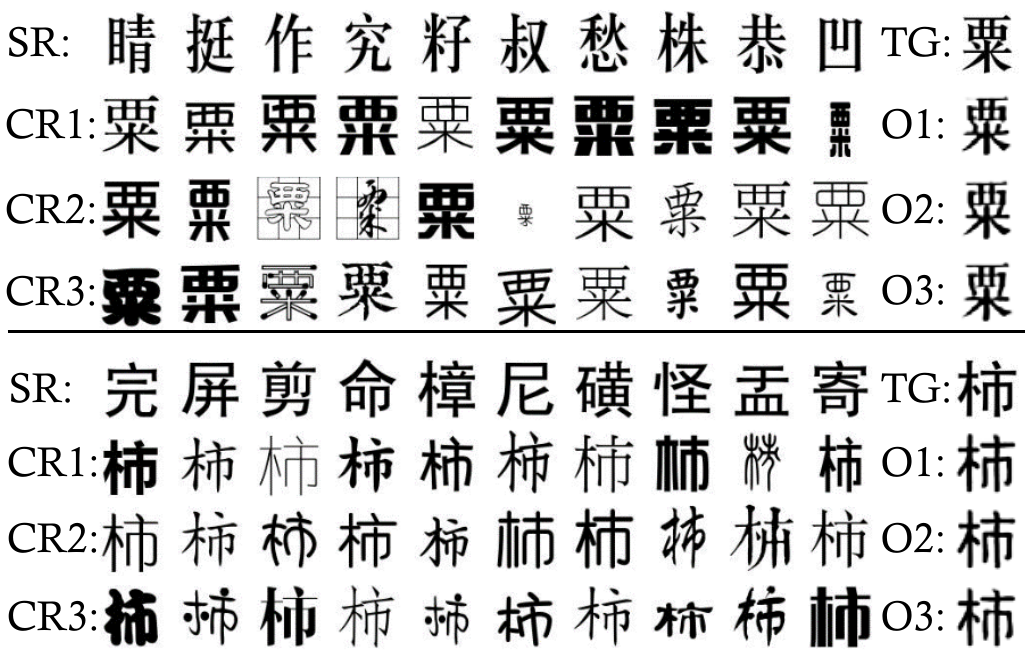}
\caption{Validation of pure content extraction. SR: the style reference set, TG: the target image, O1, O2 and O3 are generated using SR but three different content reference sets CR1, CR2 and CR3.}
\vspace{-15pt}
\label{fig:vali-content}
\end{figure}

\begin{figure*}[!t]
\centering
\setlength{\abovecaptionskip}{-15pt}
\hspace{-20pt}
\subfigure{
\begin{minipage}{0.42\textwidth}{\vspace{-10pt}
\begin{minipage}{0.17\textwidth}Source:\end{minipage}
\begin{minipage}{0.38\textwidth}
\includegraphics[width=2.4in,height=0.25in]{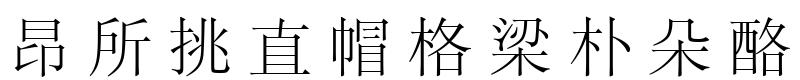}
\end{minipage}\\
\begin{minipage}{0.17\textwidth}Pix2pix:\end{minipage}
\begin{minipage}{0.38\textwidth}
\includegraphics[width=2.4in,height=0.25in]{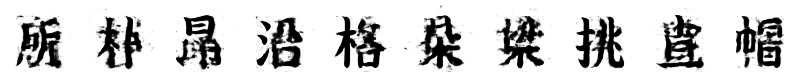}
\end{minipage}\\
\begin{minipage}{0.17\textwidth}AEGN:\end{minipage}
\begin{minipage}{0.38\textwidth}
\includegraphics[width=2.4in,height=0.25in]{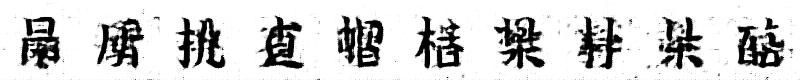}
\end{minipage}\\
\begin{minipage}{0.17\textwidth}Zitozi:\end{minipage}
\begin{minipage}{0.38\textwidth}
\includegraphics[width=2.4in,height=0.25in]{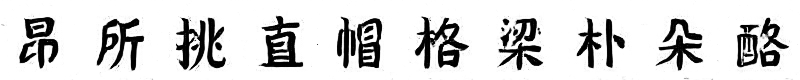}
\end{minipage}\\
\begin{minipage}{0.17\textwidth}C-GAN:\end{minipage}
\begin{minipage}{0.38\textwidth}
\includegraphics[width=2.4in,height=0.25in]{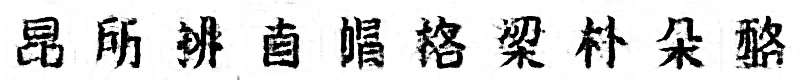}
\end{minipage}\\
\begin{minipage}{0.17\textwidth}EMD:\end{minipage}
\begin{minipage}{0.38\textwidth}
\includegraphics[width=2.4in,height=0.25in]{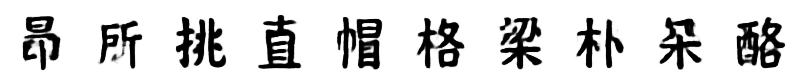}
\end{minipage}\\
\begin{minipage}{0.17\textwidth}Target:\end{minipage}
\begin{minipage}{0.38\textwidth}
\includegraphics[width=2.4in,height=0.25in]{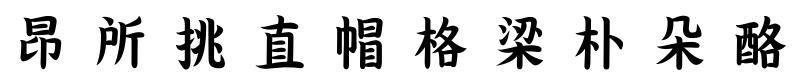}
\end{minipage}
}
\end{minipage}
\rulev
\begin{minipage}{0.35\textwidth}{\vspace{-5pt}
\begin{minipage}{0.38\textwidth}
\includegraphics[width=2.4in,height=0.25in]{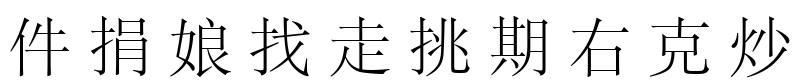}
\end{minipage}\\
\begin{minipage}{0.38\textwidth}
\includegraphics[width=2.4in,height=0.25in]{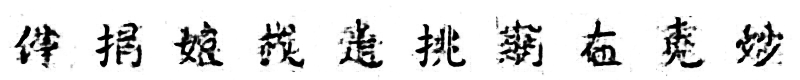}
\end{minipage}\\
\begin{minipage}{0.38\textwidth}
\includegraphics[width=2.4in,height=0.25in]{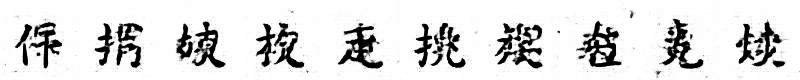}
\end{minipage}\\
\begin{minipage}{0.44\textwidth}
\includegraphics[width=2.4in,height=0.25in]{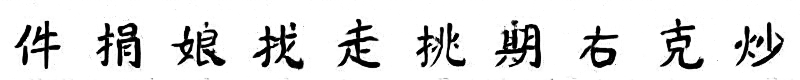}
\end{minipage}\\
\begin{minipage}{0.44\textwidth}
\includegraphics[width=2.4in,height=0.25in]{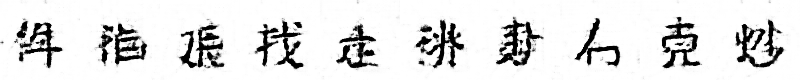}
\end{minipage}\\
\begin{minipage}{0.35\textwidth}
\includegraphics[width=2.4in,height=0.25in]{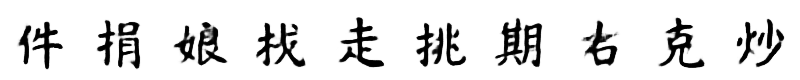}
\end{minipage}\\
\begin{minipage}{0.35\textwidth}
\includegraphics[width=2.4in,height=0.25in]{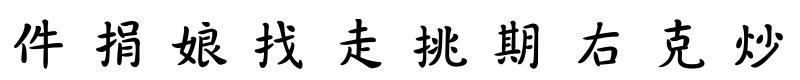}
\end{minipage}\\
}
\end{minipage}
\rulev
\begin{minipage}{0.45\textwidth}{\vspace{-15pt}
\begin{minipage}{0.15\textwidth}L1 loss\end{minipage}
\begin{minipage}{0.15\textwidth}RMSE\end{minipage}
\begin{minipage}{0.15\textwidth}PDAR\end{minipage}\vspace{7pt}\\
\begin{minipage}{0.15\textwidth}0.0105\end{minipage}
\begin{minipage}{0.15\textwidth}0.0202\end{minipage}
\begin{minipage}{0.15\textwidth}0.17\end{minipage}\vspace{7pt}\\
\begin{minipage}{0.15\textwidth}0.0112\end{minipage}
\begin{minipage}{0.15\textwidth}0.0202\end{minipage}
\begin{minipage}{0.15\textwidth}0.3001\end{minipage}\vspace{7pt}\\
\begin{minipage}{0.15\textwidth}0.0091\end{minipage}
\begin{minipage}{0.15\textwidth}\textbf{0.0184}\end{minipage}
\begin{minipage}{0.15\textwidth}0.1659\end{minipage}\vspace{7pt}\\
\begin{minipage}{0.15\textwidth}0.0112\end{minipage}
\begin{minipage}{0.15\textwidth}0.02\end{minipage}
\begin{minipage}{0.15\textwidth}0.3685\end{minipage}\vspace{7pt}\\
\begin{minipage}{0.15\textwidth}\textbf{0.0087}\end{minipage}
\begin{minipage}{0.15\textwidth}\textbf{0.0184}\end{minipage}
\begin{minipage}{0.15\textwidth}\textbf{0.1332}\end{minipage}\\
}
\end{minipage}
}
\vspace{-10pt}
\caption{Comparison of image generation for known styles and novel contents. Equal number of image pairs with source and target styles are used to train the baselines.}
\label{fig:com-area2}
\vspace{-15pt}
\end{figure*}

\subsubsection{Validation of Style and Content Separation}

Separating style and content is the key feature of the proposed {\em EMD} model. To validate the clear separation of style and content, we combine one content representation with style representations from a few disjoint style reference sets for one style and check whether the generated images are the same. For better validation, the content reference sets and style reference sets are all for novel styles and contents and we generate images with novel style and novel content. Similarly, we combine one style representation with content representations from a few disjoint content reference sets. The results are displayed in Figure~\ref{fig:vali-style} and Figure~\ref{fig:vali-content}, respectively. As shown in Figure~\ref{fig:vali-style}, the generated O1, O2 and O3 are similar though the style reference sets used are different, demonstrating that the $\textsl{Style Encoder}$ extracts accurate style representations since the only one thing the three style reference sets share is the style. Similar results can be found in Figure~\ref{fig:vali-content}, showing that the $\textsl{Content Encoder}$ extracts accurate content representations.

\subsubsection{Comparison with Baseline Methods}
In this subsection, we compare our method with the following baselines for character style transfer.
\begin{itemize}
    \item Pix2pix~\cite{isola}: Pix2pix is a conditional GAN based image translation network, which also adopts the skip-connection to connect encoder and decoder. Pix2pix is optimized by L1 distance loss and adversarial loss.
 \item Auto-encoder guided GAN (AEGN)~\cite{lyu}: AEGN consists of two encoder-decoder networks, one for image transfer and another acting as an auto-encoder to guide the transfer to learn detailed stroke information.
\item Zi-to-zi~\cite{zitozi}: Zi-to-zi is proposed for Chinese typeface transfer which is based on the encoder-decoder architecture followed by a discriminator. In discriminator, there are two fully connected layers to predict the real/fake and the style category respectively.
\item CycleGAN (C-GAN)~\cite{Zhu2017}: CycleGAN consists of two mapping networks which translate images from style A to B and from style B to A, respectively and construct a cycle process. 
\end{itemize}

\begin{figure*}[!t]
\centering
\setlength{\abovecaptionskip}{-10pt}
\hspace{-20pt}
\subfigure{
\begin{minipage}{0.45\textwidth}{
\begin{minipage}{0.26\textwidth}Source:\end{minipage}
\begin{minipage}{0.44\textwidth}
\includegraphics[width=2.3in,height=0.25in]{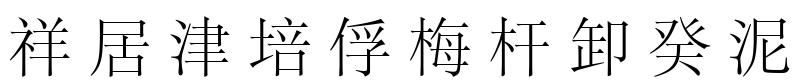}
\end{minipage}\\
\begin{minipage}{0.26\textwidth}Pix2pix-300:\end{minipage}
\begin{minipage}{0.44\textwidth}
\includegraphics[width=2.3in,height=0.25in]{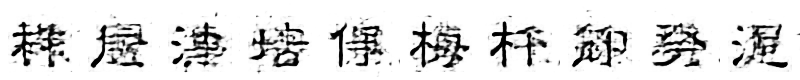}
\end{minipage}\\
\begin{minipage}{0.26\textwidth}Pix2pix-500:\end{minipage}
\begin{minipage}{0.44\textwidth}
\includegraphics[width=2.3in,height=0.25in]{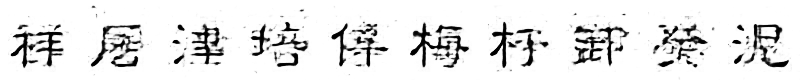}
\end{minipage}\\
\begin{minipage}{0.26\textwidth}Pix2pix-1299:\end{minipage}
\begin{minipage}{0.44\textwidth}
\includegraphics[width=2.3in,height=0.25in]{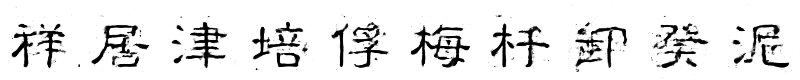}
\end{minipage}\\
\begin{minipage}{0.26\textwidth}AEGN-300:\end{minipage}
\begin{minipage}{0.44\textwidth}
\includegraphics[width=2.3in,height=0.25in]{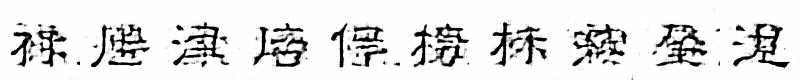}
\end{minipage}\\
\begin{minipage}{0.26\textwidth}AEGN-500:\end{minipage}
\begin{minipage}{0.44\textwidth}
\includegraphics[width=2.3in,height=0.25in]{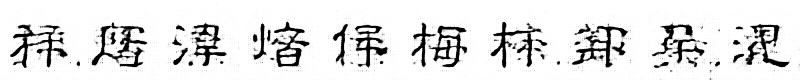}
\end{minipage}\\
\begin{minipage}{0.26\textwidth}AEGN-1299:\end{minipage}
\begin{minipage}{0.44\textwidth}
\includegraphics[width=2.3in,height=0.25in]{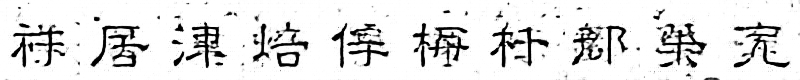}
\end{minipage}\\
\begin{minipage}{0.26\textwidth}Zitozi-300:\end{minipage}
\begin{minipage}{0.44\textwidth}
\includegraphics[width=2.3in,height=0.25in]{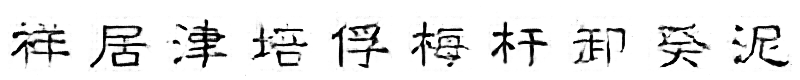}
\end{minipage}\\
\begin{minipage}{0.26\textwidth}Zitozi-500:\end{minipage}
\begin{minipage}{0.44\textwidth}
\includegraphics[width=2.3in,height=0.25in]{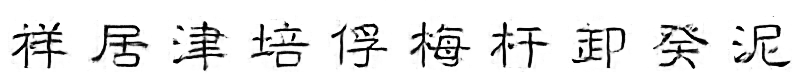}
\end{minipage}\\
\begin{minipage}{0.26\textwidth}Zitozi-1299:\end{minipage}
\begin{minipage}{0.44\textwidth}
\includegraphics[width=2.3in,height=0.25in]{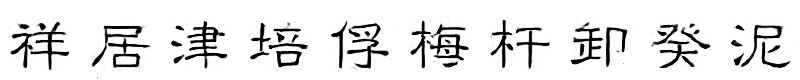}
\end{minipage}\\
\begin{minipage}{0.26\textwidth}C-GAN-300:\end{minipage}
\begin{minipage}{0.44\textwidth}
\includegraphics[width=2.3in,height=0.25in]{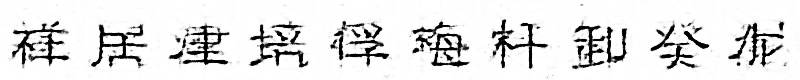}
\end{minipage}\\
\begin{minipage}{0.26\textwidth}C-GAN-500:\end{minipage}
\begin{minipage}{0.44\textwidth}
\includegraphics[width=2.3in,height=0.25in]{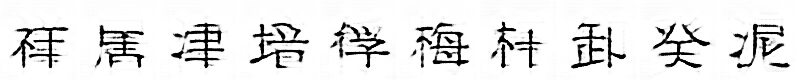}
\end{minipage}\\
\begin{minipage}{0.26\textwidth}C-GAN-1299:\end{minipage}
\begin{minipage}{0.44\textwidth}
\includegraphics[width=2.3in,height=0.25in]{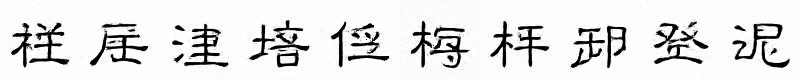}
\end{minipage}\\
\begin{minipage}{0.26\textwidth}EMD-10:\end{minipage}
\begin{minipage}{0.44\textwidth}
\includegraphics[width=2.3in,height=0.25in]{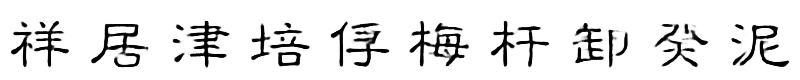}
\end{minipage}\\
\begin{minipage}{0.26\textwidth}Target:\end{minipage}
\begin{minipage}{0.44\textwidth}
\includegraphics[width=2.3in,height=0.25in]{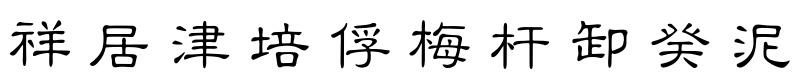}
\end{minipage}\\
}
\end{minipage}
\rulev
\begin{minipage}{0.33\textwidth}{\vspace{-5pt}
\begin{minipage}{0.44\textwidth}
\includegraphics[width=2.3in,height=0.25in]{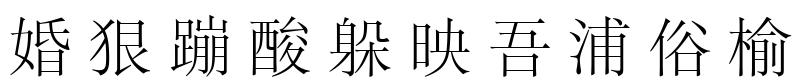}
\end{minipage}\\
\begin{minipage}{0.44\textwidth}
\includegraphics[width=2.3in,height=0.25in]{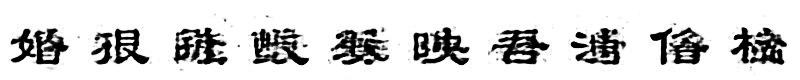}
\end{minipage}\\
\begin{minipage}{0.44\textwidth}
\includegraphics[width=2.3in,height=0.25in]{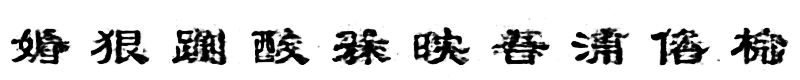}
\end{minipage}\\
\begin{minipage}{0.44\textwidth}
\includegraphics[width=2.3in,height=0.25in]{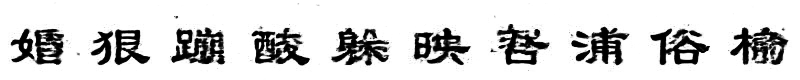}
\end{minipage}\\
\begin{minipage}{0.44\textwidth}
\includegraphics[width=2.3in,height=0.25in]{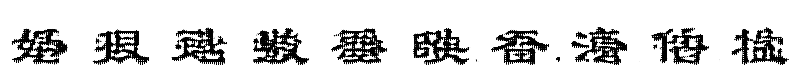}
\end{minipage}\\
\begin{minipage}{0.44\textwidth}
\includegraphics[width=2.3in,height=0.25in]{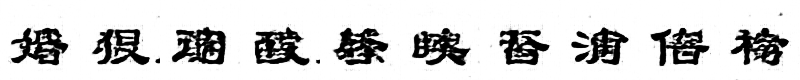}
\end{minipage}\\
\begin{minipage}{0.44\textwidth}
\includegraphics[width=2.3in,height=0.25in]{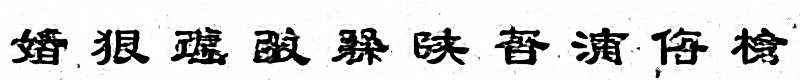}
\end{minipage}\\
\begin{minipage}{0.44\textwidth}
\includegraphics[width=2.3in,height=0.25in]{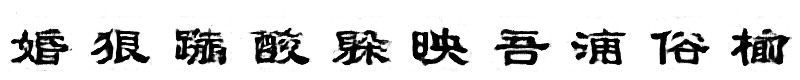}
\end{minipage}\\
\begin{minipage}{0.44\textwidth}
\includegraphics[width=2.3in,height=0.25in]{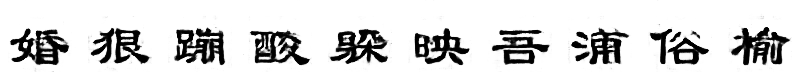}
\end{minipage}\\
\begin{minipage}{0.44\textwidth}
\includegraphics[width=2.3in,height=0.25in]{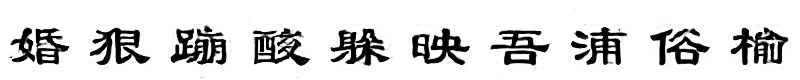}
\end{minipage}\\
\begin{minipage}{0.44\textwidth}
\includegraphics[width=2.3in,height=0.25in]{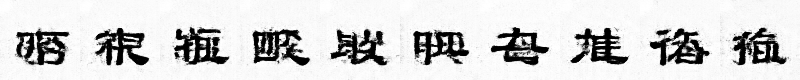}
\end{minipage}\\
\begin{minipage}{0.44\textwidth}
\includegraphics[width=2.3in,height=0.25in]{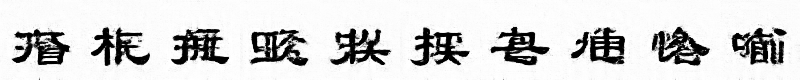}
\end{minipage}\\
\begin{minipage}{0.44\textwidth}
\includegraphics[width=2.3in,height=0.25in]{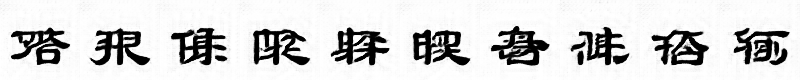}
\end{minipage}\\
\begin{minipage}{0.44\textwidth}
\includegraphics[width=2.3in,height=0.25in]{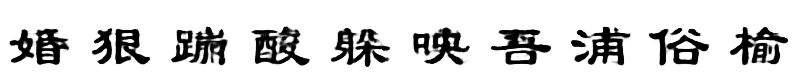}
\end{minipage}\\
\begin{minipage}{0.44\textwidth}
\includegraphics[width=2.3in,height=0.25in]{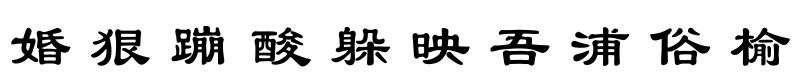}
\end{minipage}
}
\end{minipage}
\rulev
\begin{minipage}{0.40\textwidth}{\vspace{-22pt}
\begin{minipage}{0.15\textwidth}L1 loss\end{minipage}
\begin{minipage}{0.15\textwidth}RMSE\end{minipage}
\begin{minipage}{0.15\textwidth}PDAR\end{minipage}\vspace{7pt}\\
\begin{minipage}{0.15\textwidth}0.0109\end{minipage}
\begin{minipage}{0.15\textwidth}0.0206\end{minipage}
\begin{minipage}{0.15\textwidth}0.1798\end{minipage}\vspace{7pt}\\
\begin{minipage}{0.15\textwidth}0.0106\end{minipage}
\begin{minipage}{0.15\textwidth}0.0202\end{minipage}
\begin{minipage}{0.15\textwidth}0.1765\end{minipage}\vspace{7pt}\\
\begin{minipage}{0.15\textwidth}0.01\end{minipage}
\begin{minipage}{0.15\textwidth}0.0196\end{minipage}
\begin{minipage}{0.15\textwidth}0.1531\end{minipage}\vspace{7pt}\\
\begin{minipage}{0.15\textwidth}0.0117\end{minipage}
\begin{minipage}{0.15\textwidth}0.02\end{minipage}
\begin{minipage}{0.15\textwidth}0.3951\end{minipage}\vspace{7pt}\\
\begin{minipage}{0.15\textwidth}0.0108\end{minipage}
\begin{minipage}{0.15\textwidth}0.02\end{minipage}
\begin{minipage}{0.15\textwidth}0.2727\end{minipage}\vspace{7pt}\\
\begin{minipage}{0.15\textwidth}0.0105\end{minipage}
\begin{minipage}{0.15\textwidth}0.0196\end{minipage}
\begin{minipage}{0.15\textwidth}0.26\end{minipage}\vspace{7pt}\\
\begin{minipage}{0.15\textwidth}0.0091\end{minipage}
\begin{minipage}{0.15\textwidth}0.0187\end{minipage}
\begin{minipage}{0.15\textwidth}0.1612\end{minipage}\vspace{7pt}\\
\begin{minipage}{0.15\textwidth}\textbf{0.009}\end{minipage}
\begin{minipage}{0.15\textwidth}0.0185\end{minipage}
\begin{minipage}{0.15\textwidth}0.1599\end{minipage}\vspace{7pt}\\
\begin{minipage}{0.15\textwidth}\textbf{0.009}\end{minipage}
\begin{minipage}{0.15\textwidth}\textbf{0.0183}\end{minipage}
\begin{minipage}{0.15\textwidth}0.1624\end{minipage}\vspace{7pt}\\
\begin{minipage}{0.15\textwidth}0.0143\end{minipage}
\begin{minipage}{0.15\textwidth}0.0215\end{minipage}
\begin{minipage}{0.15\textwidth}0.5479\end{minipage}\vspace{7pt}\\
\begin{minipage}{0.15\textwidth}0.0126\end{minipage}
\begin{minipage}{0.15\textwidth}0.0203\end{minipage}
\begin{minipage}{0.15\textwidth}0.4925\end{minipage}\vspace{7pt}\\
\begin{minipage}{0.15\textwidth}0.0128\end{minipage}
\begin{minipage}{0.15\textwidth}0.0203\end{minipage}
\begin{minipage}{0.15\textwidth}0.4885\end{minipage}\vspace{7pt}\\
\begin{minipage}{0.15\textwidth}\textbf{0.009}\end{minipage}
\begin{minipage}{0.15\textwidth}0.0186\end{minipage}
\begin{minipage}{0.15\textwidth}\textbf{0.1389}\end{minipage}
}

\end{minipage}
}
\vspace{-10pt}
\caption{Comparison of image generation for novel styles and contents given $r$=10. The baseline methods are trained with 300, 500, 1299 image pairs respectively.}
\vspace{-15pt}
\label{fig:com-area4}
\end{figure*}

For comparison, we use the font Song as the source font which is simple and commonly used and transfer it to target fonts. Our model is trained with $N_{t}$=300k and $r$=10 and as an average, we use less than 500 images for each style. We compare our method with baselines on generating images with known styles and novel styles, respectively. For novel style, the baselines is re-trained from scratch.

\textbf{Known styles as target style.} Taking known styles as the target style, baselines are trained using the same number of paired images as the images our model used for the target style. The results are displayed in Figure~\ref{fig:com-area2} where CycleGAN is denoted as C-GAN for simplicity. We can observe that for known styles and novel contents, our method performs much better than pix2pix, AEGN and CycleGAN and close to or a little better than zi-to-zi. This is because pix2pix and AEGN usually need more samples to learn a style as Lyu did in~\cite{lyu}. CycleGAN performs poorly and it only generates part of characters or some strokes, which may be because it learns the domain mappings and without the domain knowledge, it may perform poorly. Zitozi performs well since it learns multiple styles at the same time and the contrast among different styles helps the model learn styles better.

For quantification analysis, we calculate the L1 loss, Root Mean Square Error (RMSE) and the Pixel Disagreement Ratio (PDAR)~\cite{Zhu2017} between generated images and target images. PDAR is the number of pixels with different values in the two images divided by the total image size after image binaryzation. We conduct experiments for 10 randomly sampled styles and the average results are displayed at the last three columns in Figure~\ref{fig:com-area2} and the best performance is bold. We can observe that our method performs best and achieves the lowest L1 loss, RMSE and PDAR.

\textbf{Novel styles as target style.} Taking novel styles as the target style, we test our model to generate images of novel styles and contents given $r$=10 style/content reference images without retraining. As for baselines, retraining is needed. Here, we conduct two experiments for baselines. One is that we first pretrain a model for each baseline method using the training set our method used and then fine-tune the pretrained model with the same 10 reference images as our method used. The results show that all baseline methods preform poorly and it is unfeasible to learn a style by fine-tuning on only 10 reference images. Thus, we omit the experiment results here.

The other setting is training the baseline model from scratch. Since it is unrealistic to train baseline models with only 10 samples, we train them using 300, 500, 1299 images of the target style respectively. Here we use 1299 is because the number of train contents is 1299 in our data set. The results are presented in Figure~\ref{fig:com-area4}. As shown in the figure, the proposed {\em EMD} model can generalize to novel styles from only 10 style reference images but other methods need to be retrained with more samples. The pix2pix, AEGN and CycleGAN perform worst even learned on all 1299 training images, which demonstrates that these three methods are not effective for character style transfer especially when the training data are not enough. With only 10 style reference images, our model performs better than zi-to-zi-300 namely zi-to-zi model learned with 300 examples for each style, close to zi-to-zi-500 and a little worse than zi-to-zi-1299. This may be because zi-to-zi learns multiple styles at the same time and learning with style contrast helps model learning better.

The quantitative comparison results including L1 loss, RMSE and PDAR are shown at the last three columns of Figure~\ref{fig:com-area4} and we can observe that though given only 10 style reference images, our method performs better than all pix2pix, AEGN and CycleGAN models and zi-to-zi-300, and close to zi-to-zi-500 and zi-to-zi-1299, which demonstrates the effectiveness of our method.

In conclusion, these baseline methods require many images of source styles and target styles to learn, which may hard to collect for some styles. Besides, the learned baseline model can only transfer styles appearing in train set and for new styles, they have to be retrained which is time-consuming. But our method can generalize to novel styles given only a few reference images. In addition, baseline models can only use images of target styles. However, since the proposed {\em EMD} model learns feature representations instead of transformation among specific styles, it can leverage images of any styles and make the most of existing data.

\section{Conclusion and Future Work}
In this paper, we propose a generalized style transfer network named {\em EMD} which could generate images with new styles and contents given only a few style and content reference images. The main idea is that from these reference images, the $\textsl{Style Encoder}$ and $\textsl{Content Encoder}$ could extract style and content representations, respectively. Then the extracted style and content representations will be mixed by a $\textsl{Mixer}$ to generate images with target styles and contents. To separate style and content, we leverage the conditional dependence of styles and contents given an image. This learning framework allows simultaneous style transfer among multiple styles and can be deemed as a special `multi-task' learning scenario. Then the learned encoders and mixer will be taken as the shared knowledge and transferred to new styles and contents. We evaluate the proposed method on Chinese Typeface transfer task and extensive experiments demonstrate its effectiveness.

In our study, the learning process consists of a series of image generation tasks and we try to learn a model which can generalize to novel but related tasks by learning a high-level strategy, namely learning the feature representations. This resembles to ``learning-to-learn" program. In the future, we will explore more about ``learning-to-learn" and integrate it with our framework.

\section*{Acknowledgment}
The work is partially supported by the High Technology Research and Development Program of China 2015AA015801, NSFC 61521062, STCSM 15DZ2270400.

{\small
\bibliographystyle{ieee}
\bibliography{style}

\begin{thebibliography}{10}\itemsep=-1pt

\bibitem{rewrite}
Rewrite.
\newblock \url{https://github.com/kaonashi-tyc/Rewrite}.

\bibitem{zitozi}
Zi-to-zi.
\newblock \url{https://kaonashi-tyc.github.io/2017/04/06/zi2zi.html}.

\bibitem{Bousmalis}
K.~Bousmalis, N.~Silberman, D.~Dohan, D.~Erhan, and D.~Krishnan.
\newblock Unsupervised pixel-level domain adaptation with generative
  adversarial networks.
\newblock In {\em Proceedings of the IEEE Conference on Computer Vision and
  Pattern Recognition}, 2017.

\bibitem{changpinyo}
S.~Changpinyo, W.~Chao, B.~Gong, and F.~Sha.
\newblock Synthesized classifiers for zero-shot learning.
\newblock In {\em Proceedings of the IEEE Conference on Computer Vision and
  Pattern Recognition}, pages 5327--5336, 2016.

\bibitem{chen2017}
D.~Chen, L.~Yuan, J.~Liao, N.~Yu, and G.~Hua.
\newblock Stylebank: An explicit representation for neural image style
  transfer.
\newblock In {\em Proceedings of the IEEE Conference on Computer Vision and
  Pattern Recognition}, 2017.

\bibitem{chen2016fast}
T.~Q. Chen and M.~Schmidt.
\newblock Fast patch-based style transfer of arbitrary style.
\newblock {\em arXiv preprint arXiv:1612.04337}, 2016.

\bibitem{frome}
A.~Frome, G.~Corrado, J.~Shlens, S.~Bengio, J.~Dean, T.~Mikolov, et~al.
\newblock Devise: A deep visual-semantic embedding model.
\newblock In {\em Advances in neural information processing systems}, pages
  2121--2129, 2013.

\bibitem{Gatys}
A.~Gatys, A.~Ecker, and M.~Bethge.
\newblock Image style transfer using convolutional neural networks.
\newblock In {\em Proceedings of the IEEE Conference on Computer Vision and
  Pattern Recognition}, pages 2414--2423, 2016.

\bibitem{Huang_2017_ICCV}
X.~Huang and S.~Belongie.
\newblock Arbitrary style transfer in real-time with adaptive instance
  normalization.
\newblock In {\em Proceedings of the IEEE International Conference on Computer
  Vision (ICCV)}, Oct 2017.

\bibitem{isola}
P.~Isola, J.~Zhu, T.~Zhou, and A.~Efros.
\newblock Image-to-image translation with conditional adversarial networks.
\newblock In {\em Proceedings of the IEEE Conference on Computer Vision and
  Pattern Recognition}, 2017.

\bibitem{jegou}
S.~J{\'e}gou, M.~Drozdzal, D.~Vazquez, A.~Romero, and Y.~Bengio.
\newblock The one hundred layers tiramisu: Fully convolutional densenets for
  semantic segmentation.
\newblock In {\em Proceedings of the IEEE Conference on Computer Vision and
  Pattern Recognition Workshops (CVPRW)}, pages 1175--1183. IEEE, 2017.

\bibitem{johnson}
J.~Johnson, A.~Alahi, and F.~Li.
\newblock Perceptual losses for real-time style transfer and super-resolution.
\newblock In {\em Proceedings of the European Conference on Computer Vision},
  pages 694--711. Springer, 2016.

\bibitem{liu2017}
M.~Y. Liu, T.~Breuel, and J.~Kautz.
\newblock Unsupervised image-to-image translation networks.
\newblock In {\em Advances in Neural Information Processing Systems}, pages
  700--708, 2017.

\bibitem{liu2016}
M.~Y. Liu and O.~Tuzel.
\newblock Coupled generative adversarial networks.
\newblock In {\em Advances in Neural Information Processing Systems 29}, pages
  469--477. 2016.

\bibitem{long}
J.~Long, E.~Shelhamer, and T.~Darrell.
\newblock Fully convolutional networks for semantic segmentation.
\newblock In {\em Proceedings of the IEEE Conference on Computer Vision and
  Pattern Recognition}, pages 3431--3440, 2015.

\bibitem{lyu}
P.~Lyu, X.~Bai, C.~Yao, Z.~Zhu, T.~Huang, and W.~Liu.
\newblock Auto-encoder guided gan for chinese calligraphy synthesis.
\newblock In {\em arXiv preprint arXiv:1706.08789}, 2017.

\bibitem{mordvintsev}
A.~Mordvintsev, C.~Olah, and M.~Tyka.
\newblock Inceptionism: Going deeper into neural networks.
\newblock {\em Google Research Blog. Retrieved June}, 20(14), 2015.

\bibitem{Radford}
A.~Radford, L.~Metz, and S.~Chintala.
\newblock Unsupervised representation learning with deep convolutional
  generative adversarial networks.
\newblock In {\em Proceedings of the International Conference on Learning
  Representations}, 2016.

\bibitem{ronneberger}
O.~Ronneberger, P.~Fischer, and T.~Brox.
\newblock U-net: Convolutional networks for biomedical image segmentation.
\newblock In {\em Proceedings of the International Conference on Medical Image
  Computing and Computer-Assisted Intervention}, pages 234--241. Springer,
  2015.

\bibitem{Shrivastava}
A.~Shrivastava, T.~Pfister, O.~Tuzel, J.~Susskind, W.~Wang, and R.~Webb.
\newblock Learning from simulated and unsupervised images through adversarial
  training.
\newblock In {\em Proceedings of the IEEE Conference on Computer Vision and
  Pattern Recognition}, 2017.

\bibitem{Taigmand}
Y.~Taigman, A.~Polyak, and L.~Wolf.
\newblock Unsupervised cross-domain image generation.
\newblock In {\em arXiv preprint arXiv:1611.02200}, 2016.

\bibitem{Tenenbaum}
J.~Tenenbaum and W.~Freeman.
\newblock Separating style and content.
\newblock In {\em Proceedings of the Advances in neural information processing
  systems}, pages 662--668, 1997.

\bibitem{ulyanov}
D.~Ulyanov, V.~Lebedev, A.~Vedaldi, and V.~Lempitsky.
\newblock Texture networks: Feed-forward synthesis of textures and stylized
  images.
\newblock In {\em Proceedings of the International Conference on Machine
  Learning}, pages 1349--1357, 2016.

\bibitem{upchurch}
P.~Upchurch, N.~Snavely, and K.~Bala.
\newblock From a to z: supervised transfer of style and content using deep
  neural network generators.
\newblock In {\em arXiv preprint arXiv:1603.02003}, 2016.

\bibitem{wilmot}
P.~Wilmot, E.~Risser, and C.~Barnes.
\newblock Stable and controllable neural texture synthesis and style transfer
  using histogram losses.
\newblock {\em arXiv preprint arXiv:1701.08893}, 2017.

\bibitem{xian}
Y.~Xian, Z.~Akata, G.~Sharma, Q.~Nguyen, M.~Hein, and B.~Schiele.
\newblock Latent embeddings for zero-shot classification.
\newblock In {\em Proceedings of the IEEE Conference on Computer Vision and
  Pattern Recognition}, pages 69--77, 2016.

\bibitem{zhang2017multi}
H.~Zhang and K.~Dana.
\newblock Multi-style generative network for real-time transfer.
\newblock In {\em arXiv preprint arXiv:1703.06953}, 2017.

\bibitem{Zhu2017}
J.~Y. Zhu, T.~Park, P.~Isola, and A.~A. Efros.
\newblock Unpaired image-to-image translation using cycle-consistent
  adversarial networks.
\newblock In {\em Proceedings of the IEEE International Conference on Computer
  Vision (ICCV)}, Oct 2017.

\end{thebibliography}
}

\clearpage
\onecolumn 
\begin{center}
\textbf{\large Supplemental Materials: Separating Style and Content for Generalized Style Transfer}
\end{center}

In this supplementary, we present more experimental results to better validate the effectiveness of our proposed method. We first conduct experiments to perform morphing between two styles. Then, we give more experimental results for factors influencing the model performance and present both the quantitative and qualitative results. Finally, we present some results for neural style transfer.

\setcounter{section}{0}
\setcounter{table}{0}
\setcounter{figure}{0}
\setcounter{equation}{0}

\section{Morphing}

\begin{figure*}[!ht]
\centering
\setlength{\abovecaptionskip}{-2pt}
\hspace{-5pt}
\subfigure{
\begin{minipage}{0.49\textwidth}{
\begin{minipage}{0.08\textwidth}$\mathcal{R}_{S_1}$:\end{minipage}
\begin{minipage}{0.44\textwidth}
\includegraphics[width=3.0in,height=0.20in]{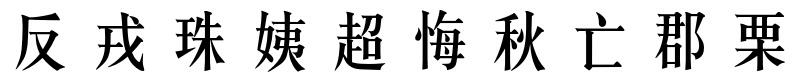}
\end{minipage}\\
\begin{minipage}{0.08\textwidth}$\mathcal{R}_{S_2}$:\end{minipage}
\begin{minipage}{0.44\textwidth}
\includegraphics[width=3.0in,height=0.20in]{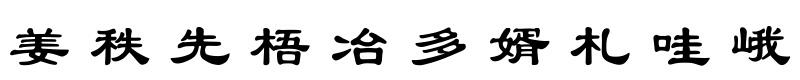}
\end{minipage}\\
\begin{minipage}{0.08\textwidth}TG1:\end{minipage}
\begin{minipage}{0.44\textwidth}
\includegraphics[width=3.0in,height=0.20in]{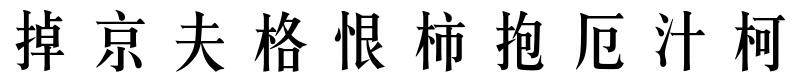}
\end{minipage}\\
\begin{minipage}{0.08\textwidth}0.0:\end{minipage}
\begin{minipage}{0.44\textwidth}
\includegraphics[width=3.0in,height=0.20in]{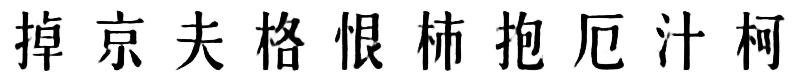}
\end{minipage}\\
\begin{minipage}{0.08\textwidth}0.1:\end{minipage}
\begin{minipage}{0.44\textwidth}
\includegraphics[width=3.0in,height=0.20in]{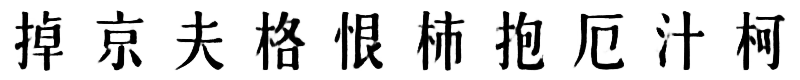}
\end{minipage}\\
\begin{minipage}{0.08\textwidth}0.2:\end{minipage}
\begin{minipage}{0.44\textwidth}
\includegraphics[width=3.0in,height=0.20in]{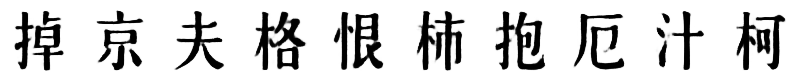}
\end{minipage}\\
\begin{minipage}{0.08\textwidth}0.3:\end{minipage}
\begin{minipage}{0.44\textwidth}
\includegraphics[width=3.0in,height=0.20in]{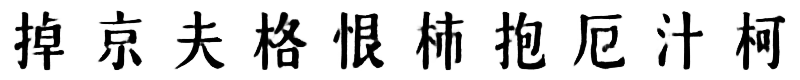}
\end{minipage}\\
\begin{minipage}{0.08\textwidth}0.4:\end{minipage}
\begin{minipage}{0.44\textwidth}
\includegraphics[width=3.0in,height=0.20in]{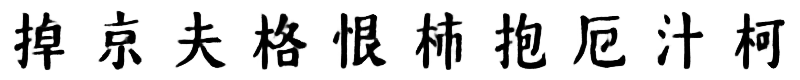}
\end{minipage}\\
\begin{minipage}{0.08\textwidth}0.5:\end{minipage}
\begin{minipage}{0.44\textwidth}
\includegraphics[width=3.0in,height=0.20in]{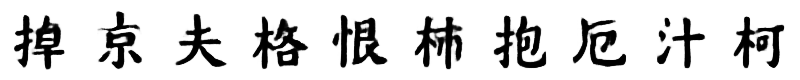}
\end{minipage}\\
\begin{minipage}{0.08\textwidth}0.6:\end{minipage}
\begin{minipage}{0.44\textwidth}
\includegraphics[width=3.0in,height=0.20in]{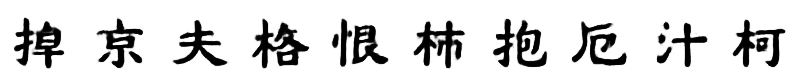}
\end{minipage}\\
\begin{minipage}{0.08\textwidth}0.7:\end{minipage}
\begin{minipage}{0.44\textwidth}
\includegraphics[width=3.0in,height=0.20in]{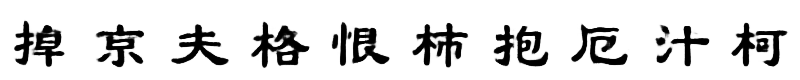}
\end{minipage}\\
\begin{minipage}{0.08\textwidth}0.8:\end{minipage}
\begin{minipage}{0.44\textwidth}
\includegraphics[width=3.0in,height=0.20in]{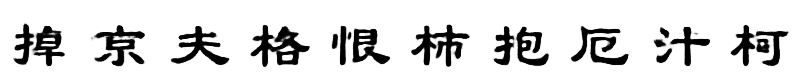}
\end{minipage}\\
\begin{minipage}{0.08\textwidth}0.9:\end{minipage}
\begin{minipage}{0.44\textwidth}
\includegraphics[width=3.0in,height=0.20in]{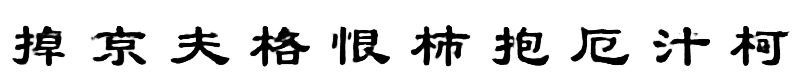}
\end{minipage}\\
\begin{minipage}{0.08\textwidth}1.0:\end{minipage}
\begin{minipage}{0.44\textwidth}
\includegraphics[width=3.0in,height=0.20in]{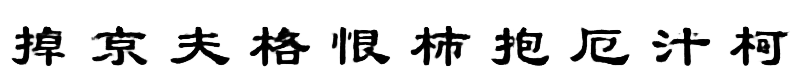}
\end{minipage}\\
\begin{minipage}{0.08\textwidth}TG2:\end{minipage}
\begin{minipage}{0.44\textwidth}
\includegraphics[width=3.0in,height=0.20in]{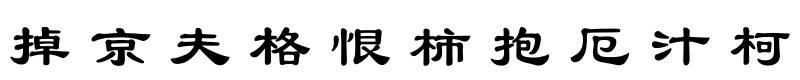}
\end{minipage}
}
\end{minipage}
\rulev
\begin{minipage}{0.49\textwidth}{
\begin{minipage}{0.08\textwidth}$\mathcal{R}_{S_1}$:\end{minipage}
\begin{minipage}{0.44\textwidth}
\includegraphics[width=3.0in,height=0.20in]{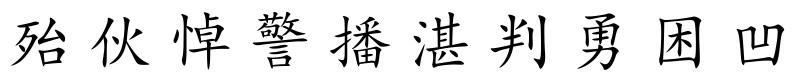}
\end{minipage}\\
\begin{minipage}{0.08\textwidth}$\mathcal{R}_{S_2}$:\end{minipage}
\begin{minipage}{0.44\textwidth}
\includegraphics[width=3.0in,height=0.20in]{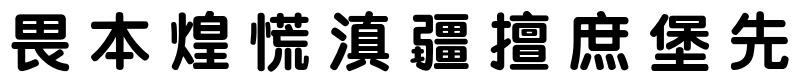}
\end{minipage}\\
\begin{minipage}{0.08\textwidth}TG1:\end{minipage}
\begin{minipage}{0.44\textwidth}
\includegraphics[width=3.0in,height=0.20in]{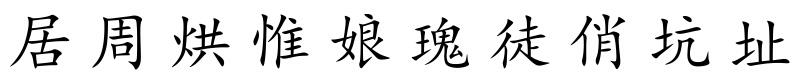}
\end{minipage}\\
\begin{minipage}{0.08\textwidth}0.0:\end{minipage}
\begin{minipage}{0.44\textwidth}
\includegraphics[width=3.0in,height=0.20in]{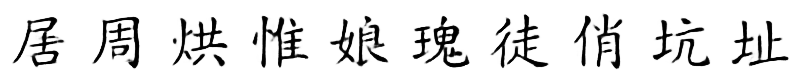}
\end{minipage}\\
\begin{minipage}{0.08\textwidth}0.1:\end{minipage}
\begin{minipage}{0.44\textwidth}
\includegraphics[width=3.0in,height=0.20in]{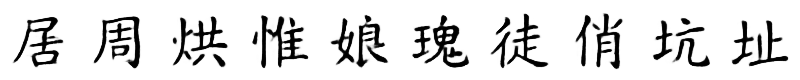}
\end{minipage}\\
\begin{minipage}{0.08\textwidth}0.2:\end{minipage}
\begin{minipage}{0.44\textwidth}
\includegraphics[width=3.0in,height=0.20in]{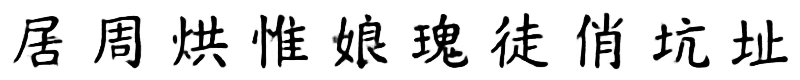}
\end{minipage}\\
\begin{minipage}{0.08\textwidth}0.3:\end{minipage}
\begin{minipage}{0.44\textwidth}
\includegraphics[width=3.0in,height=0.20in]{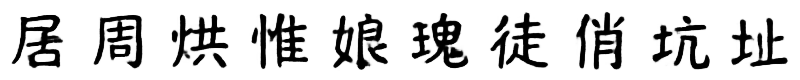}
\end{minipage}\\
\begin{minipage}{0.08\textwidth}0.4:\end{minipage}
\begin{minipage}{0.44\textwidth}
\includegraphics[width=3.0in,height=0.20in]{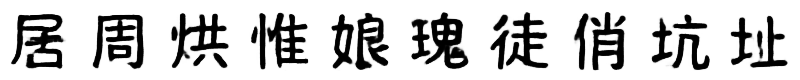}
\end{minipage}\\
\begin{minipage}{0.08\textwidth}0.5:\end{minipage}
\begin{minipage}{0.44\textwidth}
\includegraphics[width=3.0in,height=0.20in]{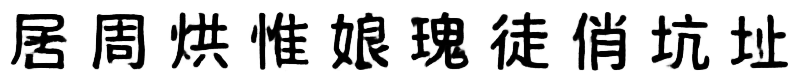}
\end{minipage}\\
\begin{minipage}{0.08\textwidth}0.6:\end{minipage}
\begin{minipage}{0.44\textwidth}
\includegraphics[width=3.0in,height=0.20in]{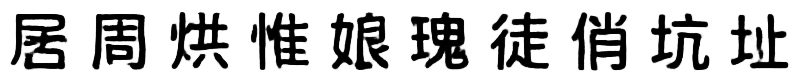}
\end{minipage}\\
\begin{minipage}{0.08\textwidth}0.7:\end{minipage}
\begin{minipage}{0.44\textwidth}
\includegraphics[width=3.0in,height=0.20in]{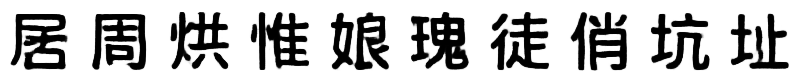}
\end{minipage}\\
\begin{minipage}{0.08\textwidth}0.8:\end{minipage}
\begin{minipage}{0.44\textwidth}
\includegraphics[width=3.0in,height=0.20in]{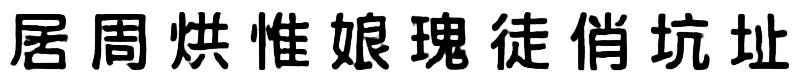}
\end{minipage}\\
\begin{minipage}{0.08\textwidth}0.9:\end{minipage}
\begin{minipage}{0.44\textwidth}
\includegraphics[width=3.0in,height=0.20in]{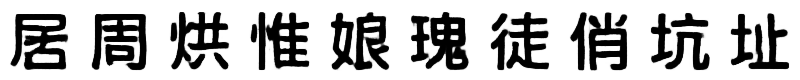}
\end{minipage}\\
\begin{minipage}{0.08\textwidth}1.0:\end{minipage}
\begin{minipage}{0.44\textwidth}
\includegraphics[width=3.0in,height=0.20in]{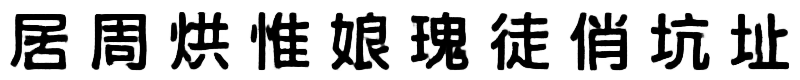}
\end{minipage}\\
\begin{minipage}{0.08\textwidth}TG2:\end{minipage}
\begin{minipage}{0.44\textwidth}
\includegraphics[width=3.0in,height=0.20in]{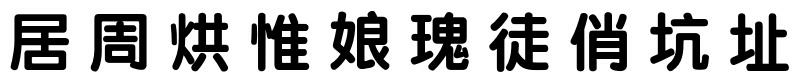}
\end{minipage}
}
\end{minipage}
}
\caption{Results of morphing between two styles. $\mathcal{R}_{S_1}$: Reference set for style $S_1$, $\mathcal{R}_{S_2}$: Reference set for style $S_2$, TG1: Target images for style $S_1$, TG2: Target images for style $S_2$, 0.0-1.0: Outputs for $\lambda$ = [0.0, 0.1, 0.2, 0.3, 0.4, 0.5, 0.6, 0.7, 0.8, 0.9, 1.0].}
\vspace{-15pt}
\label{fig:interpolation1}
\end{figure*}

\begin{figure*}[thb!]
\centering
\setlength{\abovecaptionskip}{-2pt}
\hspace{-5pt}
\subfigure{
\begin{minipage}{0.49\textwidth}{
\begin{minipage}{0.08\textwidth}$\mathcal{R}_{S_1}$:\end{minipage}
\begin{minipage}{0.44\textwidth}
\includegraphics[width=3.0in,height=0.20in]{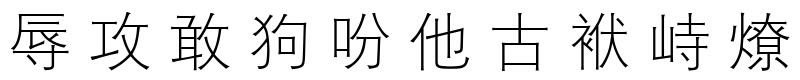}
\end{minipage}\\
\begin{minipage}{0.08\textwidth}$\mathcal{R}_{S_2}$:\end{minipage}
\begin{minipage}{0.44\textwidth}
\includegraphics[width=3.0in,height=0.20in]{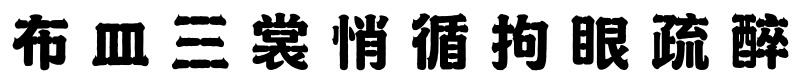}
\end{minipage}\\
\begin{minipage}{0.08\textwidth}TG1:\end{minipage}
\begin{minipage}{0.44\textwidth}
\includegraphics[width=3.0in,height=0.20in]{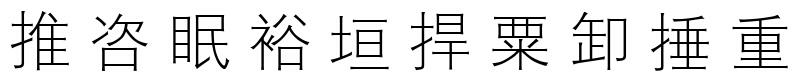}
\end{minipage}\\
\begin{minipage}{0.08\textwidth}0.0:\end{minipage}
\begin{minipage}{0.44\textwidth}
\includegraphics[width=3.0in,height=0.20in]{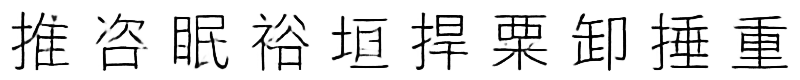}
\end{minipage}\\
\begin{minipage}{0.08\textwidth}0.1:\end{minipage}
\begin{minipage}{0.44\textwidth}
\includegraphics[width=3.0in,height=0.20in]{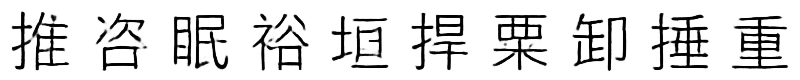}
\end{minipage}\\
\begin{minipage}{0.08\textwidth}0.2:\end{minipage}
\begin{minipage}{0.44\textwidth}
\includegraphics[width=3.0in,height=0.20in]{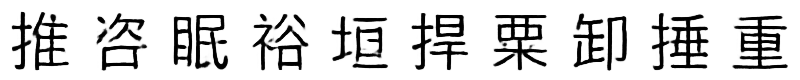}
\end{minipage}\\
\begin{minipage}{0.08\textwidth}0.3:\end{minipage}
\begin{minipage}{0.44\textwidth}
\includegraphics[width=3.0in,height=0.20in]{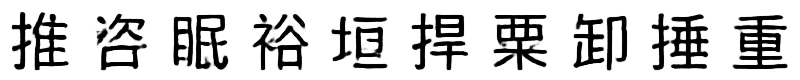}
\end{minipage}\\
\begin{minipage}{0.08\textwidth}0.4:\end{minipage}
\begin{minipage}{0.44\textwidth}
\includegraphics[width=3.0in,height=0.20in]{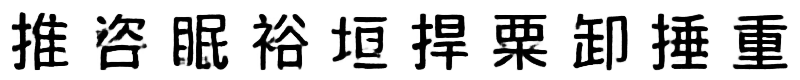}
\end{minipage}\\
\begin{minipage}{0.08\textwidth}0.5:\end{minipage}
\begin{minipage}{0.44\textwidth}
\includegraphics[width=3.0in,height=0.20in]{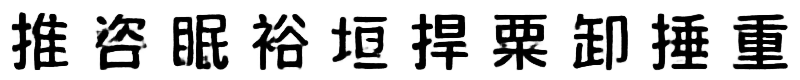}
\end{minipage}\\
\begin{minipage}{0.08\textwidth}0.6:\end{minipage}
\begin{minipage}{0.44\textwidth}
\includegraphics[width=3.0in,height=0.20in]{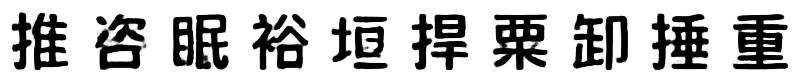}
\end{minipage}\\
\begin{minipage}{0.08\textwidth}0.7:\end{minipage}
\begin{minipage}{0.44\textwidth}
\includegraphics[width=3.0in,height=0.20in]{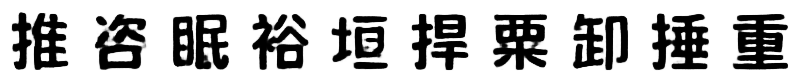}
\end{minipage}\\
\begin{minipage}{0.08\textwidth}0.8:\end{minipage}
\begin{minipage}{0.44\textwidth}
\includegraphics[width=3.0in,height=0.20in]{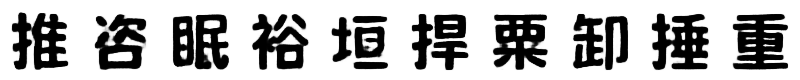}
\end{minipage}\\
\begin{minipage}{0.08\textwidth}0.9:\end{minipage}
\begin{minipage}{0.44\textwidth}
\includegraphics[width=3.0in,height=0.20in]{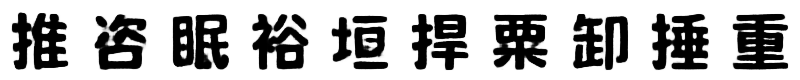}
\end{minipage}\\
\begin{minipage}{0.08\textwidth}1.0:\end{minipage}
\begin{minipage}{0.44\textwidth}
\includegraphics[width=3.0in,height=0.20in]{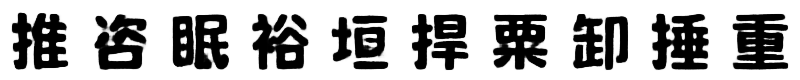}
\end{minipage}\\
\begin{minipage}{0.08\textwidth}TG2:\end{minipage}
\begin{minipage}{0.44\textwidth}
\includegraphics[width=3.0in,height=0.20in]{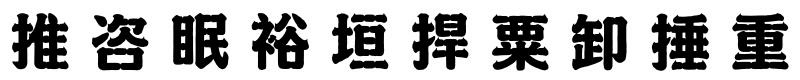}
\end{minipage}
}
\end{minipage}
\rulev
\begin{minipage}{0.49\textwidth}{
\begin{minipage}{0.08\textwidth}$\mathcal{R}_{S_1}$:\end{minipage}
\begin{minipage}{0.44\textwidth}
\includegraphics[width=3.0in,height=0.20in]{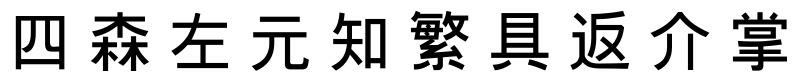}
\end{minipage}\\
\begin{minipage}{0.08\textwidth}$\mathcal{R}_{S_2}$:\end{minipage}
\begin{minipage}{0.44\textwidth}
\includegraphics[width=3.0in,height=0.20in]{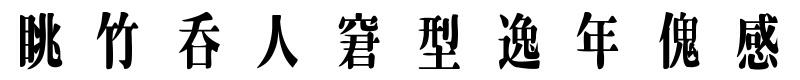}
\end{minipage}\\
\begin{minipage}{0.08\textwidth}TG1:\end{minipage}
\begin{minipage}{0.44\textwidth}
\includegraphics[width=3.0in,height=0.20in]{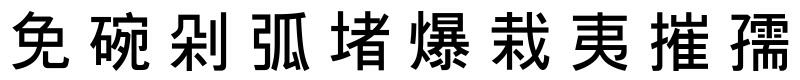}
\end{minipage}\\
\begin{minipage}{0.08\textwidth}0.0:\end{minipage}
\begin{minipage}{0.44\textwidth}
\includegraphics[width=3.0in,height=0.20in]{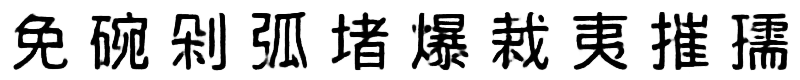}
\end{minipage}\\
\begin{minipage}{0.08\textwidth}0.1:\end{minipage}
\begin{minipage}{0.44\textwidth}
\includegraphics[width=3.0in,height=0.20in]{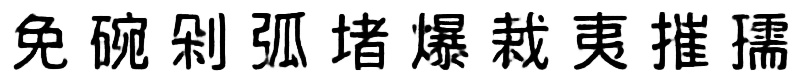}
\end{minipage}\\
\begin{minipage}{0.08\textwidth}0.2:\end{minipage}
\begin{minipage}{0.44\textwidth}
\includegraphics[width=3.0in,height=0.20in]{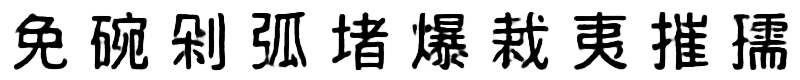}
\end{minipage}\\
\begin{minipage}{0.08\textwidth}0.3:\end{minipage}
\begin{minipage}{0.44\textwidth}
\includegraphics[width=3.0in,height=0.20in]{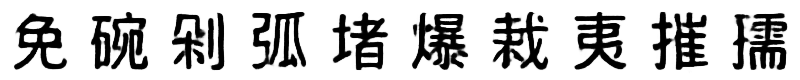}
\end{minipage}\\
\begin{minipage}{0.08\textwidth}0.4:\end{minipage}
\begin{minipage}{0.44\textwidth}
\includegraphics[width=3.0in,height=0.20in]{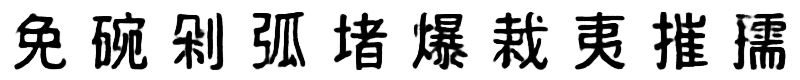}
\end{minipage}\\
\begin{minipage}{0.08\textwidth}0.5:\end{minipage}
\begin{minipage}{0.44\textwidth}
\includegraphics[width=3.0in,height=0.20in]{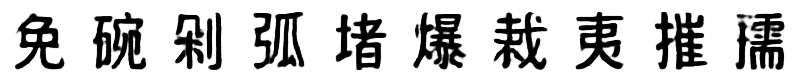}
\end{minipage}\\
\begin{minipage}{0.08\textwidth}0.6:\end{minipage}
\begin{minipage}{0.44\textwidth}
\includegraphics[width=3.0in,height=0.20in]{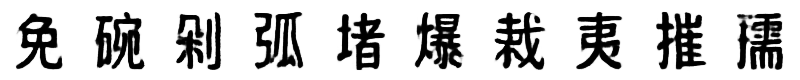}
\end{minipage}\\
\begin{minipage}{0.08\textwidth}0.7:\end{minipage}
\begin{minipage}{0.44\textwidth}
\includegraphics[width=3.0in,height=0.20in]{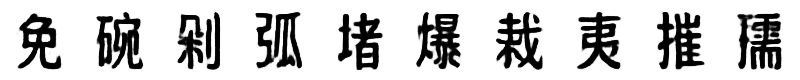}
\end{minipage}\\
\begin{minipage}{0.08\textwidth}0.8:\end{minipage}
\begin{minipage}{0.44\textwidth}
\includegraphics[width=3.0in,height=0.20in]{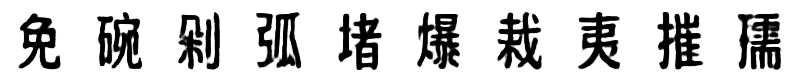}
\end{minipage}\\
\begin{minipage}{0.08\textwidth}0.9:\end{minipage}
\begin{minipage}{0.44\textwidth}
\includegraphics[width=3.0in,height=0.20in]{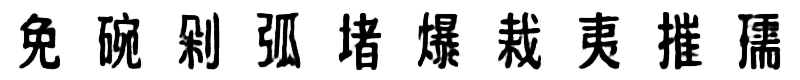}
\end{minipage}\\
\begin{minipage}{0.08\textwidth}1.0:\end{minipage}
\begin{minipage}{0.44\textwidth}
\includegraphics[width=3.0in,height=0.20in]{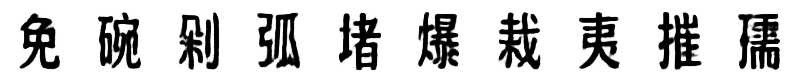}
\end{minipage}\\
\begin{minipage}{0.08\textwidth}TG2:\end{minipage}
\begin{minipage}{0.44\textwidth}
\includegraphics[width=3.0in,height=0.20in]{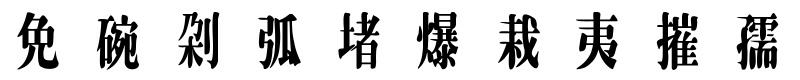}
\end{minipage}
}
\end{minipage}
}
\caption{Results of morphing between two styles. $\mathcal{R}_{S_1}$: Reference set for style $S_1$, $\mathcal{R}_{S_2}$: Reference set for style $S_2$, TG1: Target images for style $S_1$, TG2: Target images for style $S_2$, 0.0-1.0: Outputs for $\lambda$ = [0.0, 0.1, 0.2, 0.3, 0.4, 0.5, 0.6, 0.7, 0.8, 0.9, 1.0].}
\vspace{-15pt}
\label{fig:interpolation2}
\end{figure*}

In this subsection, we perform morphing between two styles. We synthesize new styles by changing the weight between two styles $S_1$ and $S_2$ according to the following function:
\begin{equation}
S_{\textrm{New}} = (1-\lambda) \times S_1 + \lambda \times S_2, \quad \quad 0 \leq \lambda \leq 1.
\end{equation}
The styles and contents used in this experiment are all novel. During experiment, we first extract the style features for the two styles from style reference sets $\mathcal{R}_{S_1}$ and $\mathcal{R}_{S_2}$ and then combine them with different weight $\lambda$. Finally, the new style feature $S_{\textrm{New}}$ will be combined with the content feature and generate the image. The results are presented in Figure~\ref{fig:interpolation1} and Figure~\ref{fig:interpolation2}. From the figure, we can observe the changing process from style $S_1$ to style $S_2$. This experiment further validates that the $\textsl{Style Encoder}$ can extract accurate and pure style features. Besides, by separating style and content, we can leverage the style representations to create new styles.

\newpage

\clearpage
\section{Influence of the Training Set Size}

In this section, we present the quantitative results of different training set size in Table~\ref{tab:com_train_size_supp} and the qualitative results in Figure~\ref{fig:train-size_supp}. We can observe that for both quantitative results and qualitative results, the larger the training set size, the better the performance. In addition, the model performance saturates with the increase of the training set size.

 \begin{table*}[!htbp]\small
     \centering
     \addtolength{\tabcolsep}{-1pt}
     \setlength{\abovecaptionskip}{-2pt}
     \setlength{\belowcaptionskip}{-8pt}
     \caption{Quantitative comparison of models with different training set size.}
     \begin{tabular}{c|ccc|ccc|ccc|ccc}
     \hline
         & \multicolumn{3}{c}{$D_1$} & \multicolumn{3}{|c}{$D_2$} & \multicolumn{3}{|c}{$D_3$} & \multicolumn{3}{|c}{$D_4$} \\

     \hline
                 & L1 loss & RMSE  & PDAR  & L1 loss & RMSE  & PDAR  & L1 loss & RMSE  & PDAR  & L1 loss & RMSE  & PDAR  \\
     \hline
     20k  &  0.0096  &  0.0192  &  0.1801  & 0.0096  &  0.0192  &  0.1806  & 0.0095  &  0.0191  &  0.1758  & 0.0095  &  0.0191  &  0.1764  \\

     \hline
     50k     & 0.0096  &  0.0191  &  0.1713  & 0.0097  &  0.0192  &  0.1726  & 0.0095  &  0.0191  &  0.1668  & 0.0096  &  0.0192  &  0.1679  \\

     \hline
     100k     & 0.0093  &  0.0188  &  0.1662  & \textbf{0.0094}  &  \textbf{0.0189}  &  0.1686  & 0.0093  &  0.0188  &  0.1633  & \textbf{0.0094}  &  \textbf{0.0189}  &  0.1654 \\

     \hline
     300k     & \textbf{0.0091}  &  \textbf{0.0185}  &  0.1549 &  \textbf{0.0094}  &  \textbf{0.0189}  &  0.1604 &  \textbf{0.0092}  &  \textbf{0.0187}  &  0.1549 &  \textbf{0.0094}  &  \textbf{0.0189}  &  0.1592\\

     \hline
     500k     & \textbf{0.0091}  &  \textbf{0.0185}  &  \textbf{0.1509}  &  \textbf{0.0094}  &  \textbf{0.0189}  &  \textbf{0.1578}  & \textbf{0.0092}  &  \textbf{0.0187}  &  \textbf{0.1519}  &  0.0095  &  0.019  &  \textbf{0.1569}   \\

     \hline
     \end{tabular}%
     \vspace{-15pt}
   \label{tab:com_train_size_supp}%
 \end{table*}

\begin{figure*}[!ht]
\centering
\setlength{\abovecaptionskip}{5pt}
\hspace{-6pt}
\subfigure{
\begin{minipage}{0.49\textwidth}{\vspace{-5pt}
\begin{minipage}{0.05\textwidth}TG:\end{minipage}
\begin{minipage}{0.45\textwidth}
\includegraphics[width=3.2in,height=0.15in]{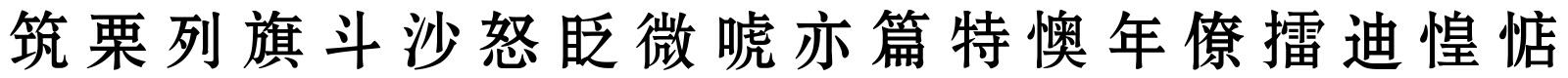}
\end{minipage}\\
\begin{minipage}{0.05\textwidth}O1:\end{minipage}
\begin{minipage}{0.45\textwidth}
\includegraphics[width=3.2in,height=0.15in]{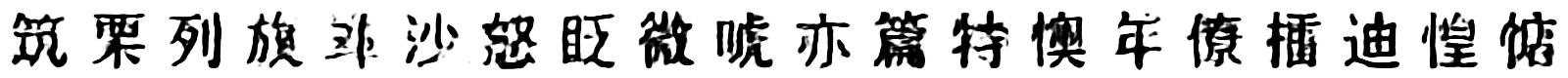}
\end{minipage}\\
\begin{minipage}{0.05\textwidth}O2:\end{minipage}
\begin{minipage}{0.45\textwidth}
\includegraphics[width=3.2in,height=0.15in]{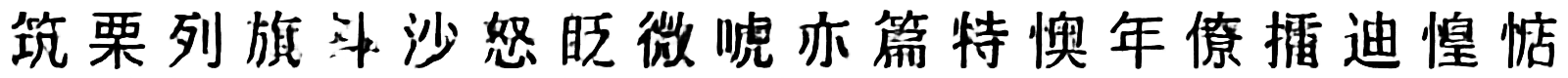}
\end{minipage}\\
\begin{minipage}{0.05\textwidth}O3:\end{minipage}
\begin{minipage}{0.45\textwidth}
\includegraphics[width=3.2in,height=0.15in]{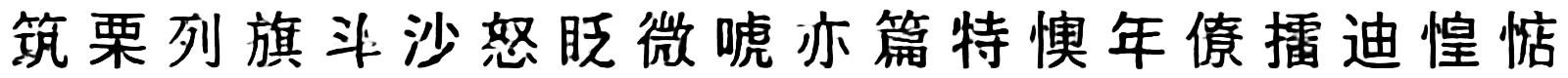}
\end{minipage}\\
\begin{minipage}{0.05\textwidth}O4:\end{minipage}
\begin{minipage}{0.45\textwidth}
\includegraphics[width=3.2in,height=0.15in]{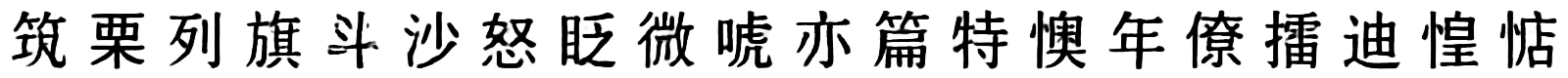}
\end{minipage}\\
\begin{minipage}{0.05\textwidth}O5:\end{minipage}
\begin{minipage}{0.45\textwidth}
\includegraphics[width=3.2in,height=0.15in]{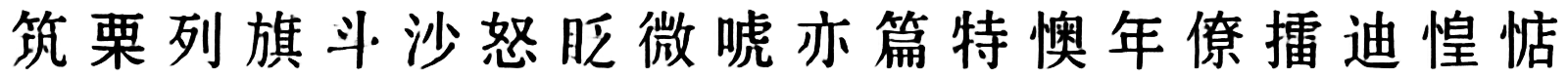}
\end{minipage}
\vspace{3pt}\\
\begin{minipage}{0.05\textwidth}TG:\end{minipage}
\begin{minipage}{0.45\textwidth}
\includegraphics[width=3.2in,height=0.15in]{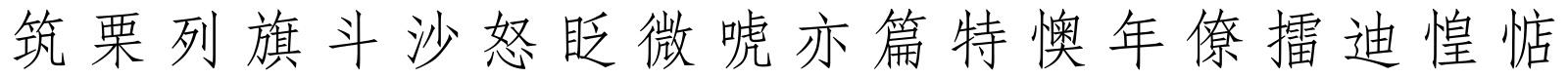}
\end{minipage}\\
\begin{minipage}{0.05\textwidth}O1:\end{minipage}
\begin{minipage}{0.45\textwidth}
\includegraphics[width=3.2in,height=0.15in]{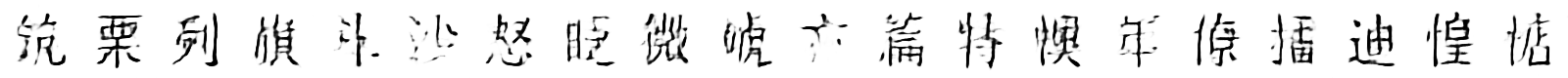}
\end{minipage}\\
\begin{minipage}{0.05\textwidth}O2:\end{minipage}
\begin{minipage}{0.45\textwidth}
\includegraphics[width=3.2in,height=0.15in]{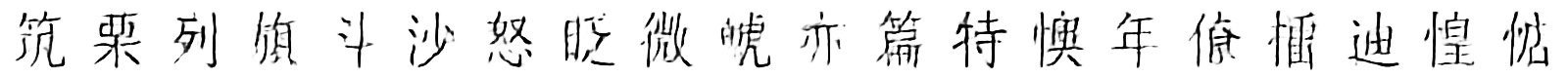}
\end{minipage}\\
\begin{minipage}{0.05\textwidth}O3:\end{minipage}
\begin{minipage}{0.45\textwidth}
\includegraphics[width=3.2in,height=0.15in]{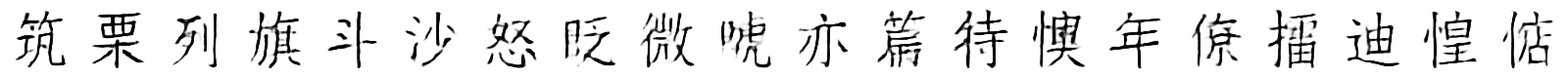}
\end{minipage}\\
\begin{minipage}{0.05\textwidth}O4:\end{minipage}
\begin{minipage}{0.45\textwidth}
\includegraphics[width=3.2in,height=0.15in]{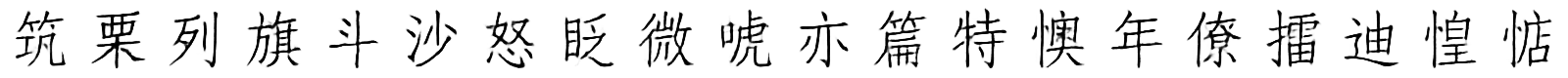}
\end{minipage}\\
\begin{minipage}{0.05\textwidth}O5:\end{minipage}
\begin{minipage}{0.45\textwidth}
\includegraphics[width=3.2in,height=0.15in]{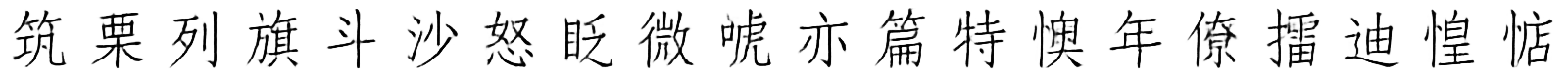}
\end{minipage}
\vspace{3pt}\\
\begin{minipage}{0.05\textwidth}TG:\end{minipage}
\begin{minipage}{0.45\textwidth}
\includegraphics[width=3.2in,height=0.15in]{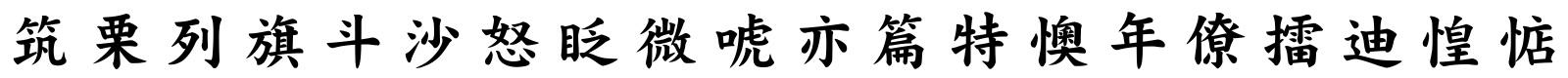}
\end{minipage}\\
\begin{minipage}{0.05\textwidth}O1:\end{minipage}
\begin{minipage}{0.45\textwidth}
\includegraphics[width=3.2in,height=0.15in]{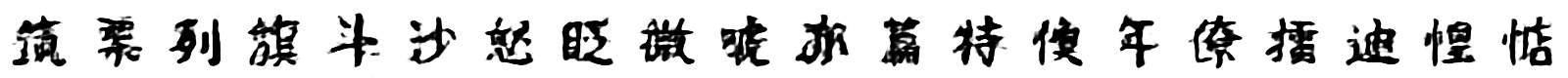}
\end{minipage}\\
\begin{minipage}{0.05\textwidth}O2:\end{minipage}
\begin{minipage}{0.45\textwidth}
\includegraphics[width=3.2in,height=0.15in]{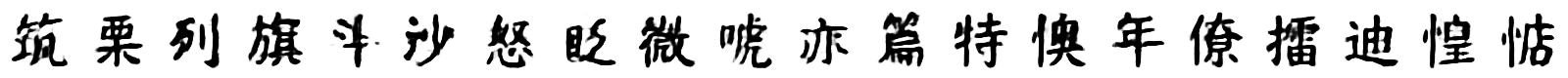}
\end{minipage}\\
\begin{minipage}{0.05\textwidth}O3:\end{minipage}
\begin{minipage}{0.45\textwidth}
\includegraphics[width=3.2in,height=0.15in]{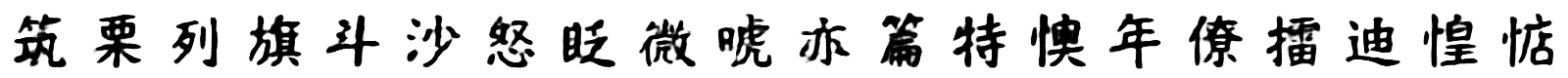}
\end{minipage}\\
\begin{minipage}{0.05\textwidth}O4:\end{minipage}
\begin{minipage}{0.45\textwidth}
\includegraphics[width=3.2in,height=0.15in]{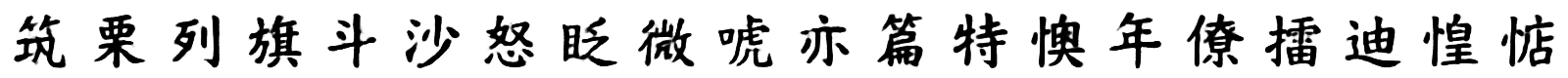}
\end{minipage}\\
\begin{minipage}{0.05\textwidth}O5:\end{minipage}
\begin{minipage}{0.45\textwidth}
\includegraphics[width=3.2in,height=0.15in]{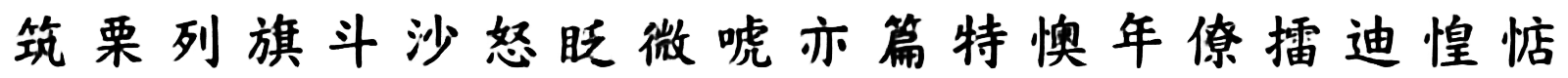}
\end{minipage}
\ruleh
\\
\begin{minipage}{0.05\textwidth}TG:\end{minipage}
\begin{minipage}{0.45\textwidth}
\includegraphics[width=3.2in,height=0.15in]{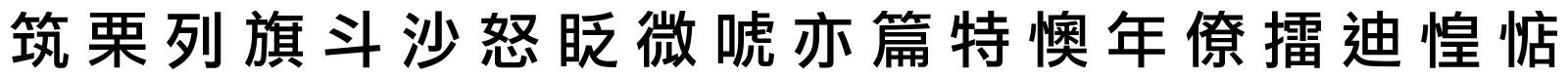}
\end{minipage}\\
\begin{minipage}{0.05\textwidth}O1:\end{minipage}
\begin{minipage}{0.45\textwidth}
\includegraphics[width=3.2in,height=0.15in]{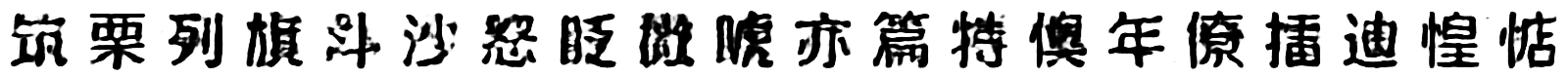}
\end{minipage}\\
\begin{minipage}{0.05\textwidth}O2:\end{minipage}
\begin{minipage}{0.45\textwidth}
\includegraphics[width=3.2in,height=0.15in]{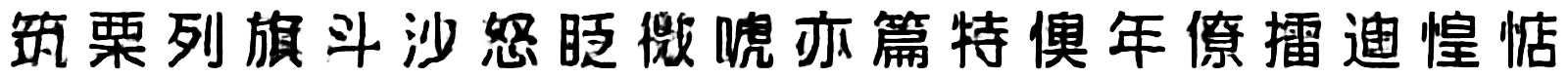}
\end{minipage}\\
\begin{minipage}{0.05\textwidth}O3:\end{minipage}
\begin{minipage}{0.45\textwidth}
\includegraphics[width=3.2in,height=0.15in]{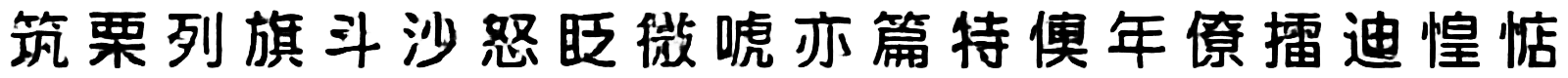}
\end{minipage}\\
\begin{minipage}{0.05\textwidth}O4:\end{minipage}
\begin{minipage}{0.45\textwidth}
\includegraphics[width=3.2in,height=0.15in]{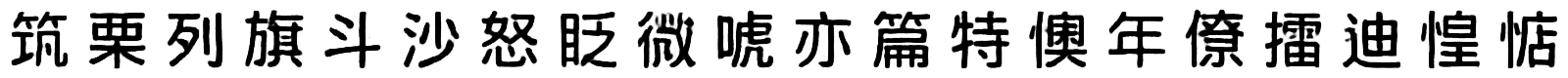}
\end{minipage}\\
\begin{minipage}{0.05\textwidth}O5:\end{minipage}
\begin{minipage}{0.45\textwidth}
\includegraphics[width=3.2in,height=0.15in]{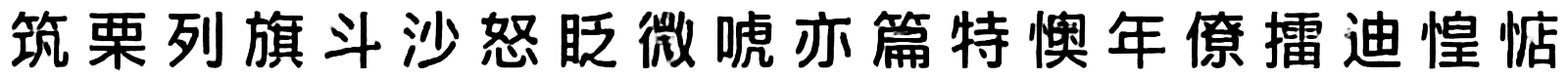}
\end{minipage}
\vspace{3pt}\\
\begin{minipage}{0.05\textwidth}TG:\end{minipage}
\begin{minipage}{0.45\textwidth}
\includegraphics[width=3.2in,height=0.15in]{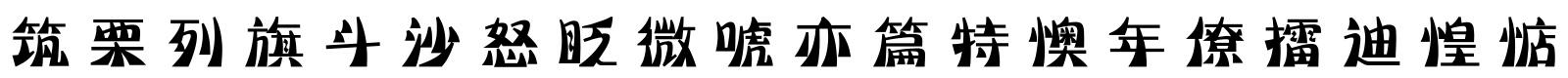}
\end{minipage}\\
\begin{minipage}{0.05\textwidth}O1:\end{minipage}
\begin{minipage}{0.45\textwidth}
\includegraphics[width=3.2in,height=0.15in]{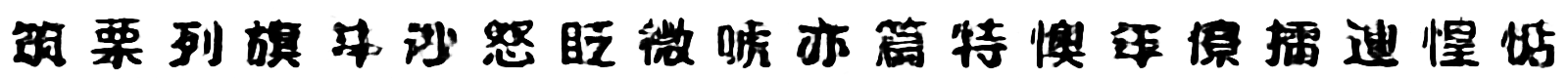}
\end{minipage}\\
\begin{minipage}{0.05\textwidth}O2:\end{minipage}
\begin{minipage}{0.45\textwidth}
\includegraphics[width=3.2in,height=0.15in]{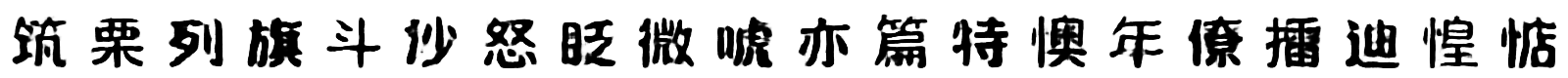}
\end{minipage}\\
\begin{minipage}{0.05\textwidth}O3:\end{minipage}
\begin{minipage}{0.45\textwidth}
\includegraphics[width=3.2in,height=0.15in]{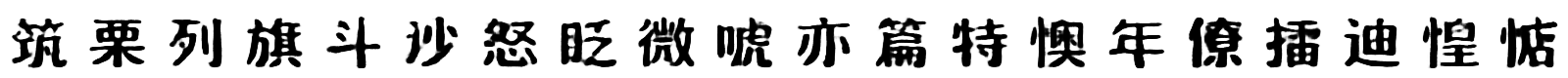}
\end{minipage}\\
\begin{minipage}{0.05\textwidth}O4:\end{minipage}
\begin{minipage}{0.45\textwidth}
\includegraphics[width=3.2in,height=0.15in]{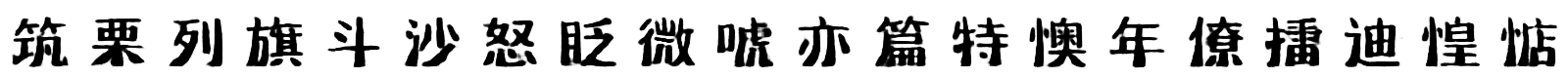}
\end{minipage}\\
\begin{minipage}{0.05\textwidth}O5:\end{minipage}
\begin{minipage}{0.45\textwidth}
\includegraphics[width=3.2in,height=0.15in]{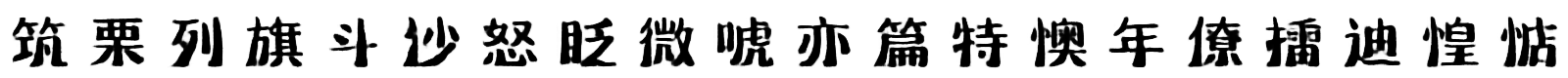}
\end{minipage}
\vspace{3pt}\\
\begin{minipage}{0.05\textwidth}TG:\end{minipage}
\begin{minipage}{0.45\textwidth}
\includegraphics[width=3.2in,height=0.15in]{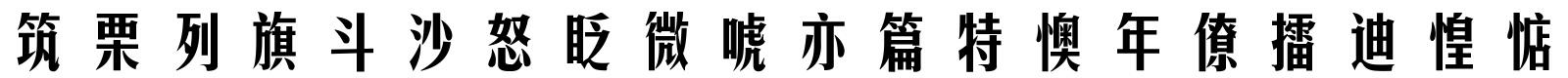}
\end{minipage}\\
\begin{minipage}{0.05\textwidth}O1:\end{minipage}
\begin{minipage}{0.45\textwidth}
\includegraphics[width=3.2in,height=0.15in]{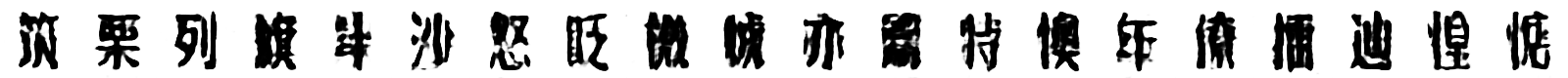}
\end{minipage}\\
\begin{minipage}{0.05\textwidth}O2:\end{minipage}
\begin{minipage}{0.45\textwidth}
\includegraphics[width=3.2in,height=0.15in]{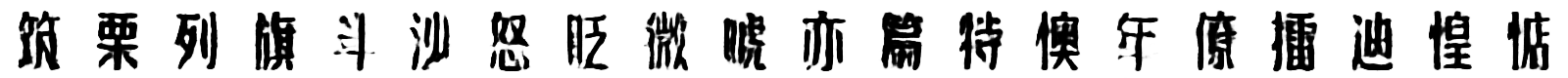}
\end{minipage}\\
\begin{minipage}{0.05\textwidth}O3:\end{minipage}
\begin{minipage}{0.45\textwidth}
\includegraphics[width=3.2in,height=0.15in]{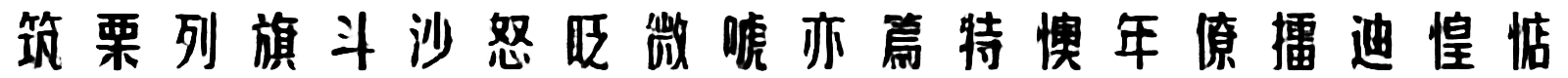}
\end{minipage}\\
\begin{minipage}{0.05\textwidth}O4:\end{minipage}
\begin{minipage}{0.45\textwidth}
\includegraphics[width=3.2in,height=0.15in]{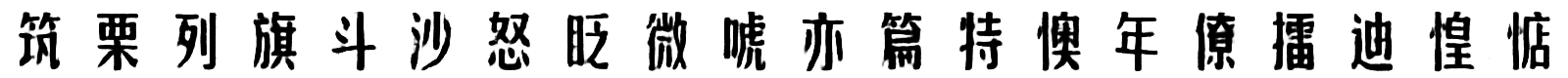}
\end{minipage}\\
\begin{minipage}{0.05\textwidth}O5:\end{minipage}
\begin{minipage}{0.45\textwidth}
\includegraphics[width=3.2in,height=0.15in]{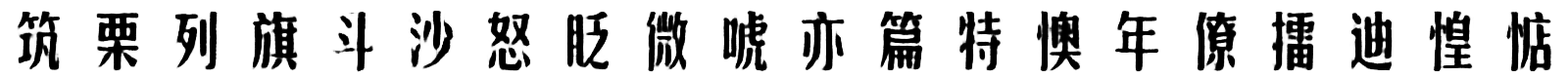}
\end{minipage}
}
\end{minipage}
}
\vspace{-5pt}
\rulev
\hspace{-3pt}
\subfigure{
\begin{minipage}{0.45\textwidth}{\vspace{-5pt}
\includegraphics[width=3.2in,height=0.15in]{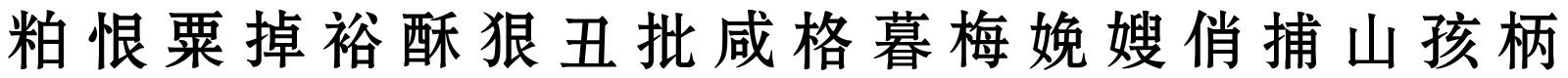}\\
\includegraphics[width=3.2in,height=0.15in]{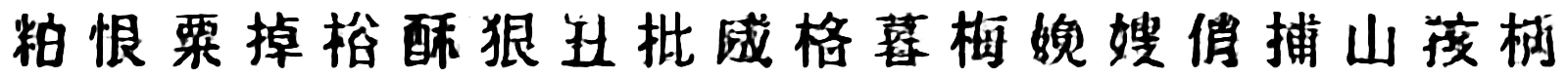}\\
\includegraphics[width=3.2in,height=0.15in]{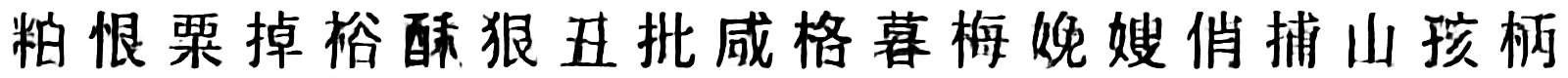}\\
\includegraphics[width=3.2in,height=0.15in]{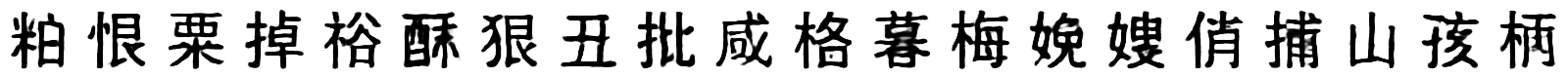}\\
\includegraphics[width=3.2in,height=0.15in]{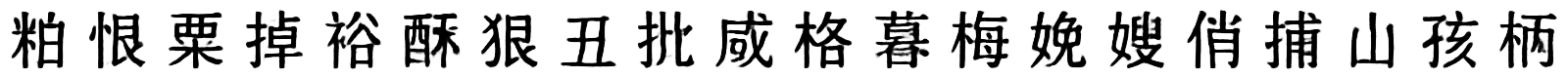}\\
\includegraphics[width=3.2in,height=0.15in]{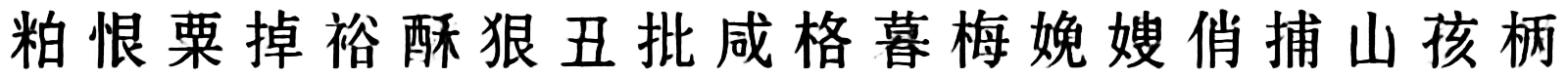}\vspace{3pt}\\
\includegraphics[width=3.2in,height=0.15in]{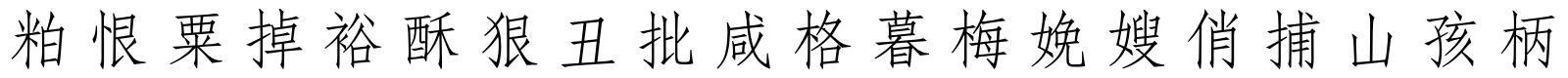}\\
\includegraphics[width=3.2in,height=0.15in]{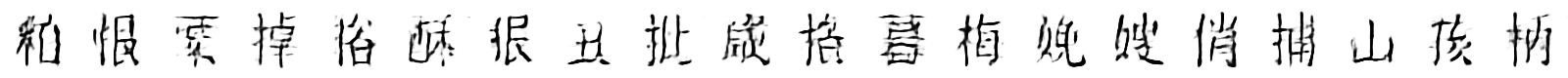}\\
\includegraphics[width=3.2in,height=0.15in]{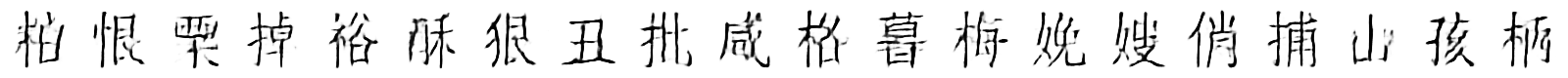}\\
\includegraphics[width=3.2in,height=0.15in]{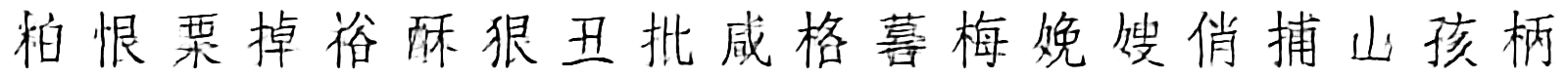}\\
\includegraphics[width=3.2in,height=0.15in]{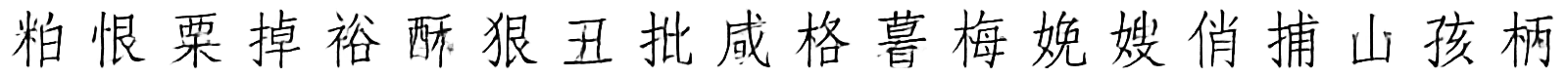}\\
\includegraphics[width=3.2in,height=0.15in]{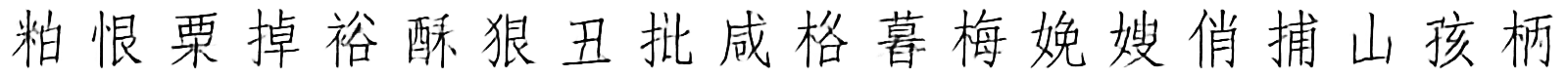}\vspace{3pt}
\includegraphics[width=3.2in,height=0.15in]{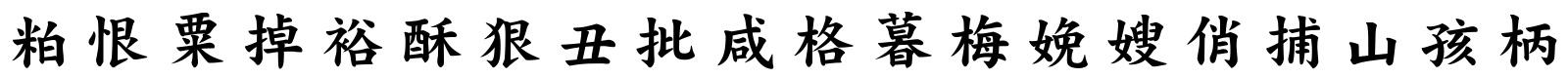}\\
\includegraphics[width=3.2in,height=0.15in]{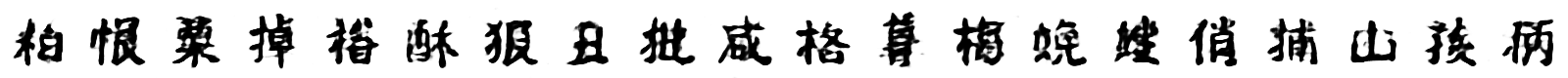}\\
\includegraphics[width=3.2in,height=0.15in]{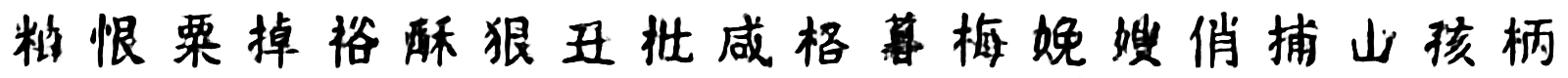}\\
\includegraphics[width=3.2in,height=0.15in]{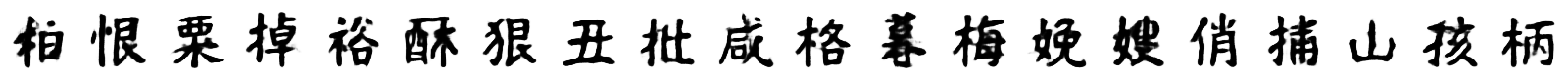}\\
\includegraphics[width=3.2in,height=0.15in]{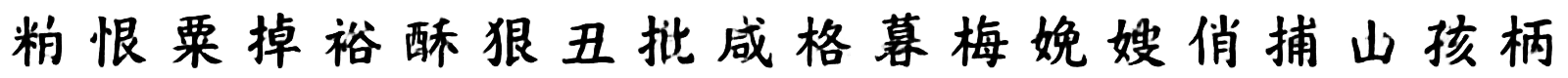}\\
\includegraphics[width=3.2in,height=0.15in]{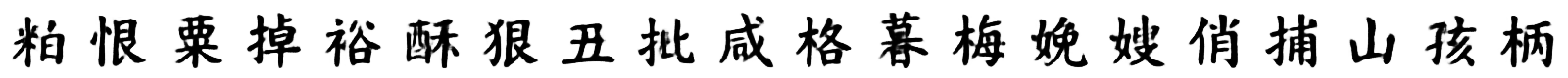}
\ruleh
\vspace{3pt}
\\
\includegraphics[width=3.2in,height=0.15in]{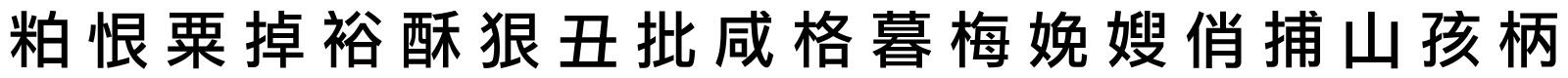}\\
\includegraphics[width=3.2in,height=0.15in]{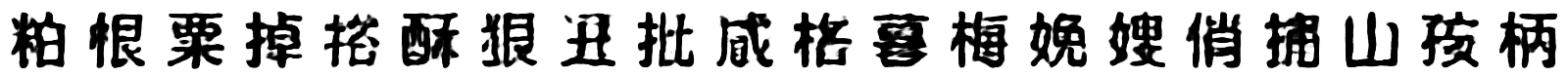}\\
\includegraphics[width=3.2in,height=0.15in]{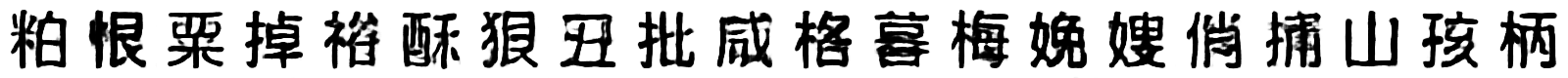}\\
\includegraphics[width=3.2in,height=0.15in]{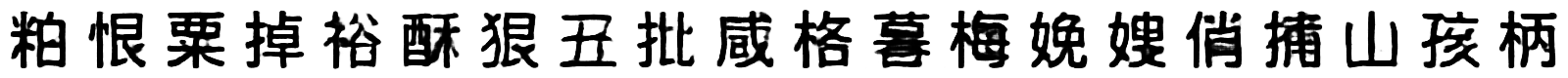}\\
\includegraphics[width=3.2in,height=0.15in]{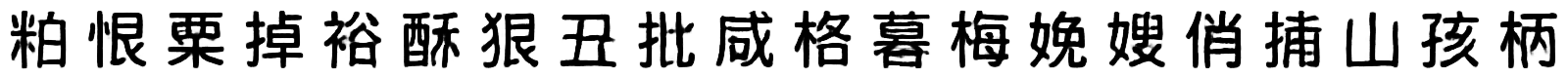}\\
\includegraphics[width=3.2in,height=0.15in]{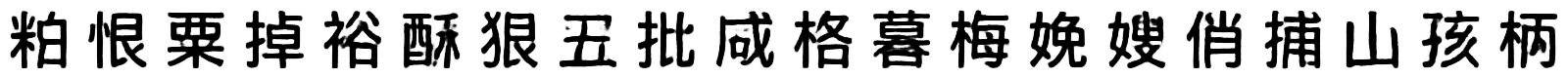}\vspace{3pt}\\
\includegraphics[width=3.2in,height=0.15in]{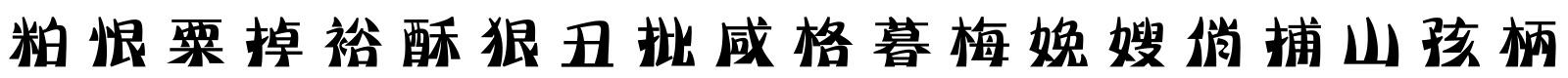}\\
\includegraphics[width=3.2in,height=0.15in]{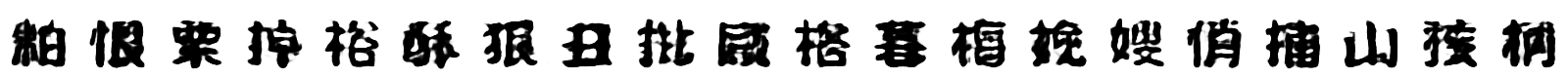}\\
\includegraphics[width=3.2in,height=0.15in]{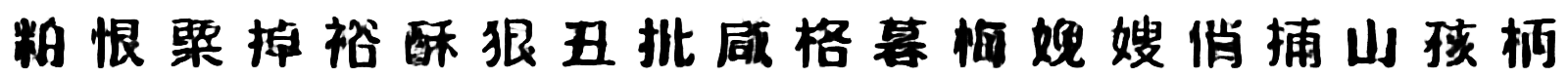}\\
\includegraphics[width=3.2in,height=0.15in]{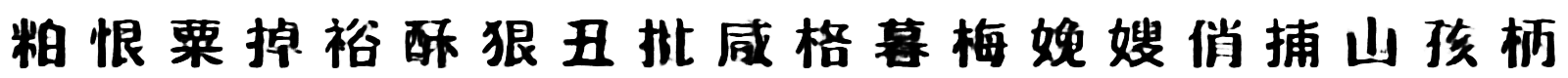}\\
\includegraphics[width=3.2in,height=0.15in]{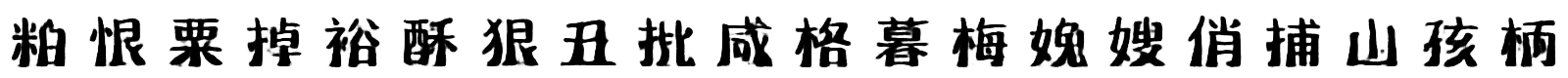}\\
\includegraphics[width=3.2in,height=0.15in]{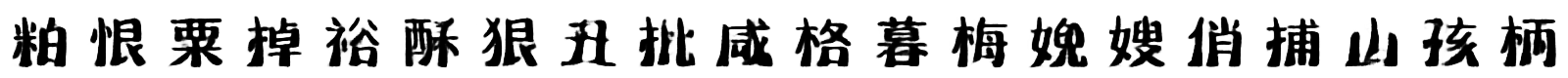}\vspace{3pt}\\
\includegraphics[width=3.2in,height=0.15in]{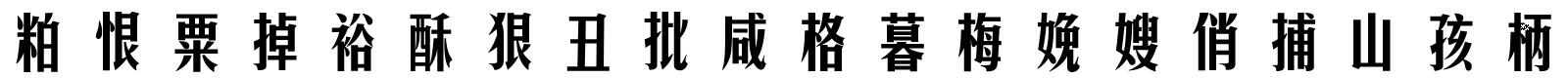}\\
\includegraphics[width=3.2in,height=0.15in]{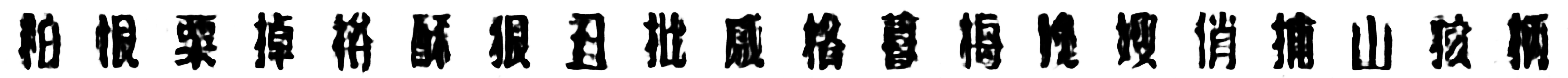}\\
\includegraphics[width=3.2in,height=0.15in]{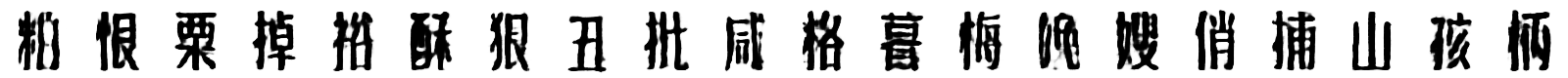}\\
\includegraphics[width=3.2in,height=0.15in]{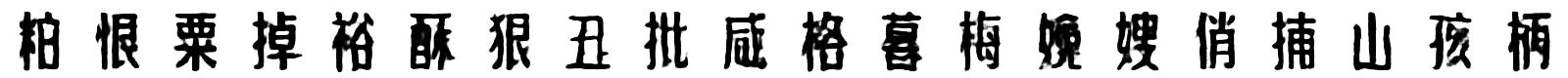}\\
\includegraphics[width=3.2in,height=0.15in]{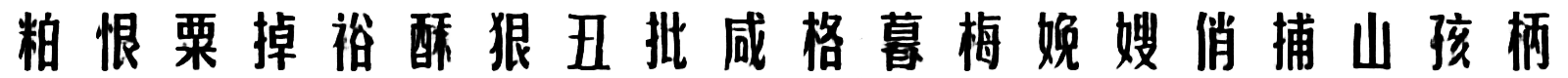}\\
\includegraphics[width=3.2in,height=0.15in]{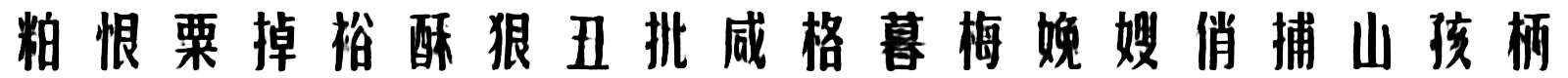}

}
\end{minipage}
}
\caption{Generation for $D_1$, $D_2$, $D_3$, $D_4$ (from upper left to lower right) with different training set size. TG: Target image, O1: Output for $N_{t}$=20k, O2: Output for $N_{t}$=50k, O3: Output for $N_{t}$=100k, O4: Output for $N_{t}$=300k, O5: Output for $N_{t}$=500k. In all cases, $r$=10.}
\label{fig:train-size_supp}
\vspace{-5pt}
\end{figure*}

\section{Influence of Reference Set Size}

Following, we present the quantitative results of different reference set size in Table~\ref{tab:com_ref_num_supp} and more generated images in Figure~\ref{fig:ref-num_supp}. From the figure, we can observe that $r$=2 performs worst and $r$=10 and $r$=15 perform closely, indicating that more reference images will provide more information and the performance will be saturated with the increase of reference set size.

 \begin{table*}[!htbp]\small
     \centering
     \addtolength{\tabcolsep}{-1pt}
     \setlength{\abovecaptionskip}{-2pt}
     \setlength{\belowcaptionskip}{-6pt}
     \caption{Quantitative comparison of models with different reference set size. In all cases, $N_{t}$=300k.}
     \begin{tabular}{c|ccc|ccc|ccc|ccc}
     \hline
         & \multicolumn{3}{c}{$D_1$} & \multicolumn{3}{|c}{$D_2$} & \multicolumn{3}{|c}{$D_3$} & \multicolumn{3}{|c}{$D_4$} \\

     \hline
            & L1 loss & RMSE  & PDAR  & L1 loss & RMSE  & PDAR  & L1 loss & RMSE  & PDAR  & L1 loss & RMSE  & PDAR  \\
     \hline
     r=2  &  0.0096  &  0.0191  &  0.1635 &  0.0098  &  0.0193  &  0.1677 &  0.0097  &  0.0192  &  0.1611 &  0.0098  &  0.0193  &  0.1649\\

     \hline
     r=5  &  0.0093  &  0.0188  &  0.1594  &  0.0095  &  0.019  &  0.1641  &  0.0094  &  0.0189  &  0.1578  &  0.0096  &  0.0192  &  0.1615 \\

     \hline
     r=10  & 0.0091  &  0.0185  &  0.1549 &  0.0094  &  0.0189  &  0.1604 &  0.0092  &  0.0187  &  0.1549 &  0.0094  &  0.0189  &  0.1592\\

     \hline
     r=15    &  0.0091  &  0.0186  &  0.1557 &  0.0094  & 0.0189  &  0.1601 &  0.0092  &  0.0187  &  0.1552 &  0.0095  &  0.019  &  0.1584 \\

     \hline
     r=50    &  \textbf{0.009}  &  \textbf{0.0184}  &  \textbf{0.1533} &  \textbf{0.0092}  &  \textbf{0.0187}  &  \textbf{0.1585} &  \textbf{0.0091}  &  \textbf{0.0185}  &  \textbf{0.1537} &  \textbf{0.0093}  &  \textbf{0.0188}  &  \textbf{0.1571} \\

     \hline
     \end{tabular}%
     \vspace{-15pt}
   \label{tab:com_ref_num_supp}%
 \end{table*}

\begin{figure*}[!htpb]
\centering
\setlength{\abovecaptionskip}{-2pt}
\hspace{-15pt}
\subfigure{
\begin{minipage}{0.49\textwidth}{\vspace{-5pt}
\begin{minipage}{0.05\textwidth}TG:\end{minipage}
\begin{minipage}{0.45\textwidth}
\includegraphics[width=3.2in,height=0.15in]{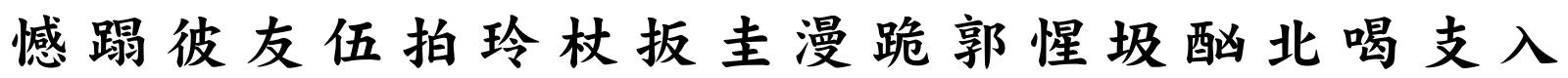}
\end{minipage}\\
\begin{minipage}{0.05\textwidth}O1:\end{minipage}
\begin{minipage}{0.45\textwidth}
\includegraphics[width=3.2in,height=0.15in]{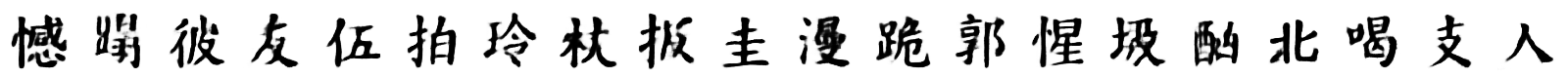}
\end{minipage}\\
\begin{minipage}{0.05\textwidth}O2:\end{minipage}
\begin{minipage}{0.45\textwidth}
\includegraphics[width=3.2in,height=0.15in]{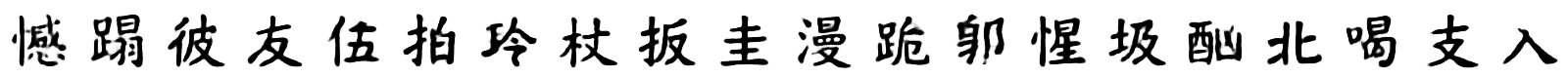}
\end{minipage}\\
\begin{minipage}{0.05\textwidth}O3:\end{minipage}
\begin{minipage}{0.45\textwidth}
\includegraphics[width=3.2in,height=0.15in]{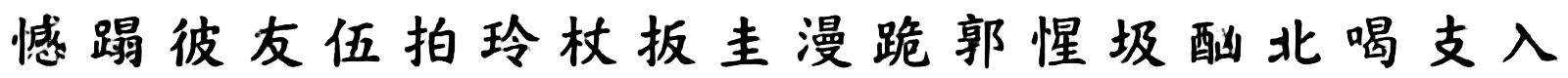}
\end{minipage}\\
\begin{minipage}{0.05\textwidth}O4:\end{minipage}
\begin{minipage}{0.45\textwidth}
\includegraphics[width=3.2in,height=0.15in]{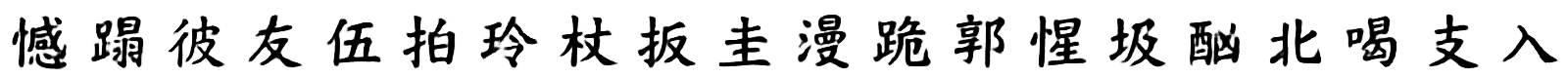}
\end{minipage}\\
\begin{minipage}{0.05\textwidth}O5:\end{minipage}
\begin{minipage}{0.45\textwidth}
\includegraphics[width=3.2in,height=0.15in]{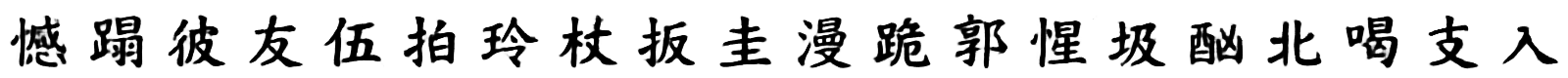}
\end{minipage}
\vspace{5pt}\\
\begin{minipage}{0.05\textwidth}TG:\end{minipage}
\begin{minipage}{0.45\textwidth}
\includegraphics[width=3.2in,height=0.15in]{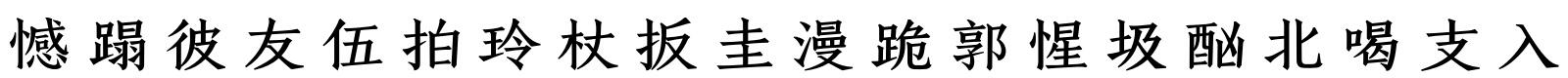}
\end{minipage}\\
\begin{minipage}{0.05\textwidth}O1:\end{minipage}
\begin{minipage}{0.45\textwidth}
\includegraphics[width=3.2in,height=0.15in]{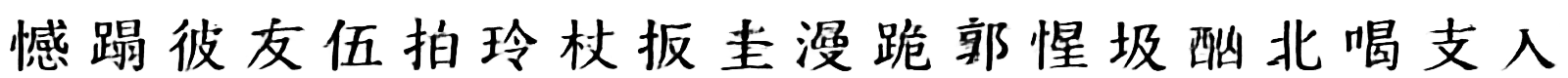}
\end{minipage}\\
\begin{minipage}{0.05\textwidth}O2:\end{minipage}
\begin{minipage}{0.45\textwidth}
\includegraphics[width=3.2in,height=0.15in]{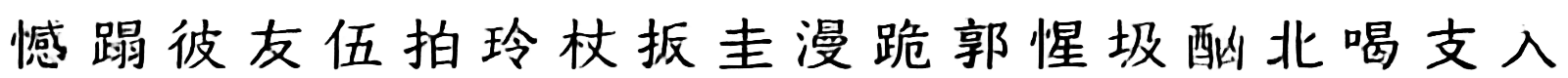}
\end{minipage}\\
\begin{minipage}{0.05\textwidth}O3:\end{minipage}
\begin{minipage}{0.45\textwidth}
\includegraphics[width=3.2in,height=0.15in]{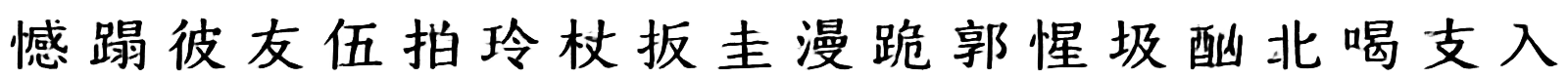}
\end{minipage}\\
\begin{minipage}{0.05\textwidth}O4:\end{minipage}
\begin{minipage}{0.45\textwidth}
\includegraphics[width=3.2in,height=0.15in]{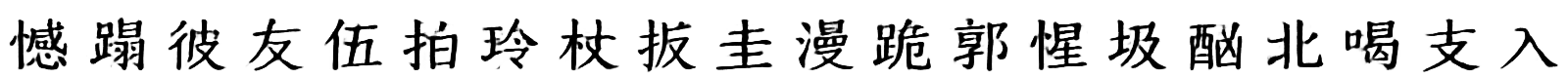}
\end{minipage}\\
\begin{minipage}{0.05\textwidth}O5:\end{minipage}
\begin{minipage}{0.45\textwidth}
\includegraphics[width=3.2in,height=0.15in]{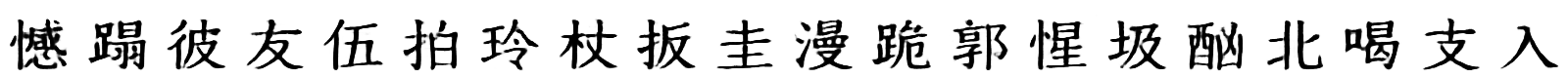}
\end{minipage}
\vspace{3pt}\\
\begin{minipage}{0.05\textwidth}TG:\end{minipage}
\begin{minipage}{0.45\textwidth}
\includegraphics[width=3.2in,height=0.15in]{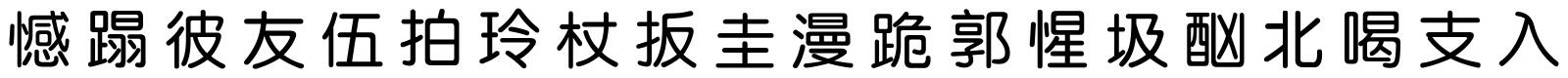}
\end{minipage}\\
\begin{minipage}{0.05\textwidth}O1:\end{minipage}
\begin{minipage}{0.45\textwidth}
\includegraphics[width=3.2in,height=0.15in]{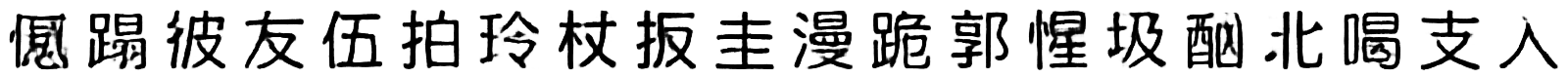}
\end{minipage}\\
\begin{minipage}{0.05\textwidth}O2:\end{minipage}
\begin{minipage}{0.45\textwidth}
\includegraphics[width=3.2in,height=0.15in]{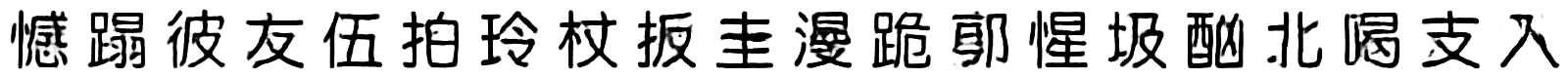}
\end{minipage}\\
\begin{minipage}{0.05\textwidth}O3:\end{minipage}
\begin{minipage}{0.45\textwidth}
\includegraphics[width=3.2in,height=0.15in]{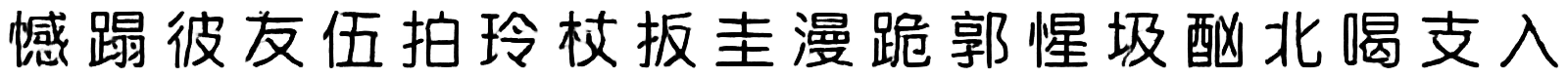}
\end{minipage}\\
\begin{minipage}{0.05\textwidth}O4:\end{minipage}
\begin{minipage}{0.45\textwidth}
\includegraphics[width=3.2in,height=0.15in]{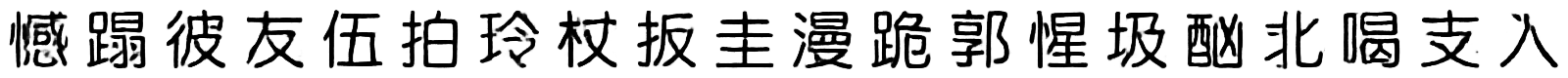}
\end{minipage}\\
\begin{minipage}{0.05\textwidth}O5:\end{minipage}
\begin{minipage}{0.45\textwidth}
\includegraphics[width=3.2in,height=0.15in]{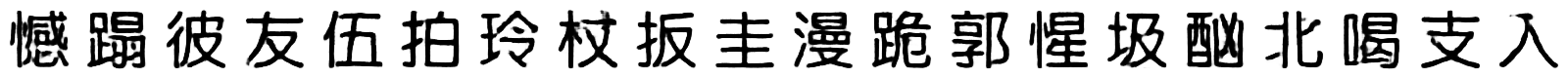}
\end{minipage}
\vspace{3pt}
\ruleh
\\
\begin{minipage}{0.05\textwidth}TG:\end{minipage}
\begin{minipage}{0.45\textwidth}
\includegraphics[width=3.2in,height=0.15in]{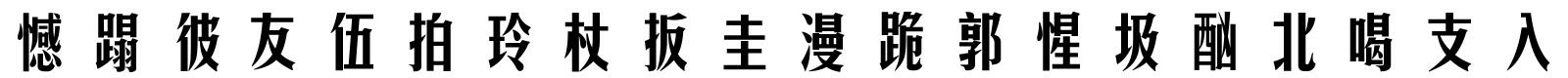}
\end{minipage}\\
\begin{minipage}{0.05\textwidth}O1:\end{minipage}
\begin{minipage}{0.45\textwidth}
\includegraphics[width=3.2in,height=0.15in]{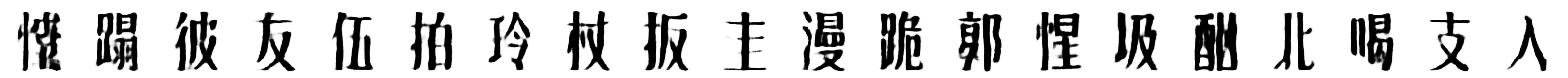}
\end{minipage}\\
\begin{minipage}{0.05\textwidth}O2:\end{minipage}
\begin{minipage}{0.45\textwidth}
\includegraphics[width=3.2in,height=0.15in]{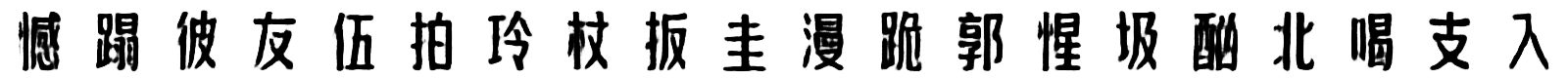}
\end{minipage}\\
\begin{minipage}{0.05\textwidth}O3:\end{minipage}
\begin{minipage}{0.45\textwidth}
\includegraphics[width=3.2in,height=0.15in]{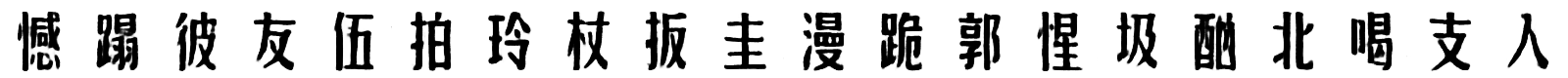}
\end{minipage}\\
\begin{minipage}{0.05\textwidth}O4:\end{minipage}
\begin{minipage}{0.45\textwidth}
\includegraphics[width=3.2in,height=0.15in]{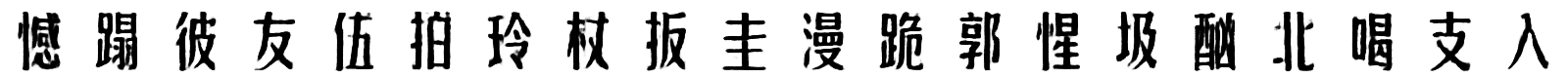}
\end{minipage}\\
\begin{minipage}{0.05\textwidth}O5:\end{minipage}
\begin{minipage}{0.45\textwidth}
\includegraphics[width=3.2in,height=0.15in]{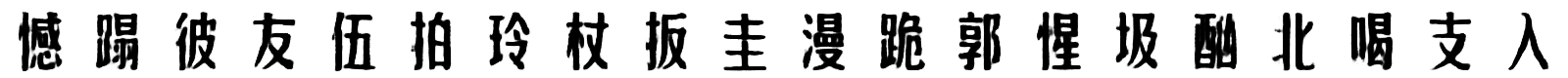}
\end{minipage}
\vspace{3pt}\\
\begin{minipage}{0.05\textwidth}TG:\end{minipage}
\begin{minipage}{0.45\textwidth}
\includegraphics[width=3.2in,height=0.15in]{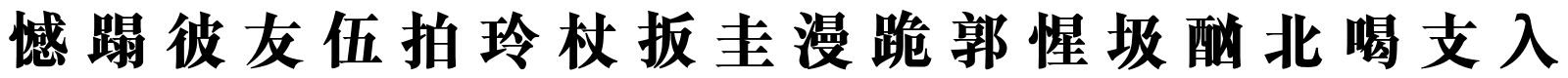}
\end{minipage}\\
\begin{minipage}{0.05\textwidth}O1:\end{minipage}
\begin{minipage}{0.45\textwidth}
\includegraphics[width=3.2in,height=0.15in]{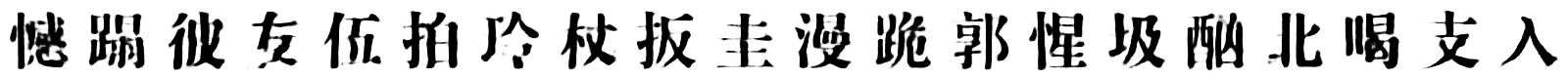}
\end{minipage}\\
\begin{minipage}{0.05\textwidth}O2:\end{minipage}
\begin{minipage}{0.45\textwidth}
\includegraphics[width=3.2in,height=0.15in]{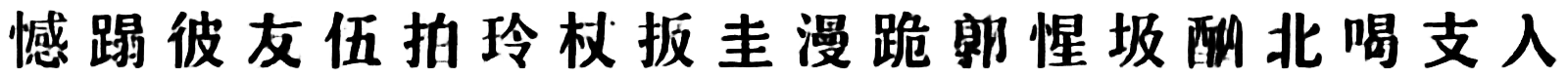}
\end{minipage}\\
\begin{minipage}{0.05\textwidth}O3:\end{minipage}
\begin{minipage}{0.45\textwidth}
\includegraphics[width=3.2in,height=0.15in]{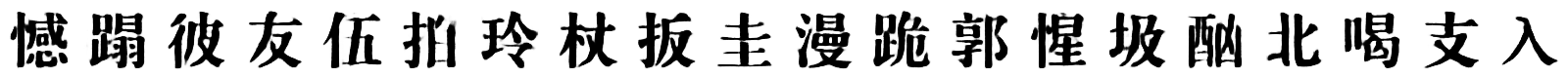}
\end{minipage}\\
\begin{minipage}{0.05\textwidth}O4:\end{minipage}
\begin{minipage}{0.45\textwidth}
\includegraphics[width=3.2in,height=0.15in]{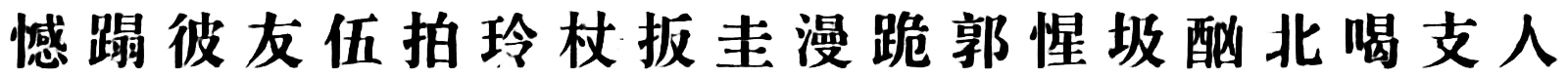}
\end{minipage}\\
\begin{minipage}{0.05\textwidth}O5:\end{minipage}
\begin{minipage}{0.45\textwidth}
\includegraphics[width=3.2in,height=0.15in]{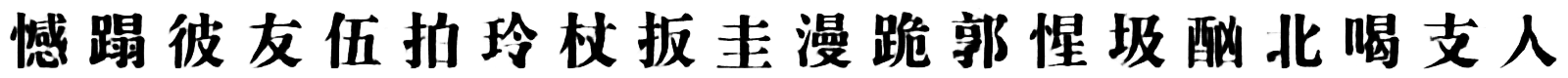}
\end{minipage}
\vspace{3pt}\\
\begin{minipage}{0.05\textwidth}TG:\end{minipage}
\begin{minipage}{0.45\textwidth}
\includegraphics[width=3.2in,height=0.15in]{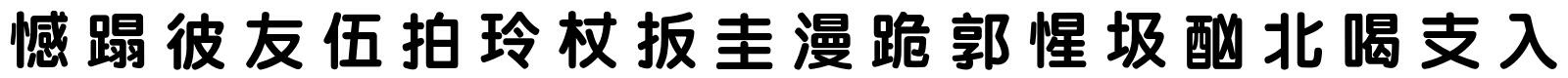}
\end{minipage}\\
\begin{minipage}{0.05\textwidth}O1:\end{minipage}
\begin{minipage}{0.45\textwidth}
\includegraphics[width=3.2in,height=0.15in]{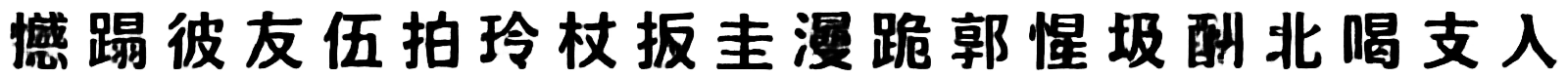}
\end{minipage}\\
\begin{minipage}{0.05\textwidth}O2:\end{minipage}
\begin{minipage}{0.45\textwidth}
\includegraphics[width=3.2in,height=0.15in]{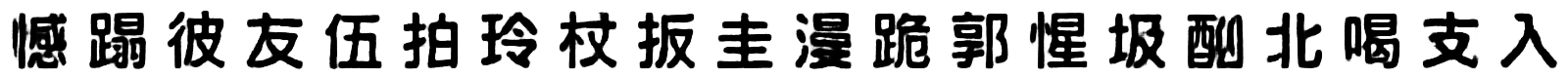}
\end{minipage}\\
\begin{minipage}{0.05\textwidth}O3:\end{minipage}
\begin{minipage}{0.45\textwidth}
\includegraphics[width=3.2in,height=0.15in]{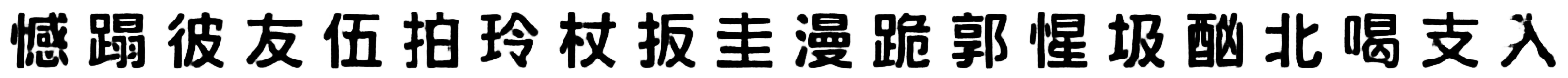}
\end{minipage}\\
\begin{minipage}{0.05\textwidth}O4:\end{minipage}
\begin{minipage}{0.45\textwidth}
\includegraphics[width=3.2in,height=0.15in]{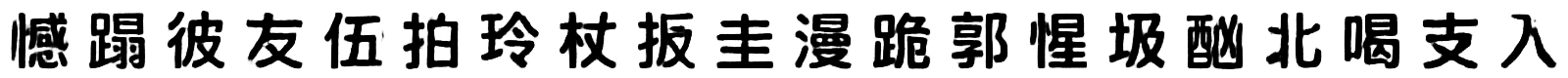}
\end{minipage}\\
\begin{minipage}{0.05\textwidth}O5:\end{minipage}
\begin{minipage}{0.45\textwidth}
\includegraphics[width=3.2in,height=0.15in]{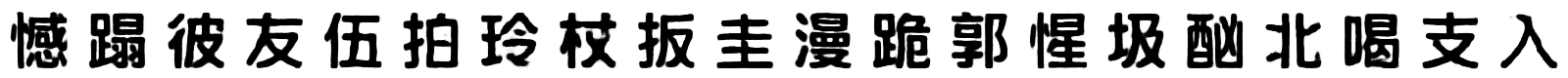}
\end{minipage}
}
\end{minipage}
}
\rulev
\hspace{-3pt}
\subfigure{
\begin{minipage}{0.45\textwidth}{\vspace{-5pt}
\includegraphics[width=3.2in,height=0.15in]{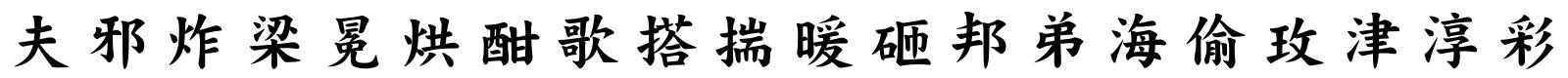}\\
\includegraphics[width=3.2in,height=0.15in]{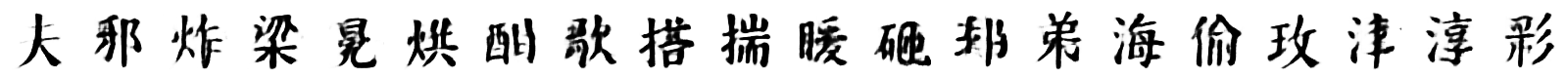}\\
\includegraphics[width=3.2in,height=0.15in]{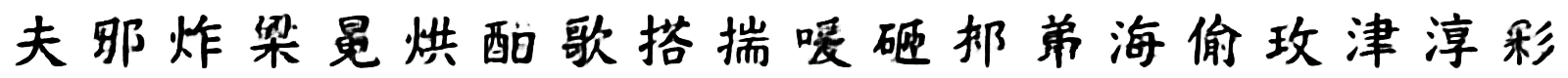}\\
\includegraphics[width=3.2in,height=0.15in]{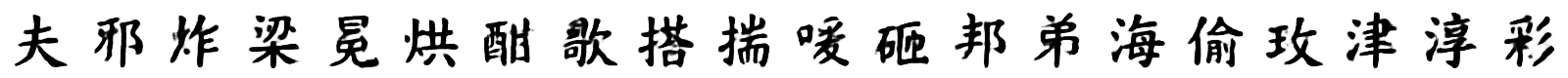}\\
\includegraphics[width=3.2in,height=0.15in]{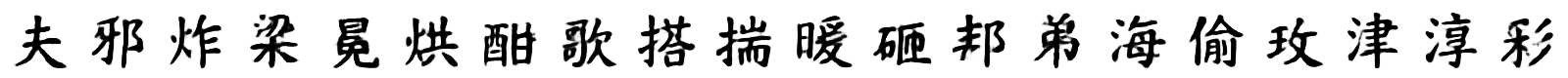}\\
\includegraphics[width=3.2in,height=0.15in]{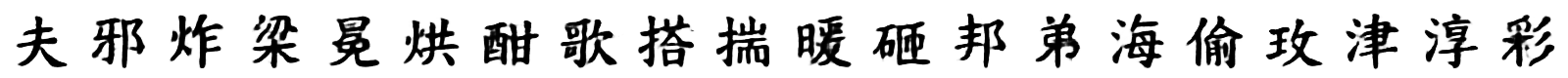}\vspace{3pt}\\
\includegraphics[width=3.2in,height=0.15in]{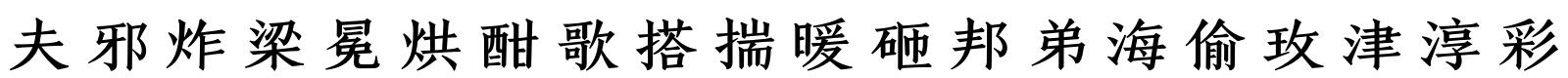}\\
\includegraphics[width=3.2in,height=0.15in]{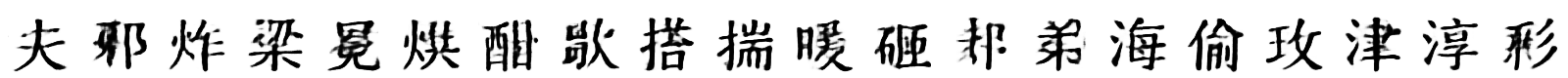}\\
\includegraphics[width=3.2in,height=0.15in]{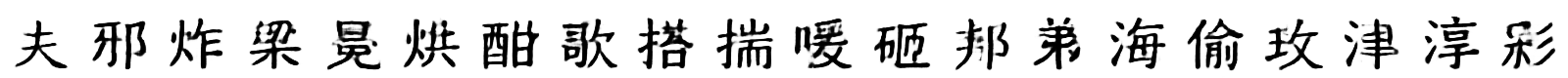}\\
\includegraphics[width=3.2in,height=0.15in]{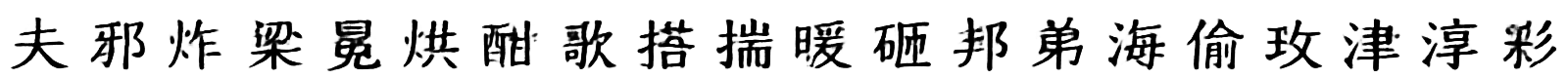}\\
\includegraphics[width=3.2in,height=0.15in]{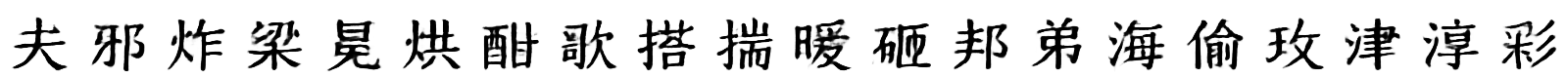}\\
\includegraphics[width=3.2in,height=0.15in]{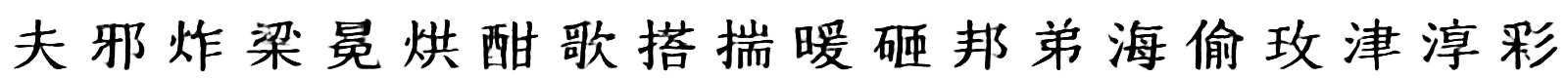}\vspace{3pt}\\
\includegraphics[width=3.2in,height=0.15in]{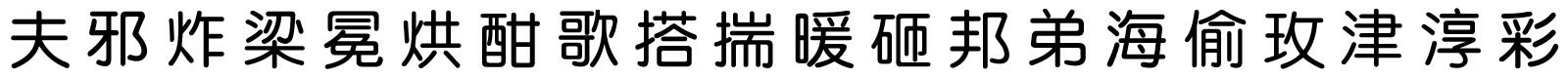}\\
\includegraphics[width=3.2in,height=0.15in]{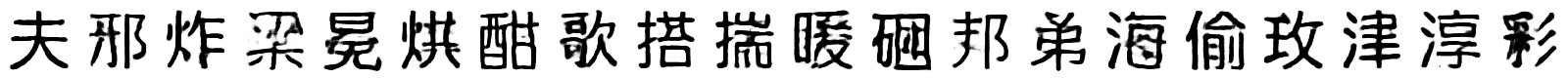}\\
\includegraphics[width=3.2in,height=0.15in]{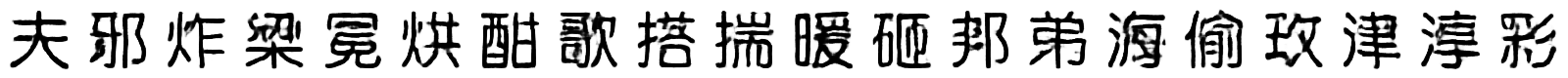}\\
\includegraphics[width=3.2in,height=0.15in]{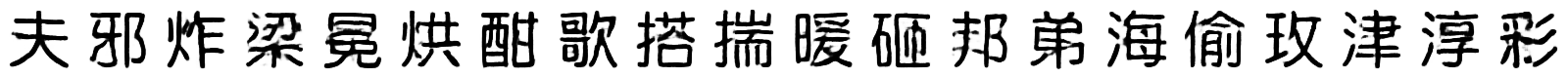}\\
\includegraphics[width=3.2in,height=0.15in]{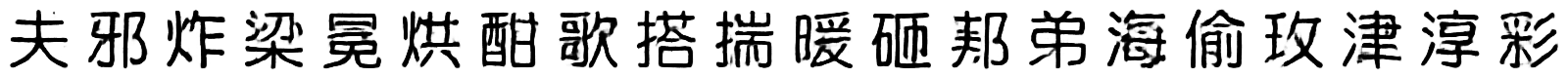}\\
\includegraphics[width=3.2in,height=0.15in]{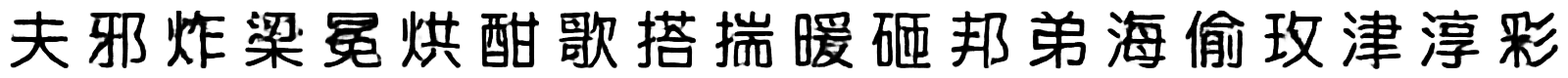}\vspace{3pt}
\ruleh
\vspace{4pt}
\\
\includegraphics[width=3.2in,height=0.15in]{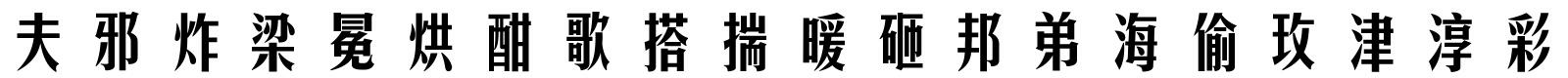}\\
\includegraphics[width=3.2in,height=0.15in]{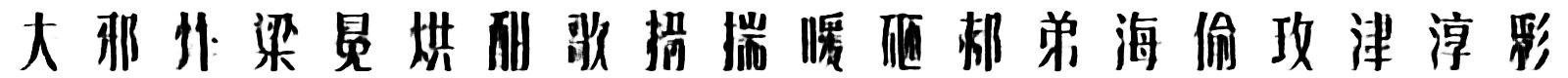}\\
\includegraphics[width=3.2in,height=0.15in]{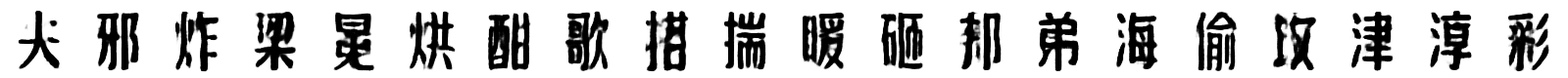}\\
\includegraphics[width=3.2in,height=0.15in]{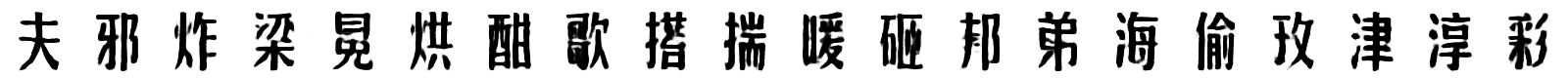}\\
\includegraphics[width=3.2in,height=0.15in]{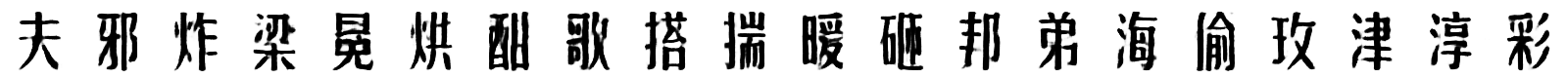}\\
\includegraphics[width=3.2in,height=0.15in]{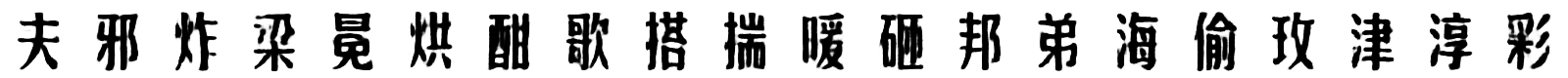}\vspace{3pt}\\
\includegraphics[width=3.2in,height=0.15in]{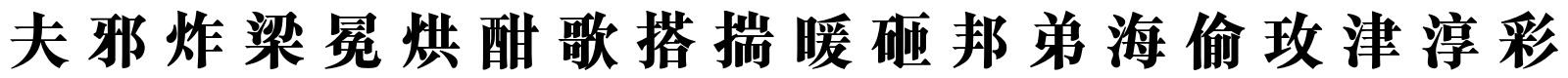}\\
\includegraphics[width=3.2in,height=0.15in]{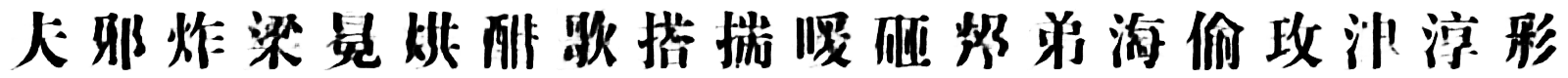}\\
\includegraphics[width=3.2in,height=0.15in]{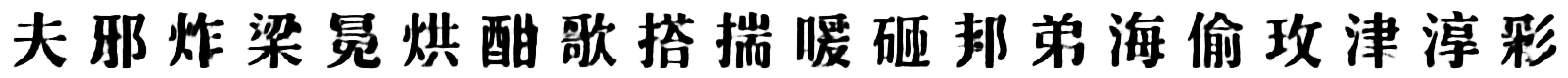}\\
\includegraphics[width=3.2in,height=0.15in]{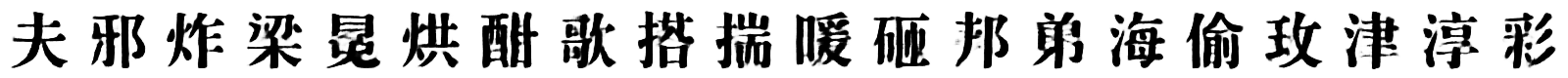}\\
\includegraphics[width=3.2in,height=0.15in]{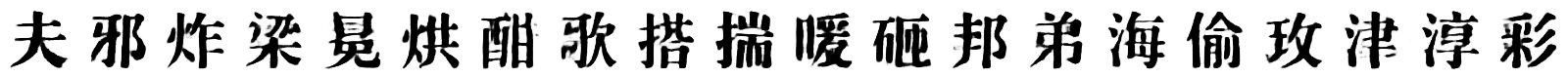}\\
\includegraphics[width=3.2in,height=0.15in]{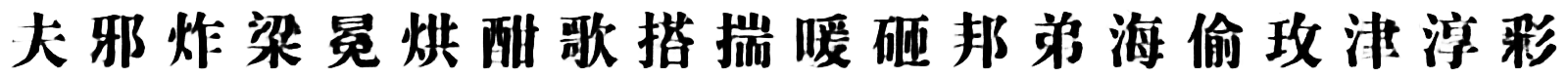}\vspace{3pt}\\
\includegraphics[width=3.2in,height=0.15in]{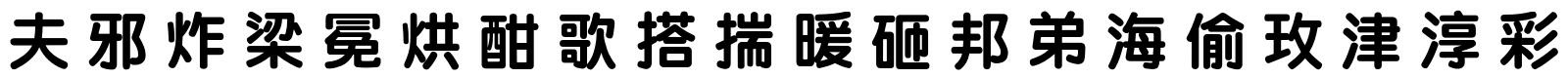}\\
\includegraphics[width=3.2in,height=0.15in]{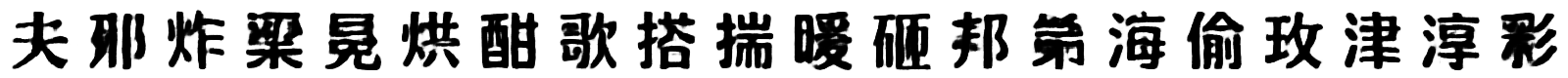}\\
\includegraphics[width=3.2in,height=0.15in]{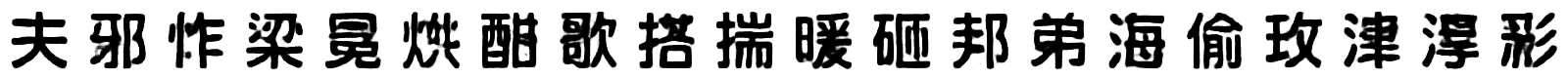}\\
\includegraphics[width=3.2in,height=0.15in]{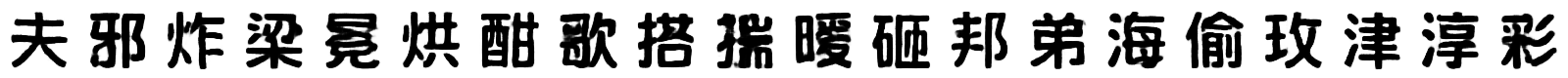}\\
\includegraphics[width=3.2in,height=0.15in]{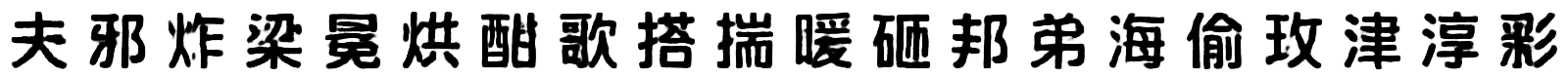}\\
\includegraphics[width=3.2in,height=0.15in]{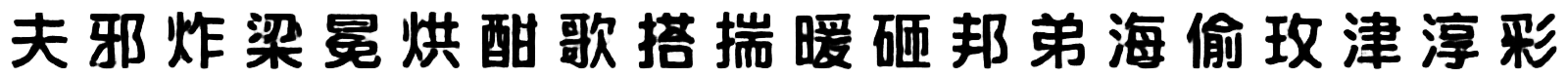}

}
\end{minipage}
}
\caption{The impact of the number of reference images on the generation of images in $D_1$, $D_2$, $D_3$, $D_4$, respectively (from upper left to lower right). TG: Target image, O1: Output for $r$=2, O2: Output for $r$=5, O3: Output for $r$=10, O4: Output for $r$=15, O5: Output for $r$=50. In all cases, $N_{t}$=300k.}
\vspace{-15pt}
\label{fig:ref-num_supp}
\end{figure*}

\section{Effect of the Weighted Loss}

In this subsection, we compare the model trained with L1 loss and weighted L1 loss. The quantitative results are displayed in Table~\ref{tab:com_loss_supp} and the qualitative results are shown in Figure~\ref{fig:comparison-loss_supp}. From the figure, we can observe that images with thin and light characters are generated better with weighted loss.

 \begin{table*}[!htbp]\small
     \centering
     \addtolength{\tabcolsep}{-3pt}
     \setlength{\abovecaptionskip}{-2pt}
     \setlength{\belowcaptionskip}{-6pt}
     \caption{Quantitative comparison of models with L1 loss and weighted L1 loss.}
     \begin{tabular}{c|ccc|ccc|ccc|ccc}
     \hline
         & \multicolumn{3}{c}{$D_1$} & \multicolumn{3}{|c}{$D_2$} & \multicolumn{3}{|c}{$D_3$} & \multicolumn{3}{|c}{$D_4$} \\

     \hline
                 & L1 loss & RMSE  & PDAR  & L1 loss & RMSE  & PDAR  & L1 loss & RMSE  & PDAR  & L1 loss & RMSE  & PDAR  \\
     \hline
     L1 loss &  \textbf{0.0091}  &  0.0186  &  0.1561  &  \textbf{0.0094}  &  \textbf{0.0189}  &  0.161  &  0.0093  &  \textbf{0.0187}  &  0.1554  &  0.0095  &  0.019  &  \textbf{0.1592}
 \\

     \hline
     Weighted L1 loss & \textbf{0.0091}  &  \textbf{0.0185}  &  \textbf{0.1549} &  \textbf{0.0094}  &  \textbf{0.0189}  &  \textbf{0.1604} &  \textbf{0.0092}  &  \textbf{0.0187}  &  \textbf{0.1549} &  \textbf{0.0094}  &  \textbf{0.0189}  &  \textbf{0.1592}\\

     \hline
     \end{tabular}%
     \vspace{-20pt}
   \label{tab:com_loss_supp}%
 \end{table*}

\begin{figure*}[!ht]
\centering
\setlength{\abovecaptionskip}{-2pt}
\hspace{-5pt}
\subfigure{
\begin{minipage}{0.47\textwidth}{
\begin{minipage}{0.06\textwidth}TG:\end{minipage}
\begin{minipage}{0.44\textwidth}
\includegraphics[width=3.0in,height=0.15in]{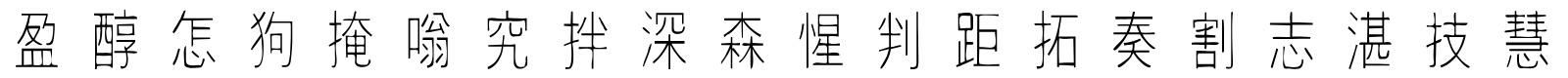}
\end{minipage}\\
\begin{minipage}{0.06\textwidth}O1:\end{minipage}
\begin{minipage}{0.44\textwidth}
\includegraphics[width=3.0in,height=0.15in]{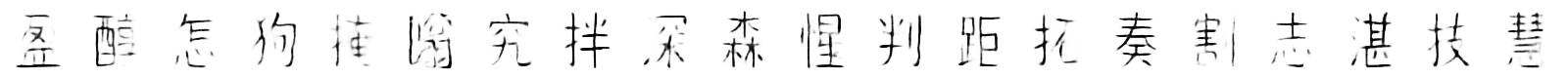}
\end{minipage}\\
\begin{minipage}{0.06\textwidth}O2:\end{minipage}
\begin{minipage}{0.44\textwidth}
\includegraphics[width=3.0in,height=0.15in]{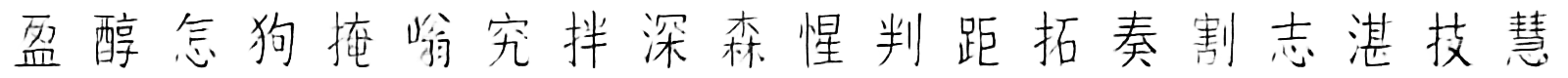}
\end{minipage}
}
\end{minipage}
\rulev
\begin{minipage}{0.47\textwidth}{
\begin{minipage}{0.06\textwidth}TG:\end{minipage}
\begin{minipage}{0.44\textwidth}
\includegraphics[width=3.0in,height=0.15in]{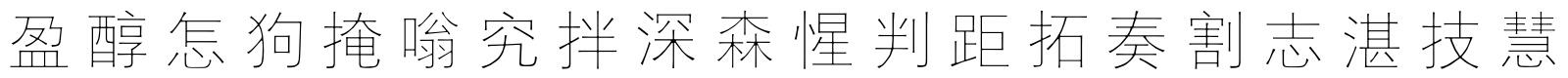}
\end{minipage}\\
\begin{minipage}{0.06\textwidth}O1:\end{minipage}
\begin{minipage}{0.44\textwidth}
\includegraphics[width=3.0in,height=0.15in]{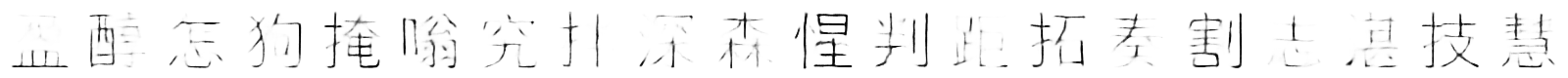}
\end{minipage}\\
\begin{minipage}{0.06\textwidth}O2:\end{minipage}
\begin{minipage}{0.44\textwidth}
\includegraphics[width=3.0in,height=0.15in]{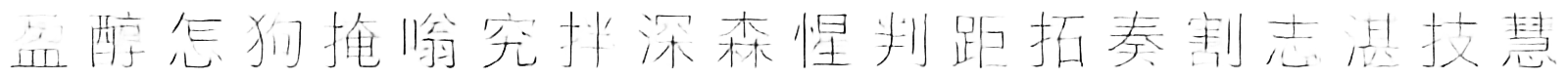}
\end{minipage}
}
\end{minipage}
}
\caption{Results of different loss functions with $N_{t}$=300k, $r$=10. TG: Target image, O1: Output for L1 as the loss function, O2: Output for weighted L1 loss as the loss function.}
\vspace{-15pt}
\label{fig:comparison-loss_supp}
\end{figure*}

\section{Results of One Reference Image}

\begin{figure*}[htb]
\centering
\setlength{\abovecaptionskip}{2pt}
\hspace{-6pt}
\includegraphics[height=1.4in,width=3.4in]{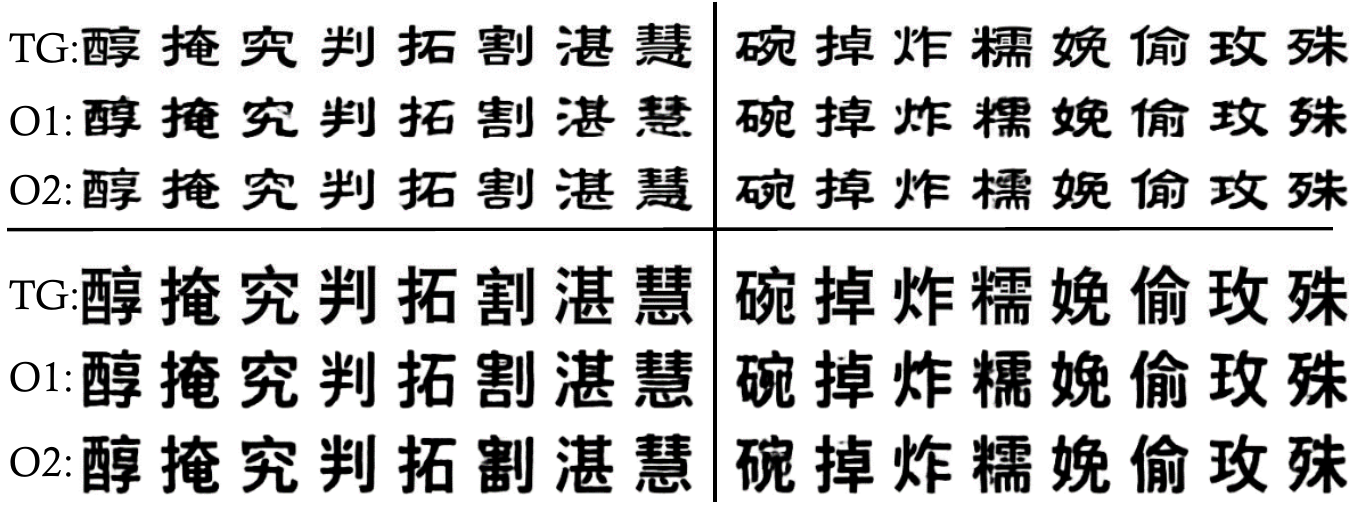}
\caption{Generation for $D_1$, $D_2$, $D_3$, $D_4$ (from upper left to lower right) for $N_{t}$=300k. TG: Target image, O1: Output for $r$=10, O2: Output for $r$=1.}
\label{fig:r=1_supp}
\vspace{-15pt}
\end{figure*}

We compare two models: $r$=10 vs. $r$=1 (splitting each former triplet into 100 triplets). As shown in Figure \ref{fig:r=1_supp}, the two models perform similarly, but the first model is more time efficient since it learns from $r^2$ style-content pairs at one time.

\section{Experiment for Neural Style Transfer}

\begin{figure*}[!htbp]
\setlength{\abovecaptionskip}{-2pt}
\centering
\vspace{-8pt}
\subfigure{
\includegraphics[height=2.6in,width=5.4in]{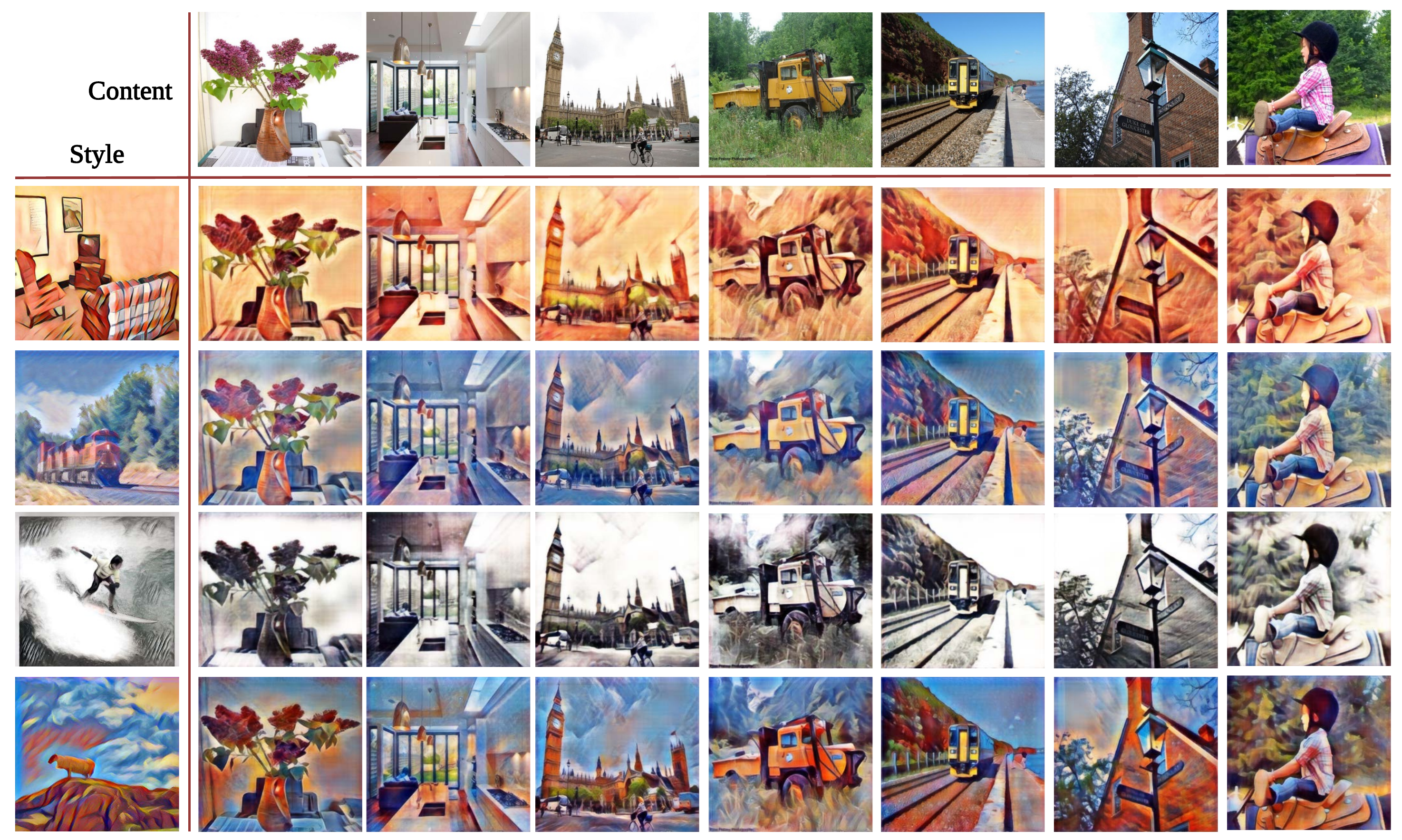}
}
\caption{Experiment results for neural style transfer.}
\label{fig:neural_image_supp}
\vspace{-10pt}
\end{figure*}

For neural style transfer, we constructed a dataset with artistic Photoshop filters which contains 106 styles and each style has 781 images with different contents. The results on the dataset are presented in Figure \ref{fig:neural_image_supp}, showing our method works well for neural images.

\end{document}